\documentclass[11pt,a4paper,onecolumn,copyright,gdm]{google}
\usepackage[authoryear,sort&compress,round]{natbib}
\uselogo{}
\renewcommand{\today}{2026-10-2}

\usepackage{hyperref}
\usepackage{xurl}

\usepackage{booktabs}
\usepackage{array}
\usepackage{longtable}
\usepackage{float}
\usepackage{pifont}
\usepackage{threeparttable}
\usepackage{graphicx}
\usepackage{caption}
\usepackage{enumitem}
\setlist[itemize]{leftmargin=*,itemsep=2pt,topsep=3pt,parsep=0pt}
\setlist[enumerate]{leftmargin=*,itemsep=2pt,topsep=3pt,parsep=0pt}
\setlist[description]{leftmargin=*,itemsep=2pt,topsep=3pt,parsep=0pt}

\newcommand{\rot}[1]{\rotatebox{90}{#1}}

\usepackage{xcolor}
\usepackage{colortbl}
\newcommand{\tablerowseparator}{\arrayrulecolor{black!35}\midrule\arrayrulecolor{black}}

\newcommand{\cmark}{\textcolor{green!55!black}{\ding{51}}}
\newcommand{\xmark}{\textcolor{red!75!black}{\ding{55}}}

\usepackage{wasysym}
\usepackage{listings}
\newcommand{\bktL}{\Circle}
\newcommand{\bktM}{\LEFTcircle}
\newcommand{\bktH}{\CIRCLE}

\DeclareRobustCommand{\pmark}{\textcolor{orange!90!black}{%
  \ooalign{\ding{51}\cr\kern0.20em\raisebox{0.45ex}{\rotatebox[origin=c]{115}{\rule{0.75em}{0.45pt}}}\cr}}}
\newcommand{\MV}{MultiVerse}

\title{MultiVerse: A Trustworthy Tool-Use Benchmark for Real-World Task Delegation under Harnesses}

\author[1,3,*,$\dagger$]{Sohee Yang}
\author[2,*]{Selcuk Gun}
\author[1,*]{Nora Kassner}
\author[5,*]{Rafael Mosquera-Gómez}
\author[1]{Minsuk Chang}
\author[1]{Yennie Jun}
\author[1]{Yuan Shangguan}
\author[1]{Julie Liu}
\author[1]{Idan Heimlich Shtacher}
\author[1]{Michele Bevilacqua}
\author[1]{Yang Song}
\author[1]{Abhijit Rao}
\author[1]{Sun Jae Lee}
\author[1]{Eric Zhu}
\author[1]{Chenmei Li}
\author[1]{Palak Jain}
\author[1]{Abhirut Gupta}
\author[1]{Jifan Zhu}
\author[1]{Shashi Narayan}
\author[1]{Stephan Lee}
\author[1]{Prateek Kolhar}
\author[2]{Anirudha Rayasam}
\author[2]{Xiaowei Li}
\author[5]{Kurt Bollacker}
\author[5,2]{Peter~Mattson}
\author[4,*]{Mor Geva}
\author[1,*]{Max Bartolo}
\author[1,*]{Sebastian Riedel}

\affil[1]{Google DeepMind}
\affil[2]{Google}
\affil[3]{UCL}
\affil[4]{Tel Aviv University}
\affil[5]{MLCommons}
\makeatletter
\g@addto@macro{\@author}{\par\vspace{\dimexpr\affilsep-\parskip\relax}{\Affilfont
\textsuperscript{*}Core contributor\quad
\textsuperscript{\(\dagger\)}Project lead\par}}
\makeatother

\begin{abstract}
As agents improve, users delegate increasingly complex tasks involving sequences of tool calls across multiple services. These requests are often incomplete or constrained, requiring agents to ask for clarification or follow instructions that change how they ordinarily use tools and communicate. Moreover, users often make requests through end-user harnesses, such as Hermes and OpenClaw, which change how models act. Therefore, a tool-use benchmark for such tasks must use realistic service interfaces, support multi-turn user interaction, assess instruction-based behavioral steering, and compare models across harnesses. Trustworthy evaluation also requires distinguishing agent failures from errors introduced by the infrastructure, harness integrations, user simulators, or verifiers. We present \MV{}, a benchmark for cross-service delegation that simulates 35 real-world services with 809 tools in a self-contained sandbox. Its tasks include behavioral instructions and multi-turn settings that test whether agents seek missing information from a reticent user simulator. We compare nine proprietary and three open-weight models on the same tasks under one bare harness (a simple loop) and four end-user harnesses. The strongest model--harness pair achieves a mean success rate of 70.0\%, compared with 52.4\% with an open-weight model. Instruction-following weaknesses differ across constraints and models. Agents struggle to obtain user-held information: in the multi-turn setting, 29.5\% of attempts end without the agent consulting the user for information needed to solve the task. No harness is consistently best. On average, end-user harnesses perform nearly identically to the bare harness, despite large gains and losses for particular model--harness--setting combinations. \MV{} enables the community to reliably test agent improvements in the various aspects that matter in real-world tool use.
\end{abstract}

\begin{document}
\maketitle
\par\noindent{\Affilfont
\href{https://multiverse-benchmark.com/}{\raisebox{-0.28\height}{\includegraphics[height=1.4em]{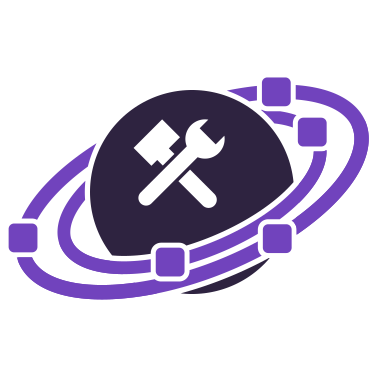}}\hspace{0.4em}\texttt{multiverse-benchmark.com}}\par}

\section{Introduction}

\begin{figure}[!t]
\centering
\includegraphics[width=\linewidth]{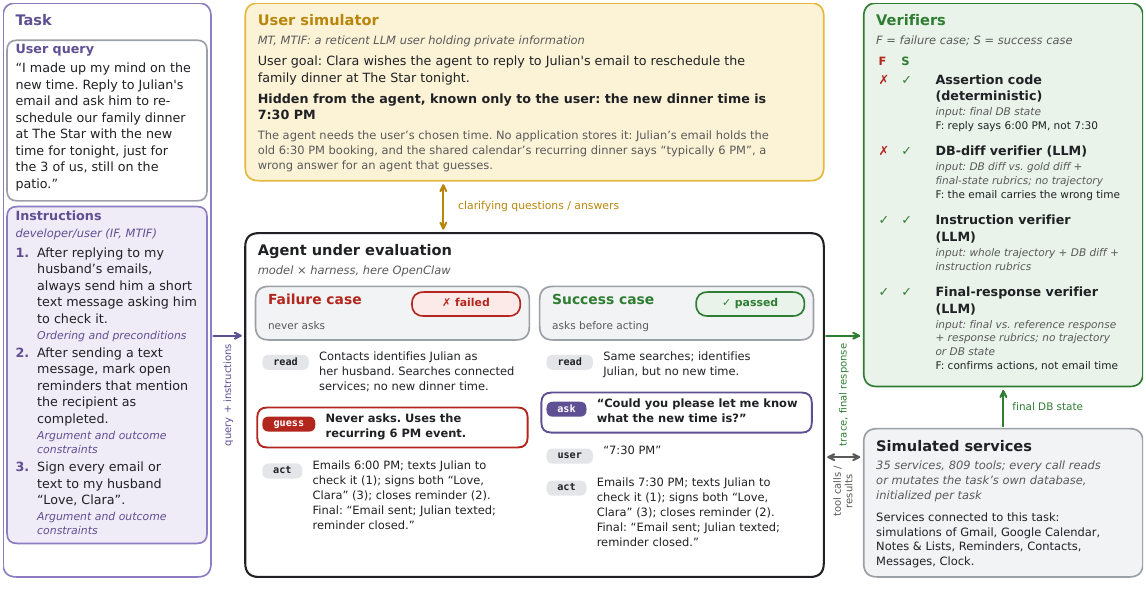}
\caption{Overview of \MV{}, adapted from a validation task, with trajectories and verdicts modified for illustrative purposes. Appendix Figures~\ref{fig:gui-initial} and~\ref{fig:gui-final} show services in a desktop GUI~(\S\ref{sec:gui}) before and after task completion.}
\label{fig:overview}
\label{fig:example-task}
\end{figure}

Consider a user running a Kimi K3 agent through OpenClaw as her personal assistant, asking it to reply to her husband's email to arrange a different time for tonight's family dinner (Figure~\ref{fig:example-task}). Her preferred time is not recorded in any connected service, so the agent needs to ask her. A standing instruction also requires it to send her husband a short text message after replying, asking him to check his email. The agent needs to search the relevant services, identify the missing information, consult the user, and follow specified constraints on how it acts. The end-user harness also shapes its behavior through its own system prompt, tool exposure, context management, and retry logic.

Evaluation of this kind of task delegation requires realistic service APIs; assessment of behavioral steering ability that measures adherence to instructions separately from task completion, on various categories of constraints including tool call orders and response formats; eliciting information that only the user knows; cross-harness comparisons; and trustable verifiers whose accuracy is checked to be high against human review. Trustworthy evaluation also must distinguish agent failures from attempts interrupted by infrastructure problems such as provider quota limits, and measure noise and errors introduced by simulated user responses, verifiers, and the model--harness integration. We organize these requirements into eleven tool-use benchmark design dimensions covering realism and trustworthy evaluation (\S\ref{sec:dimensions}).

Existing benchmarks only partially cover these dimensions (Table~\ref{tab:comparison-compact}; full comparison in \S\ref{sec:full-comparison}). Among the widely used ones, only $\tau^3$-bench~\citep{sierra2026tau3} has an interactive user simulator, and none report instruction-adherence rates across multiple categories. Terminal-Bench and Agents' Last Exam analyze only a subset of model--harness pairs~\citep{merrill2026terminalbench,sun2026agents}.

We introduce \MV{} (MULTI-serVice Evaluation of Real-world Steering and Elicitation), a benchmark designed to cover all eleven design dimensions. \MV{} simulates 35 real-world services, including Gmail, Slack, and Stripe, exposing 809 tools implemented from their documentation. Each task runs in a self-contained sandbox with the same initial state across different runs. It provides 396 validation tasks and 400 hidden test tasks across four settings: single-turn execution (ST), single-turn execution with behavioral steering instructions (IF), multi-turn clarification (MT), and multi-turn clarification with behavioral steering instructions (MTIF). The instruction-bearing settings report adherence rates by behavioral category, separately from task completion. The multi-turn settings test whether agents recognize and request information that is unavailable in the environment but held by a simulator, whose accuracy was reviewed by humans; no violations of the simulator's response guidelines were found in 97 of 100 audited conversations. We score agent attempts\footnote{An attempt is one execution of a task using a model under a certain harness.} with deterministic assertions over final service states and complementary LLM judges, whose majority-vote decisions agreed with human review in 92--100\% of audited cases.

We evaluate nine proprietary and three open-weight models, each under a bare harness and four end-user harnesses, in three runs. GPT~6 Astra with OpenHands achieves the highest mean success rate, 70.0\%, and the highest mean success rate for an open-weight model is 52.4\% (Kimi K3 with OpenClaw). In IF and MTIF, 30.6\% of failed attempts meet the task's outcome requirements but violate behavioral instructions on how to carry it out. Response content and presentation, clarification and assumptions, and process constraints have the highest average violation rates. Agents also struggle to obtain user-held information: 29.5\% of MT attempts end after a single turn due to the agent not consulting the user for more information.

Among the models and settings evaluated, no harness is consistently best. End-user harnesses score only 0.05 percentage points below the bare harness on average, but their effects vary by model and setting. On MT, OpenHands scores 10.3 points above Bare for Sonnet~5 but 10.7 points below it for Opus~4.6, with both directions holding in all three runs. For Fable~5, OpenClaw scores higher on ST but lower on MT than Bare, and more MT attempts end without a user reply across all three runs.

In summary, we release \MV{}, a realistic and trustworthy tool-use benchmark that combines high-quality tasks spanning behavioral instructions and multi-turn clarification, extensively tested simulations of 35 real-world services with a GUI, and evaluation code for comparing models across harnesses, so that the community can reliably test agent improvements against it and build on it.

\begin{table}[tbp]
\centering
\small
\setlength{\tabcolsep}{3pt}
\newcommand{\tdimension}[1]{\,\ensuremath{_{\textcolor{black!55}{\mathrm{D#1}}}}}
\caption{Benchmark comparison (D1--D11 mark design dimensions in Figure~\ref{fig:design-dimensions}). \cmark~satisfied, \pmark~partial, \xmark~not satisfied; --~not applicable, unreported, or retrieved at run time. Low (\bktL), mid (\bktM), and high (\bktH) buckets, exact counts, and citations: Appendix Table~\ref{tab:comparison} (\S\ref{sec:full-comparison}). Unlike in the other columns, a fuller \#~Tasks circle does not mean a better benchmark (D11).}
\label{tab:comparison-compact}
\resizebox{0.8\textwidth}{!}{%
\begin{tabular}{@{}lccccccccccccccc@{}}
\toprule
& \multicolumn{8}{c}{\textbf{Realism}} & \multicolumn{7}{c}{\textbf{Trustworthy evaluation}} \\
\cmidrule(lr){2-9} \cmidrule(lr){10-16}
& \rot{\shortstack[l]{Real-world\\[-2pt]tools\tdimension{1}}}
& \rot{\# Services\tdimension{2}}
& \rot{\# Tools\tdimension{2}}
& \rot{\# Tools/task\tdimension{2}}
& \rot{\shortstack[l]{\# Services/\\[-2pt]task\tdimension{2}}}
& \rot{User simulator\tdimension{3}}
& \rot{\shortstack[l]{Category-wise\\[-2pt]steering\\[-2pt]evaluation\tdimension{4}}}
& \rot{\shortstack[l]{Cross-harness\\[-2pt]evaluation\tdimension{5}}}
& \rot{\shortstack[l]{Hidden test\\[-2pt]set\tdimension{6}}}
& \rot{\shortstack[l]{Service-state\\[-2pt]verification\tdimension{7}}}
& \rot{Self-contained\tdimension{8}}
& \rot{\shortstack[l]{Infra-error\\[-2pt]isolation\tdimension{9}}}
& \rot{\shortstack[l]{Validated user\\[-2pt]simulator\tdimension{10}}}
& \rot{\shortstack[l]{Validated\\[-2pt]evaluation\tdimension{10}}}
& \rot{\# Tasks\tdimension{11}} \\
\midrule
\multicolumn{16}{@{}l}{\emph{Widely used benchmarks}} \\[1pt]
GDPval & -- & -- & -- & -- & -- & \xmark & \xmark & \xmark & \cmark & \xmark & \xmark & -- & -- & \cmark & \bktH \\
Terminal-Bench 2.0 & \cmark & -- & -- & -- & -- & \xmark & \pmark & \pmark & \xmark & \cmark & \xmark & \pmark & -- & \pmark & \bktL \\
Agents' Last Exam & \cmark & -- & -- & -- & -- & \xmark & \pmark & \pmark & \cmark & \cmark & \xmark & \cmark & -- & \pmark & \bktH \\
MCP-Atlas & \cmark & \bktH & \bktM & \bktL & -- & \xmark & \xmark & \xmark & \cmark & \xmark & \xmark & \pmark & -- & \pmark & \bktH \\
BFCL v4 & \pmark & -- & \bktH & \bktL & -- & \xmark & \pmark & \xmark & \xmark & \cmark & \xmark & \xmark & -- & \xmark & \bktH \\
$\tau^3$-bench & \xmark & \bktL & \bktM & \bktL & \bktL & \cmark & \pmark & \pmark & \xmark & \cmark & \cmark & \pmark & \cmark & \pmark & \bktM \\
\midrule
\multicolumn{16}{@{}l}{\emph{Closest benchmarks}} \\[1pt]
Toolathlon & \cmark & \bktH & \bktH & \bktH & \bktM & \xmark & \pmark & \xmark & \xmark & \cmark & \xmark & \pmark & -- & \pmark & \bktL \\
AppWorld & \cmark & \bktM & \bktM & -- & \bktH & \xmark & \pmark & \pmark & \pmark & \cmark & \cmark & \xmark & -- & \pmark & \bktM \\
HANDBOOK.md & \cmark & \bktL & \bktL & \bktH & \bktM & \xmark & \pmark & \xmark & \xmark & \cmark & \cmark & \cmark & -- & \pmark & \bktL \\
AutomationBench & \cmark & \bktH & \bktM & -- & -- & \xmark & \pmark & \xmark & \cmark & \cmark & \cmark & \pmark & -- & \xmark & \bktM \\
GAIA2 & \xmark & \bktM & \bktM & \bktH & \bktH & \xmark & \pmark & \pmark & \cmark & \pmark & \cmark & \pmark & -- & \cmark & \bktH \\
\midrule
\textbf{\MV{} (ours)} & \cmark & \bktH & \bktH & \bktH & \bktH & \cmark & \cmark & \cmark & \cmark & \cmark & \cmark & \cmark & \cmark & \cmark & \bktM \\
\bottomrule
\end{tabular}%
}
\end{table}

\section{Design dimensions for tool-use benchmarks}
\label{sec:dimensions}

\begin{figure}[htbp]
\centering
\includegraphics[width=\linewidth]{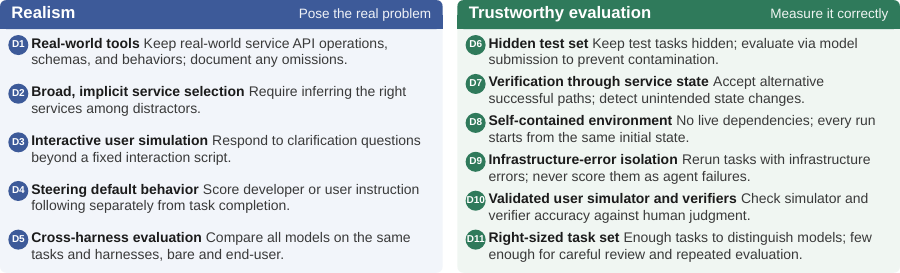}
\caption{Design dimensions for realistic delegation and trustworthy evaluation.}
\label{fig:design-dimensions}
\end{figure}

A benchmark must pose the real problem and measure it correctly. Our eleven design dimensions organize these needs around realism and trustworthy evaluation, where the latter includes accounting for errors introduced by harnesses, user simulators, and verifiers. Figure~\ref{fig:design-dimensions} summarizes the dimensions, which we use to compare benchmarks in Table~\ref{tab:comparison-compact} and Appendix Table~\ref{tab:comparison}. Full definitions appear in \S\ref{sec:dimensions-details}.

\section{Related work}
\label{sec:positioning}

Toolathlon~\citep{li2026tool}, AppWorld~\citep{trivedi2024appworld}, HANDBOOK.md~\citep{surge2026handbook}, AutomationBench~\citep{shepard2026automationbench}, and GAIA2~\citep{froger2025are} are the closest to our benchmark because they also evaluate task delegation across multiple services. None of these provides a user simulator that answers agent-initiated clarification questions (Table~\ref{tab:comparison-compact}). The first four evaluate agent execution without further user input; GAIA2 provides asynchronous interactions with simulated mobile applications, but its user turns are scripted. These setups do not test whether an agent can identify missing user-held information, ask for it, and use the reply. All five contain some instructions about agent actions or responses, but do not provide the category-level analysis of instruction adherence required by our steering criterion. In contrast, \MV{} uses a dedicated verifier to assess instruction adherence separately from task completion and reports results across nine categories, including action-order and response-format requirements.

These benchmarks also differ in reproducibility and evaluation: Toolathlon depends on live services; in the AppWorld codebase we audited, provider failures can consume the step budget while the attempt is scored normally; and GAIA2 compares actions against a reference graph. None reports a complete model-by-harness evaluation, and among the 33 other benchmarks we reviewed (\S\ref{sec:full-comparison} and \S\ref{sec:full-audit}), none combines real tools in a self-contained sandbox with responsive user simulation, steering evaluation by category of instruction, and full model-by-harness evaluation on the same tasks. On the Agentic Benchmark Checklist~\citep{zhu2025establishing}, \MV{} satisfies 28 of 33 applicable checklist items (0.85), the highest total among the six benchmarks assessed (\S\ref{sec:abc}).

\section{\MV{}}
\label{sec:benchmark}

\MV{} evaluates an agent's ability to accomplish delegation tasks by calling simulated tools of real-world services, entirely within a self-contained, reproducible simulation. Figure~\ref{fig:overview} shows the components and the evaluation process.

\subsection{Tools}
\label{sec:tools}

\MV{} simulates 35 widely used real-world services (e.g., Gmail, Google Calendar, Drive, Slack, WhatsApp, Zendesk, Salesforce, HubSpot, Stripe, Instagram, GitHub, and Spotify)\footnote{All reported runs and the released benchmark use pseudo service names (e.g., Gmail $\rightarrow$ \texttt{gridwell\_mail} and Jira $\rightarrow$ \texttt{zira}), while preserving the documented tool schemas. On an earlier version of the task set, a pilot study with Claude Opus~4.5, Gemini~3 Pro, and GPT~5.2 found no statistically significant change in success after renaming. We use the original service names in the paper; the mapping is in \S\ref{sec:service-mapping}.} that together expose 809 tools, the individual functions an agent calls (e.g., \texttt{slack\_\allowbreak{}send\_\allowbreak{}message}, \texttt{stripe\_\allowbreak{}create\_\allowbreak{}refund}, \texttt{gmail\_\allowbreak{}list\_\allowbreak{}messages}). The services were implemented from API or implementation documentation; \S\ref{sec:env-gen} describes their implementation and testing in detail, and also explains which functionalities we left out and which behaviors of the real service we reproduced. An engineering team at a vendor built, tested, and corrected the services with the help of coding agents. The services underwent internal quality control (QC) by the vendors who developed the APIs, independent QC by another vendor, and informal author QC of API subsets. The task construction vendor also reported the bugs they found during their work, and the engineering team fixed the raised issues (Appendix Figure~\ref{fig:vendor-workflow}). The implementations were validated with ~33K unit tests all passing, covering 88\% of executable statements (79--98\% per service).

The tools show the complexity of real-world APIs. Across the 809 tool definitions, input schemas average 3.2 top-level parameters and 5.9 leaf parameters (individual input fields after expanding nested objects), counting both required and optional inputs. Some schemas are substantially larger: \texttt{salesforce.update\_event} exposes 30 top-level parameters, while \texttt{google\_slides.\allowbreak batch\_update\_presentation} exposes 84 leaf fields across its supported update requests and write controls, with text-color fields nested seven levels deep. For comparison, AppWorld's 457 tools~\citep{trivedi2024appworld} average 3.2 leaf input parameters, none nested (\S\ref{sec:tool-schema-comparison}).

Each service is backed by a structured database that tool calls deterministically read and mutate, and each task ships its own initial database state, so the same agent behavior always produces the same service outcome. We also release a desktop GUI with the tools for inspecting service states and interacting with the simulated services (\S\ref{sec:gui}; Appendix Figures~\ref{fig:gui-initial} and~\ref{fig:gui-final}).

\subsection{Tasks}
\label{sec:tasks}

Each task specifies an initial service state, a natural-language query, and optional developer/user instructions.
\MV{} comprises 796 tasks~(396 validation, 400 hidden test) across four task settings that emphasize different demands. The validation set contains 100 Single-Turn (ST) tasks that test single-turn search, synthesis, and correct action; 97 Instruction Following (IF) tasks with additional developer/user instructions that require departing from default behavior; 100 Multi-Turn (MT) tasks designed to test whether agents obtain missing information from a reticent user; and 99 MTIF tasks that combine both. The hidden test set contains 100 tasks per setting. To prevent test set leakage and support fair comparison across model providers, the test set development and access have been restricted to only the lead author and the vendors, who have no access to foundational model training. Also, the test set and a model-submission-based leaderboard will be maintained by a nonprofit organization specializing in AI benchmarking.

Across the 396 validation tasks, agents have access to a median of 148.5 candidate tools (maximum 324) across a median of six services (maximum nine) (Appendix Figure~\ref{fig:tool-stats}, \S\ref{sec:diversity}). Queries often do not explicitly state which services the agent needs to search through, leaving the identification of the services that contain the necessary information as a part of the task requirements. Each task includes one reference (gold) solution: a sequence of tool calls and agent responses that solves it. Gold solutions contain a median of 17 tool calls, and every solution uses at least three distinct tools across at least two services (\S\ref{sec:diversity}).

\paragraph{Behavioral instructions} We define IF tasks in a narrower sense than the name implies: although completing the task is itself a form of instruction following, our IF tasks assess adherence to additional constraints on actions and communication, such as checking prerequisites before taking an action, grounding claims in retrieved evidence, or using a prescribed final-response format. The nine instruction categories are defined in Appendix Table~\ref{tab:if-groups} (\S\ref{sec:if-types}). An agent can reach the target final service state while violating an instruction, such as a required response format. We score instruction compliance separately from task completion, while requiring both for overall success.

\paragraph{Construction} Human annotators constructed tasks to be solvable but challenging for strong reference models~(\S\ref{sec:task-construction}). Each task had to have a successful gold solution, produced by guiding an agent when needed, while a randomly chosen reference model had to fail to solve it without guidance from the human annotator. The intended cause of failure differed by setting: in ST, the difficulty came from solving the task. In IF and MTIF, the task was designed to be solvable by the reference model, but the added behavioral steering instructions were intended to cause failure, whereas in MT, consulting the user for missing information was designed to be the challenge. Differences between settings therefore reflect this design, not the effect of adding one factor to otherwise identical tasks.

For each task, an annotator randomly sampled two to seven services for the task to use, which broadened coverage across services; sampled services the task could not use stayed connected as distractors~(\S\ref{sec:diversity}). The annotator then specified the intended task and constraints on the initial service state, and generated and verified the initial databases with the help of an LLM~(\S\ref{sec:workspace-gen}).
Each task passed a fixed sequence of automated and human QC at the annotation vendor before delivery~(\S\ref{sec:construction}): three QC passes examined task validity, verifier correctness, and consistency between the two. A second, independent vendor performed QC on every delivered task, comparing multiple model attempts per task to see how suspected defects affected execution or scoring, so that the QC team could separate task defects from model errors~(\S\ref{sec:qc}). The authors also conducted QC on subsets of tasks and sent feedback to the annotation vendor. The annotation vendor was responsible for correcting the task defects the QC vendor identified. About 300 annotators, selected through QC after training, created the tasks, and a QC team of about 60 members at the second vendor checked them, iterating on task quality across 1,000 tasks. Appendix Figure~\ref{fig:vendor-workflow} shows the overall workflow.

\subsection{User simulator}
\label{sec:usersim}

For MT and MTIF, annotators created requests with missing information that the agent was expected to obtain from the user, such as what time the new family dinner should be. The user simulator holds these details and answers questions about them. The simulator is an LLM (Gemini 3.5 Flash) driven by one user goal and a set of hidden information items written for each task (\S\ref{sec:usersim-construction}; prompt in \S\ref{sec:prompts}). The 199 multi-turn validation tasks contain 268 hidden items, a median of one per task.

The simulator is deliberately reticent. It answers direct questions using only its hidden information and declines questions outside that information. The simulator may also correct an explicitly stated incorrect value, provided the corresponding action has not been completed or can still be modified. Thus, a direct question is not the only way hidden information can surface, but the agent cannot obtain information from the simulator that it should retrieve from the services.

We confine the simulator to clarification to limit hallucination, keep its replies checkable against its guidelines and the task's hidden information, and support accurate responses. Human reviewers found no guideline violations in 97 of 100 conversations containing at least one simulator reply, covering ten models, five harnesses, and both multi-turn settings (\S\ref{sec:qc}). Prior work illustrates the importance of properly constraining the role of the user simulator: $\tau^2$-bench found errors in 40--47\% of conversations with the simulator it inherited from $\tau$-bench, and still 16\% in its new telecom domain, a drop its authors attribute to constraining user behavior~\citep{barres2025tau2bench}.

\subsection{Verifiers}
\label{sec:verifier}

Four complementary verifiers assess service state, the agent's report to the user, and instruction adherence, as applicable to each task. Rubrics specify what the LLM judges check. The requirements can overlap; an instruction may also constrain the required final state or response.

\begin{itemize}
\item \textbf{Assertion code} Deterministic code checks required properties of the final service state, such as whether one reply was sent in Julian's email thread (Figure~\ref{fig:example-task}).
\item \textbf{DB-diff verifier} An LLM judge compares the agent's database changes with those made by the gold solution, using applicable final-state rubrics and allowing valid alternatives. It checks whether the email requests the intended reservation change and whether unrelated service state was changed inappropriately; the wording need not match the gold solution.
\item \textbf{Final-response verifier} An LLM judge compares the agent's final message to the user with an annotator-written reference response, using applicable response rubrics. Here, it evaluates the report to Clara, not the email sent to Julian. It also checks answers to information-seeking requests that require no database change.
\item \textbf{Instruction verifier} An LLM judge checks compliance with the developer/user instructions against one or more rubrics per instruction, using the trajectory, database state, and/or response. In Figure~\ref{fig:example-task}, the agent must also send the additional text message, close the relevant reminders, and sign both messages ``Love, Clara''.
\end{itemize}

An attempt succeeds only if it passes every applicable verifier. The assertion code and DB-diff verifier are both state verifiers that check the DB state. Appendix Table~\ref{tab:verifier-inputs} details verifier inputs, outputs, and applicability; only the instruction verifier reads the trajectory.

The default LLM judges use Gemini 3.1 Pro, with three separately generated judgments and majority voting; the prompts are in \S\ref{sec:prompts}, and \S\ref{sec:qc} documents exceptions to this procedure. We audited 200 trajectories covering ten models and five harnesses across the four task settings. Reviewers inspected the task evidence alongside a sampled judge verdict and its explanation. The three-vote majority decisions agreed with their final assessments in 97.5\% of DB-diff checks, 92\% of final-response checks, and 100\% of instruction checks (\S\ref{sec:qc}).

\paragraph{Use of verifiers for task construction and correction}
During task construction, we ensured that the gold solutions passed all the applicable verifiers on automatic replay. After performing all the experiments, we used disagreements between assertion code and the DB-diff verifier to identify and correct overly strict assertions. An LLM review of sampled disagreements estimated that the assertion code incorrectly rejected 2.1\% of attempts before the correction (\S\ref{sec:qc}). 394 of the 396 validation tasks (99.5\%) were solved at least once in our experiments, which shows that nearly all validation tasks are empirically solvable even by the current models.

\subsection{Harnesses and evaluation protocol}
\label{sec:harness-protocol}

Agents run under the bare harness or one of four end-user harnesses (Hermes~\citep{nousresearch2026hermes}, OpenCode~\citep{anomalyco2026opencode}, OpenHands~\citep{wang2025openhands}, and OpenClaw~\citep{openclaw2026}). We compare how different end-user harnesses guide the same model through the same tasks, with the same service tools and model-call budget. We restrict native capabilities that could expose gold solutions or let agents bypass the simulated services, and repair integration bugs that prevent some models from running correctly~(\S\ref{sec:harness-config}). The harnesses still differ in their instructions to the model and how they present and execute tool calls.

We record why each attempt ended and distinguish infra problems from agent failures. Transient provider, container, and verifier failures are classified as transient infra issues; they are never scored and are marked for retry, whereas outcomes the agent could have avoided, such as malformed output, refusals, unparseable tool calls, context overflow, and budget exhaustion, count as execution-completion failures (\S\ref{sec:harness-config}). Each task runs in its own container, with models accessed through provider APIs, so that one task's execution cannot modify another's environment.

\section{Experimental results}
\label{sec:results}

\subsection{Setup}
\label{sec:setup}

We evaluate twelve models, nine proprietary and three open-weight (Appendix Table~\ref{tab:identifiers}), under five harnesses: the bare harness (Bare) and four end-user harnesses, Hermes, OpenCode, OpenHands, and OpenClaw. The evaluation uses 396 validation tasks: 100 each of ST and MT, 97 IF, and 99 MTIF. We run each experiment three times, which gives 12$\times$5$\times$396$\times$3 = 71{,}280 scored attempts. Models are configured for high reasoning effort where supported, with provider and harness settings given in \S\ref{sec:harness-config}. The open-weight models are accessed through OpenRouter at the quantization levels listed in Appendix Table~\ref{tab:identifiers}. All configurations use the same task tools and model-call budget, with a shared retry policy for provider errors, since these choices can materially affect performance~\citep{zhang2026stop}. Harness-specific tool interfaces and context-compaction settings are described in \S\ref{sec:harness-config}. Task-specific developer/user instructions are placed in the system prompt under the bare harness and in the first user message under end-user harnesses (\S\ref{sec:harness-config}). Each attempt is capped at 320 model calls. Headline average scores weight the four task settings equally. Across-run standard deviations (SDs) are sample standard deviations of the three run results. Task-sampling confidence intervals use the bootstrap~\citep{efron1979bootstrap}. We draw 10{,}000 replicates, resampling 97 IF tasks, 99 MTIF tasks, and 100 tasks within each of ST and MT, with all runs and model--harness outcomes of a sampled task kept together.

\subsection{Main results}
\label{sec:main-results}

\begin{figure}[t]
\centering
\includegraphics[width=\linewidth]{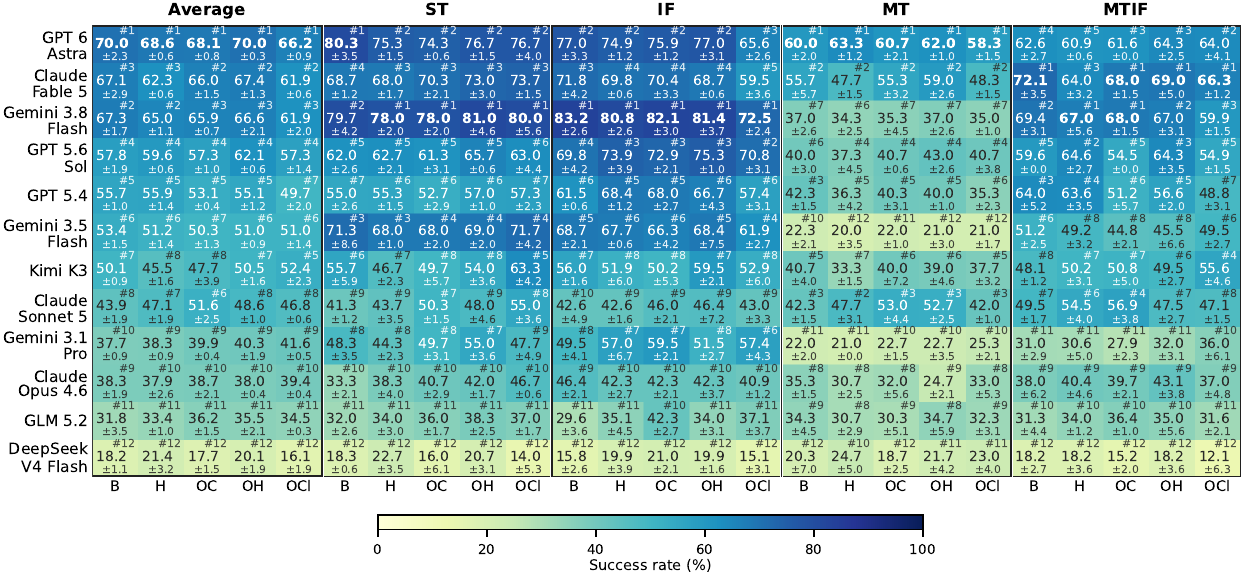}
\caption{Success rates by model, harness, and task setting. Each cell gives the three-run mean, within-column rank, and across-run SD in percentage points; rank-1 means are bold. Average weights the four task settings equally. B = Bare, H = Hermes, OC = OpenCode, OH = OpenHands, OCl = OpenClaw. Task-sampling confidence intervals appear in Appendix Figure~\ref{fig:main-results-ci}.}
\label{fig:main-results}
\end{figure}

Averaged across settings (Figure~\ref{fig:main-results}), GPT~6 Astra performs best under the bare harness and when the four end-user harnesses are averaged equally, at 70.0\% and 68.2\%. The highest individual model--harness mean is Astra with OpenHands at 70.0\%. The top-performing model varies across task settings. Gemini~3.8 Flash leads IF under all five harnesses; Astra and Fable have the highest harness-averaged MT and MTIF means, respectively. Open models show a large room for improvement; even Kimi K3, the strongest evaluated open-weight model, has substantially lower success rates than GPT~6, with the best score of 52.4\% under OpenClaw (Figure~\ref{fig:main-results}). Aggregate performance is broadly repeatable across the three operational runs, while individual outcomes vary (\S\ref{sec:repeatability-details}). The mean success rate is 49.2\%, compared with 32.9\% for success in all three runs (Appendix Table~\ref{tab:pass-consistency}).

Model rankings are largely preserved across harnesses. After averaging each model's success over the four settings, we compare every pair of the twelve models (66 pairs) under the bare harness and each end-user harness. The bare-harness ordering is preserved in 251/264 comparisons, and no model moves more than two places in the twelve-model ranking. Rankings vary more within individual settings, where models can move up to four places (\S\ref{sec:harness-results}).

\paragraph{Failure analysis}
Most failures in ST and MT involve an incorrect final service state, as judged by the assertion code or DB-diff verifier (95.0\% and 92.9\% of all failures, respectively). Attempts ending because of call or token limits, incorrect refusals, or malformed responses are classified separately as execution-completion failures (2.3\% of all failures) (\S\ref{sec:failure-details}).

\subsection{Steering default behavior under instructions}
\label{sec:if-results}

\begin{figure}[t]
\centering
\includegraphics[width=0.90\linewidth]{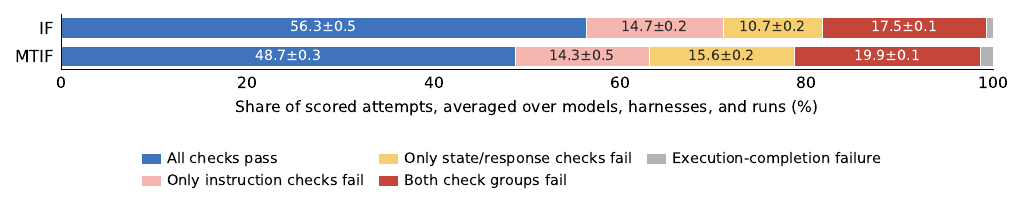}
\caption{Rates of the combined outcomes of the state, response, and instruction verifiers in IF and MTIF, averaged over all models, the five harnesses, and three runs.}
\label{fig:task-steering-outcomes}
\end{figure}

IF and MTIF were constructed by adding behavioral instructions to tasks that a reference model could otherwise solve. Many observed failures reflect this intended challenge: 30.6\% of failed attempts pass every state and final-response check but violate an instruction (Figure~\ref{fig:task-steering-outcomes}).

\begin{figure}[t]
\centering
\includegraphics[width=\linewidth]{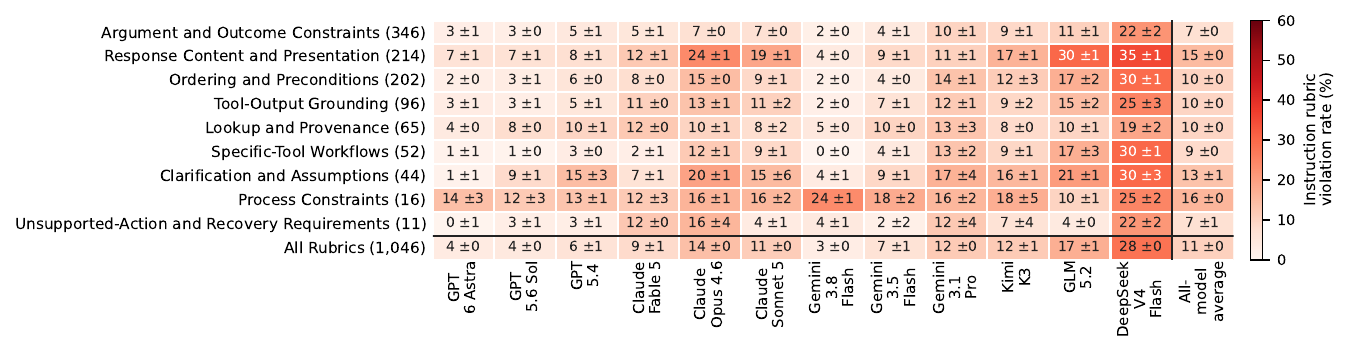}
\caption{Violation rates of instruction rubrics in IF and MTIF tasks, averaged over the five harnesses: three-run means $\pm$ across-run SD (percentage points), computed after averaging harnesses within each run. Each model column averages IF and MTIF equally. Parentheses show the number of distinct rubrics, not the number of tasks. Appendix Table~\ref{tab:if-groups} describes the rubric categories.}
\label{fig:instruction-types-by-model}
\end{figure}

\paragraph{Different models violate different kinds of instructions.}
Figure~\ref{fig:instruction-types-by-model} compares models' violation rates across instruction categories (\S\ref{sec:if-types}), averaged equally across all five harnesses. Averaged across models, instruction rubric violation rates are highest for instructions about what to say and how to format it (response content and presentation), when to ask for clarification or proceed without asking (clarification and assumptions), and which procedures to follow, such as avoiding redundant tool calls (process constraints).\footnote{Appendix Figure~\ref{fig:instruction-types-bare} shows the bare-harness results. The overall trend is similar to that of Figure~\ref{fig:instruction-types-by-model}, which means that harness-specific instructions do not change the pattern. The same three categories have the highest average violation rates.} The three GPT models are among the four most compliant overall. Models differ across categories: GPT~6 Astra has the lowest rate for clarification and assumptions (1.1\%), GLM~5.2 for process constraints (10.0\%), and Gemini~3.8 Flash for ordering (1.8\%) and response requirements (3.8\%). DeepSeek V4 Flash has the highest violation rate in all nine categories.

\begin{figure}[t]
\centering
\includegraphics[width=0.95\linewidth]{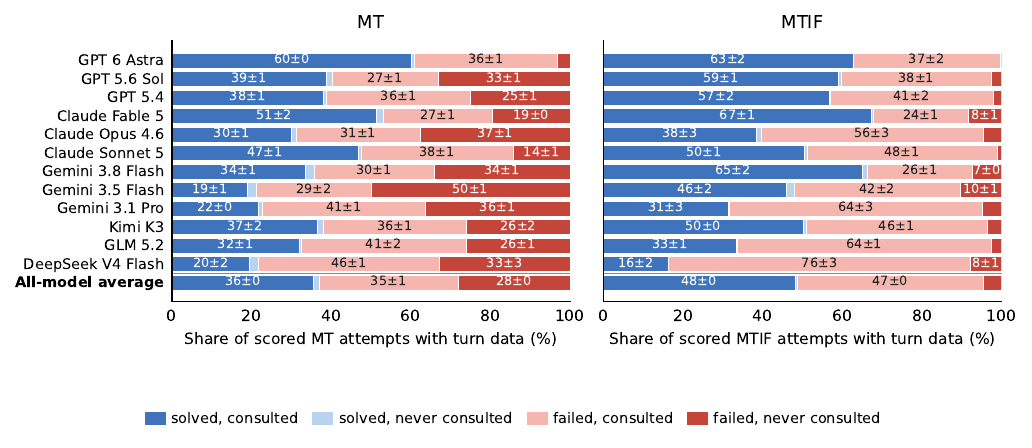}
\caption{Success and user consultation rates in MT and MTIF by model. Consultation denotes cases where at least one user reply exists other than the simulator's termination signal. Each model--type bar averages over the five harnesses and three runs.}
\label{fig:consult-outcome}
\end{figure}

\paragraph{Agents often fail to obtain missing information from the user.}
MT tasks are designed to require information available only from the user through clarification or confirmation. The MT prompt tells agents to search the apps for context and ask for information absent from them. Nevertheless, 29.5\% of MT attempts end after a single agent turn without a user reply (Figure~\ref{fig:consult-outcome}; percentage of ``solved, never consulted'' and ``failed, never consulted''). Only 4.7\% of these succeed,\footnote{Among the 18,000 MT attempts, 248 are scored as successful without consultation. These usually involve a default choice matching the annotated answer or workspace evidence suggesting it (\S\ref{sec:no-consultation-successes}).} compared with 50.5\% after a reply. Consultation is more common in MTIF, and this may reflect its construction: annotators started from a multi-turn task that the reference model could solve, then added behavioral instructions until it failed (\S\ref{sec:task-construction}).

\subsection{Multi-turn clarification}
\label{sec:mt-results}

\paragraph{Consultation differs substantially across models.}
In MT, consultation ranges from 47.7\% (Gemini~3.5 Flash) to 96.2\% (Astra). Consultation does not always mean that the agent asked a relevant question, though: in a qualitative review, 22/30 Claude Fable~5 first messages requested the withheld information, and the rest did not, e.g., reporting completion based on a guessed value (\S\ref{sec:first-turn-consultation}).

\subsection{Performance across harnesses}
\label{sec:harness-results}

\begin{figure}[t]
\centering
\includegraphics[width=0.9\linewidth]{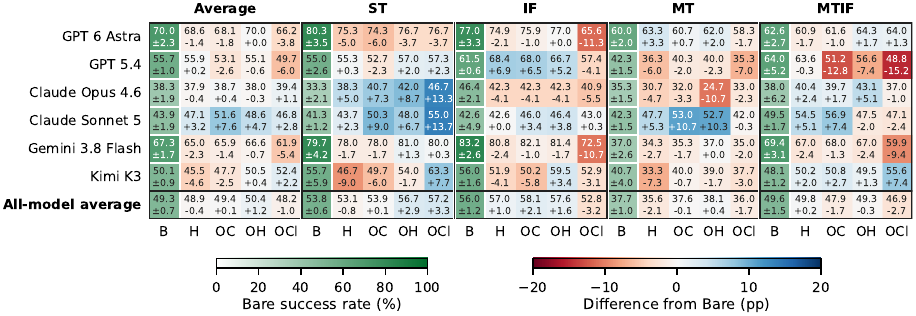}
\caption{Success rates for six models (all models: Appendix Figure~\ref{fig:harness-model-matrices-full}). Labels show three-run means $\pm$ across-run SDs. Green: absolute bare-harness success. Other cells: absolute success (above) and difference from the bare harness (below, percentage points; blue: gain, red: loss). The final row weights all twelve models equally. Comparisons are exploratory; uncertainty and repeatability are discussed in \S\ref{sec:harness-benefit-details} and \S\ref{sec:repeatability-details}. Differences are computed from unrounded means. B = Bare, H = Hermes, OC = OpenCode, OH = OpenHands, OCl = OpenClaw.}
\label{fig:harness-model-matrices}
\end{figure}

\paragraph{End-user harness benefits depend on the model and setting.}
Figure~\ref{fig:harness-model-matrices} shows each end-user harness's change in success relative to the bare harness. Averaging equally across models, settings, and end-user harnesses gives a mean difference of -0.05 percentage points. Harness--setting means range from -3.0 points for OpenClaw on IF to +3.3 for OpenClaw on ST. However, specific model--harness combinations show large gains and losses relative to the bare harness within individual task settings (\S\ref{sec:harness-extreme-cases}). OpenClaw's ST gain for Claude Sonnet~5 appears in all three runs, although individual task outcomes remain variable (\S\ref{sec:repeatability-details}). Lower bare-harness success is weakly associated with greater harness benefit, but does not reliably identify which configuration benefits (\S\ref{sec:harness-benefit-details}).

\paragraph{Task outcomes vary within as well as across harnesses.}
For the same model, task, and run, attempts under the bare harness and an end-user harness have different outcomes in 24.7\% of comparisons. Comparing different runs of the same model, harness, and task gives 21.7\%. Cross-harness disagreement thus exceeds rerun disagreement by 3.0 percentage points, but substantial variation occurs even without changing the harness. Therefore, these reversals alone do not establish that particular harnesses reliably solve distinct sets of tasks. The rerun comparison and additional outcome summaries are given in Appendix Table~\ref{tab:rerun-vs-harness} and Appendix Figure~\ref{fig:configuration-specific-outcomes}.

\paragraph{OpenClaw shows distinctive behaviors in our experimental setup.} OpenClaw's tool-call distribution has a higher upper quartile, particularly on MT (Appendix Figure~\ref{fig:harness-tool-call-quartiles}). OpenClaw's search guidance and shell-based tool interface may contribute to these higher call counts, although their contributions cannot be separated in this comparison (\S\ref{sec:harness-config}).

\paragraph{Comparison of tasks solved only by Bare or OpenClaw} In our qualitative review of Claude Sonnet~5's ST trajectories, OpenClaw-only successes often involved finding or reconciling evidence from a wide range of services that Bare missed, whereas Bare-only successes occurred when OpenClaw completed much of the workflow but made a narrow execution or final-response error (\S\ref{sec:harness-extreme-cases}). On one MT task, GPT~5.6 Sol under OpenClaw made 151--159 tool calls across the runs searching the services for information that only the user could provide, whereas under Bare it asked the user and obtained it.

\section{Conclusion}
\label{sec:conclusion}

\MV{} evaluates cross-service delegation under behavioral instructions, user clarification, and multiple harnesses. The quality of the benchmark is supported by multiple rounds of task and service quality control, human audits of the simulator and verifiers, and assessment using the Agent Benchmark Checklist. Across the twelve models, the highest model--harness mean success rate is 70.0\%, and the highest for an open-weight model is 52.4\%. Agents frequently fail to obtain needed user information, and their instruction-following strengths differ across constraints. Harnesses achieve similar success rates when averaged across all twelve models and task settings, but changing the harness can substantially improve or reduce success for a particular model and setting. \MV{} provides a trustworthy testbed for improving agents' tool-use behavior in real-world task delegation under various harnesses.

\section*{Acknowledgments}
We thank Slav Petrov for supporting this work and Charles Kwong for his constructive feedback on the paper. We also thank Pillar AI, Turing, and Mercor.

\bibliography{references}
\bibliographystyle{abbrvnat}

\clearpage
\appendix

\section{Limitations}
\label{sec:limitations}

\paragraph{Fidelity of simulated services} \MV{} simulates documented API behavior rather than calling live services, trading undocumented service-specific business logic and evolving service semantics for reproducibility; live services change between runs and would make scores incomparable over time (\S\ref{sec:dimensions}). Coverage is partial and disclosed per service: Table~\ref{tab:service-grounding} lists the reference used for each implementation, its exposed tools, and its omissions or interface adaptations, such as authentication, webhooks, media transport, quotas, and some request types of large APIs (the simulated Google Docs supports 5 of the 40 \texttt{batchUpdate} request types). Tasks were created on the implemented operations, so no task requires an omitted one, and scores measure tool use against the documented interfaces that agents are expected to rely on; the implementations also pass their unit tests. The real-world tools criterion in Table~\ref{tab:comparison-compact} is applied to every benchmark on the basis of its own description of its tools, and the per-service disclosure goes beyond that criterion. Except for Google Calendar, whose behavior was matched to captures from live accounts (\S\ref{sec:env-gen}), we did not compare the services end to end with the real APIs. Reusing the tasks' exact identifiers is not always possible: an email address, for example, may already belong to someone else. Even if we replace those identifiers with ones reserved for testing, we still need to create the corresponding accounts, permissions, and linked records across services, then restore their initial states for repeated runs. This setup is costly, and account-verification requirements and anti-abuse controls can restrict automated account creation and testing.

\paragraph{Tool defects} Like any large software system, the simulated services may retain minor defects despite several rounds of QC (\S\ref{sec:qc}). These defects may require agents to work around tool errors or unexpected outputs, but every task remains solvable: we verify at least one working solution path by replaying its gold trajectory and requiring all applicable verifiers to pass. In addition, all but two validation tasks are solved by at least one agent attempt in our experiments, providing further evidence of task solvability. Errors returned by the tools are therefore part of the task, and agents must recover from them; infrastructure failures, in contrast, are never scored, and the affected attempts are rerun (\S\ref{sec:harness-config}). By default, the simulation does not reproduce the outages, quota errors, and transport failures of live services, so the reported evaluation does not cover recovery from such backend failures. However, the services include an error simulator that injects errors with a set probability. We removed this option from the evaluation code and did not test it, because such an experiment had low priority and the injected errors would add noise to agent scores.

\paragraph{Narrow scope of the user simulator} The user simulator's reticent responses may not capture the full diversity of how people supply missing information. Therefore, the resulting scores measure clarification under this controlled interaction, not the full range of how people collaborate with agents. However, this restricted role is deliberate: grounding replies in task-specific hidden information leaves little room for hallucination and unsolicited hints, and makes the simulator's behavior checkable. This design lets us assess whether agents recognize and request information they cannot retrieve from services, without relying on an open-ended simulation of human behavior that can introduce high variance into the evaluation.

\paragraph{Assertion code strictness} Assertion code is a guardrail that checks whether an attempt meets the most obvious requirements of a task. However, if an annotator overlooks alternative ways to solve a task, the assertion code can be overly strict and reject valid solutions by requiring a value or execution logic that only one solution path needs. For example, the original assertion code for one validation task accepted an email containing ``7:30'' but not one containing ``19:30'' and would therefore have failed an agent that wrote the time as 19:30; we fixed the code to accept both.\footnote{Accepting ``7:30'' also passes an agent that writes 7:30~AM. Assertion code checks only the most obvious requirements, so it is not designed to keep false positives low; the DB-diff verifier is meant to catch them.} Such strictness lowers the success rate rather than rewarding incorrect outcomes. We identified such cases through disagreements between the assertion code and the DB-diff verifier among the reviewed attempts, reviewed them against the task and the recorded states, relaxed the assertion code, and then replayed the gold solutions and rechecked the recorded agent states (\S\ref{sec:qc}). This process reduced the identified false negatives, but assertions may still reject valid solutions.

\paragraph{Verdict-assisted human review} In the judge audit, reviewers saw the sampled verifier verdict and its explanation, and the second reviewer saw the first reviewer's assessment. This information may influence their judgments, so the reported agreement should not be interpreted as accuracy against an independently labeled reference. We chose this design because a trajectory, its database diff, and the final response form long structured records that are slow and error-prone to judge from scratch, whereas checking a stated verdict against the cited evidence directs reviewers to the decisive facts. Reviewers overturned 12 of the 400 sampled verdicts, each checked by an L3 reviewer (\S\ref{sec:qc}). We do not report inter-annotator agreement.

\paragraph{Choice of judge and user simulator models} The LLM judges and the user simulator are Gemini models, as are three of the evaluated models, so self-preference, the tendency of LLM evaluators to favor their own generations~\citep{panickssery2024llm,zheng2023judging}, could inflate the scores of the Gemini models. However, the user simulator has little room for self-preference, because it answers only from each task's hidden information (\S\ref{sec:usersim}). An early pilot that used a Claude model and Gemini 3.1 Pro as the evaluated agent or the judge found no statistically significant difference in user-simulator accuracy or agent success rates. A judge also cannot turn a failed assertion into a success, because assertion code checks every task and success requires every applicable verifier to pass. Most importantly, in the human audit of the judges (\S\ref{sec:qc}), agreement with human reviewers is similar for the two audited Gemini models (Gemini~3.1 Pro and 3.5 Flash) and the other eight models (DB-diff 39/40 vs.\ 156/160; final response 19/20 vs.\ 73/80; instruction following 20/20 vs.\ 80/80), and the judges pass Gemini trajectories that reviewers judged unsuccessful no more often than those of other models (DB-diff 1/19 vs.\ 4/67; final response 1/9 vs.\ 5/29).\footnote{Because the audit sampled one trajectory per model, harness, and setting, agreement could be compared by the model that produced the trajectory, although Gemini 3.8 Flash and GPT 6 Astra were not included since the audit was performed before they were evaluated. Counts use the three-vote majority.} The audit therefore shows no sign that the judges favor Gemini models, although a sample of this size can detect only a large bias.

\paragraph{Restricted end-user harness capabilities} Under every end-user harness, the agent's only tools are \MV{}'s actions; the harness-native capabilities that these harnesses expose in deployment, such as running terminal commands or Python code, exploring a file system, or accessing the web, are removed, or confined to a separate sandbox where removal is impossible (\S\ref{sec:harness-config}). Our results therefore do not measure the full capability of an end-user harness: an agent that could write and run a Python script, or look something up on the web, might complete delegation tasks under an unconstrained end-user harness that it fails here. We restricted these capabilities for two reasons. First, when they were available, some models kept calling native tools instead of the benchmark's tools, for example running SQL queries to look for real emails, and wasted their model-call budget on them. Second, we consider the restriction unavoidable for security reasons. With native capabilities enabled, an agent could reach the tasks and gold solutions in the file system or their publicly released copies on the web, alter the evaluation environment, or send data outside the environment, and its tool calls would no longer be recorded through the benchmark's interface, so the resulting scores would be neither safe to collect nor attributable to the model's use of the simulated services. Before these capabilities were fully removed, we observed one incident in which an agent under Hermes uploaded an image to a temporary image-sharing website to solve a task, and the verifiers scored the attempt as successful; we then hardened the restrictions. Measuring what native capabilities add would require a secure environment and is left to future work.

\paragraph{End-user harness modifications} The harnesses we evaluate are not exactly the same as what users get when they install them with their default settings. We modify them to compare models on the same tasks, with the same service tools and model-call budget. We also repair integration defects that prevent some models from running successfully under an unpatched harness. Leaving those defects uncorrected would mix failures of task execution with failures of model--harness compatibility. Also, during development, frequent upstream harness updates caused evaluation code that had worked the previous day to raise runtime errors the next day; pinning harness versions was therefore necessary both to run the experiments reliably and to make them reproducible. We release the pinned configurations and patches with the evaluation code, allowing others to reproduce the setup and reuse the compatibility fixes in their own deployments.

\paragraph{Task selection using reference models} The choice of reference models may influence relative model rankings and measured harness benefits. These comparisons may differ on tasks constructed without the reference-model failure criterion (\S\ref{sec:task-construction}). We evaluate the reference models alongside newer models to measure whether the difficulties identified during construction persist and how other models perform on the same challenges.

\section{Statements}

\subsection{AI use statement}
We used generative AI tools during the environment and task construction to generate the blueprints of the initial service states and the initial drafts of the tool implementations, which annotators and engineers then revised, and to help implement the codebase, clean and reformat the data, analyze qualitative data, write the analysis code, and interpret the results. All comparisons and analyses in this paper were proposed by the authors. The initial service state of each task was generated from an LLM-written blueprint (\S\ref{sec:workspace-gen}), and gold trajectories were recorded from agent runs under annotator guidance (\S\ref{sec:task-construction}). The simulated services and the evaluation and analysis code were written with the help of coding agents (\S\ref{sec:tools}). Several qualitative reviews that human reviewers or the authors performed were assisted by LLMs, as stated where they are reported (\S\ref{sec:qc}, \S\ref{sec:first-turn-consultation}, \S\ref{sec:no-consultation-successes}). The benchmark also uses LLMs as its user simulator and judges (\S\ref{sec:usersim}, \S\ref{sec:verifier}). The authors collected all related work and filled in the benchmark comparison and Agent Benchmark Checklist tables by hand. LLMs filled in the entries that extended these tables, and the authors verified them by hand. At least two authors filled in and/or verified each entry separately and resolved any disagreements through multiple meetings held to discuss the entries. We also used generative AI tools to write the code that creates the figures and to help draft and edit parts of the paper.

We confirm that all AI-assisted work was reviewed and verified by humans. Annotators inspected and corrected every generated initial state and guided the LLMs to generate every gold solution, and every gold solution passes all applicable verifiers on replay. The service implementations pass their unit tests, the user simulator and judges were validated against human review (\S\ref{sec:qc}), and the analysis code was tested and reported numbers were checked multiple times by the authors. The authors reviewed and revised all text manually. We take responsibility for the final content of this work, including text, claims, or artifacts produced with the aid of generative AI.

\subsection{Ethics statement}
All tasks, initial service states, tools, and reviews were produced by professional annotators and engineers engaged through three commercial vendors under paid contracts (\S\ref{sec:qc}). No personal data were collected: the initial service states were generated from a synthetic blueprint, and the simulated services hold no real user data.

\subsection{Reproducibility statement}
We release the 396 validation tasks, the simulated services, the evaluation code, and the harness configurations; the 400 test tasks remain hidden for the leaderboard. The environment is self-contained: each task starts from a fixed initial state, and no tool calls a live service (\S\ref{sec:tools}), so the same sequence of tool calls always produces the same service outcome. The evaluation code pins every Python dependency in a uv lock file. We describe the services, tasks, user simulator, and verifiers in \S\ref{sec:benchmark} and \S\ref{sec:construction}, including the grounding of each service (Table~\ref{tab:service-grounding}), initial-state generation (\S\ref{sec:workspace-gen}), task construction (\S\ref{sec:task-construction}), and the inputs of each verifier (Table~\ref{tab:verifier-inputs}). The evaluation protocol, model identifiers, access routes, and pinned harness versions are given in \S\ref{sec:setup} and \S\ref{sec:harness-config} (Table~\ref{tab:identifiers}). The general instructions and the user-simulator and judge prompts are reproduced verbatim in \S\ref{sec:prompts}, and two worked tasks with recorded attempts are listed in \S\ref{sec:worked-examples}. Because the evaluated models, the user simulator, and the judges are LLMs whose outputs vary between calls, reported success rates average three complete runs, with across-run standard deviations and task-bootstrap confidence intervals (\S\ref{sec:setup}).

\section{Extended background and benchmark comparison}
\label{sec:background-details}

\subsection{Full definitions of the design dimensions}
\label{sec:dimensions-details}

Figure~\ref{fig:design-dimensions} summarizes the eleven design dimensions in two groups. Their full definitions are as follows.

\subsubsection*{Realism}
\begin{itemize}
\item \textbf{Real-world tools} Tools should reproduce real-service interfaces from public API or implementation documentation, preserving their operations, argument schemas, and documented behaviors, with any omissions stated. Benchmark-defined backends with reduced interfaces understate the difficulty of real tool use, where arguments are nested and constraints are service-specific.
\item \textbf{Broad and implicit multi-service scope} There should be enough services that selecting the right few among plausible distractors is itself part of the task. A query that explicitly names all the services to be used, or a harness that implicitly supplies only the relevant tools for each task, solves the selection problem for the agent.
\item \textbf{Interactive multi-turn user simulator} In real-world task delegation, users interact with the agent, so evaluation must include multi-turn tasks. A scripted user cannot adapt its replies to clarification questions outside the script.
\item \textbf{Category-wise steering evaluation} A model can complete a task while violating instructions about how to act or respond. Evaluate adherence to developer or user instructions separately from task completion, using identifiable checks for at least two distinct behavioral categories and reporting quantitative adherence rates for each, with the evaluated population stated.
\item \textbf{Cross-harness evaluation} Real-world task delegation mostly happens through end-user harnesses, which change both the model's behavior and the user's experience. Attributing a score to the model or to the harness requires evaluating every model on the same tasks under the same bare harness and end-user harnesses, with model settings matched as closely as practical and any remaining differences documented.
\end{itemize}

\subsubsection*{Trustworthy evaluation}
\begin{itemize}
\item \textbf{Hidden test set} Released tasks can enter training data, so a held-out set evaluated only through model submission limits contamination.
\item \textbf{Verification through service state} Judge correctness by the resulting service state rather than by matching a prescribed tool-call sequence. This credits alternative valid solutions and recovery from intermediate mistakes, while checking that the requested changes were achieved and unrelated state was preserved.
\item \textbf{Self-contained environment} Task services and required task content should be fixed and resettable, so upstream changes cannot silently alter the task or its outcome. We assess task-time dependencies, not ordinary installation dependencies or hosted agent, simulator, and judge model calls; the latter still require separate control and error handling.
\item \textbf{Infrastructure-error isolation} Attempts affected by provider rate limits, initialization failures, or exhausted quotas must be classified separately from agent failures, excluded from reported denominators, and rerun; otherwise operational accidents are recorded as spurious agent failures.
\item \textbf{Validated user simulator and verifiers} The accuracy of every component that produces or influences a score must be measured against human judgment, deterministic checks included. Assertion code that is stricter than the task, or an exact match after normalization, rejects correct solutions just as an LLM judge can, and a user simulator that reveals or withholds the wrong information changes what the agent is scored on.
\item \textbf{Right-sized task set} A task set should be large enough to distinguish model performance reliably, but small enough that every task can undergo careful QC and the evaluation can be run repeatedly.
\end{itemize}

\subsection{Full benchmark comparison}
\label{sec:full-comparison}

Table~\ref{tab:comparison} assesses 33 other benchmarks alongside \MV{} against our design dimensions (\S\ref{sec:dimensions}). It extends Table~\ref{tab:comparison-compact} with additional benchmarks, grouped by whether and how they provide a fixed set of tools. Related instruction-following and interaction studies are discussed in \S\ref{sec:full-audit}. The bucket thresholds and column definitions are given below the table.

\begin{table*}[p]
\centering
\footnotesize
\setlength{\tabcolsep}{2.8pt}
\renewcommand{\arraystretch}{0.72}
\caption{Comparison of 33 other benchmarks and \MV{} along our benchmark design dimensions (\S\ref{sec:dimensions}). \cmark~= satisfied, \pmark~= partially satisfied or qualified, \xmark~= not satisfied, --~= not applicable, not reported, or exposed only through run-time retrieval. Numeric columns give paper-reported counts or counts derived from a specified release, with differences noted; ``\#~Tools/task'' and ``\#~Services/task'' report the action space the agent is exposed to within a single task, as the median across tasks where a distribution is available and otherwise as the reported mean or range. The notes below define the \#~Tasks column. Superscripts refer to the accompanying notes. A compact view of the twelve closest and most widely used benchmarks appears in Table~\ref{tab:comparison-compact}.}
\label{tab:comparison}
\begin{tabular}{@{}lcrrrrcccccccccr@{}}
\toprule
& \multicolumn{8}{c}{\textbf{Realism}} & \multicolumn{6}{c}{\textbf{Trustworthy evaluation}} & \\
\cmidrule(lr){2-9} \cmidrule(lr){10-15}
\textbf{Benchmark}
& \rot{Real-world tools}
& \rot{\# Services}
& \rot{\# Tools}
& \rot{\# Tools/task}
& \rot{\# Services/task}
& \rot{User simulator}
& \rot{Category-wise steering}
& \rot{Cross-harness evaluation\textsuperscript{a}}
& \rot{Hidden test set}
& \rot{Service-state verification}
& \rot{Self-contained}
& \rot{Validated user simulator\textsuperscript{b}}
& \rot{Validated evaluation\textsuperscript{c}}
& \rot{Infra-error isolation\textsuperscript{d}}
& \rot{\# Tasks\textsuperscript{$\dagger$}} \\
\midrule
\multicolumn{16}{@{}l}{\emph{No tools provided (agent brings its own)}} \\[1pt]
BrowseComp~\citeyearpar{wei2025browsecomp} & --\textsuperscript{e} & -- & -- & -- & -- & \xmark & \xmark & \xmark & \xmark & \xmark & \xmark & -- & \xmark & -- & 1{,}266 \\
GAIA~\citeyearpar{mialon2024gaia} & --\textsuperscript{e} & -- & -- & -- & -- & \xmark & \xmark & \pmark & \pmark\textsuperscript{f} & \xmark & \xmark & -- & \xmark & -- & 466 \\
GDPval~\citeyearpar{patwardhan2025gdpval} & --\textsuperscript{e} & -- & -- & -- & -- & \xmark & \xmark & \xmark & \cmark & \xmark & \xmark & -- & \cmark & -- & 1{,}320 \\
JobBench~\citeyearpar{li2026jobbench} & --\textsuperscript{e} & -- & -- & -- & -- & \xmark & \pmark\textsuperscript{u} & \pmark & \xmark & \cmark\textsuperscript{s} & \xmark & -- & \pmark\textsuperscript{i} & \pmark & 130\textsuperscript{s} \\
\addlinespace[3pt]
\multicolumn{16}{@{}l}{\emph{Computer-use, coding, and web benchmarks}} \\[1pt]
APEX-Agents~\citeyearpar{vidgen2026apex} & \pmark & 12 & 250 & -- & -- & \xmark & \xmark & \xmark & \xmark & \xmark\textsuperscript{g} & \cmark & -- & \cmark & \pmark & 480 \\
MINT~\citeyearpar{wang2024mint} & \pmark & 1 & --\textsuperscript{h} & -- & 1 & \cmark & \pmark\textsuperscript{u} & \xmark & \xmark & \cmark & \xmark & \pmark & \xmark & \pmark & 586 \\
WebArena~\citeyearpar{zhou2024webarena} & \pmark & 8 & --\textsuperscript{h} & -- & 8 & \xmark & \pmark\textsuperscript{u} & \xmark & \xmark & \cmark & \cmark & -- & \cmark & \pmark & 812 \\
AgentBench~\citeyearpar{liu2024agentbench} & \pmark & 8 & --\textsuperscript{h} & -- & -- & \xmark & \xmark & \xmark & \xmark & \cmark & \cmark & -- & \pmark & \pmark & 1{,}014 \\
Terminal-Bench~\citeyearpar{merrill2026terminalbench} & \cmark & -- & --\textsuperscript{h} & -- & -- & \xmark & \pmark\textsuperscript{u} & \pmark & \xmark & \cmark & \xmark & -- & \pmark\textsuperscript{i} & \pmark & 89 \\
TheAgentCompany~\citeyearpar{xu2025theagentcompany} & \cmark & 4 & --\textsuperscript{h} & -- & 4 & \cmark & \pmark\textsuperscript{u} & \pmark & \xmark & \cmark & \xmark & \cmark & \pmark\textsuperscript{i} & \xmark & 175 \\
OSWorld-Verified~\citeyearpar{xie2024osworld,osworldverified2025} & \cmark & -- & --\textsuperscript{h} & -- & -- & \xmark & \pmark\textsuperscript{u} & \pmark & \xmark & \cmark & \xmark & -- & \pmark\textsuperscript{i} & \pmark & 369 \\
SWE-Bench Pro~\citeyearpar{deng2025swebench} & \cmark & -- & --\textsuperscript{h} & -- & -- & \xmark & \xmark & \pmark\textsuperscript{$\ddagger$} & \cmark & \cmark & \cmark & -- & \pmark\textsuperscript{i} & \xmark & 1{,}865 \\
Agents' Last Exam~\citeyearpar{sun2026agents} & \cmark & -- & --\textsuperscript{h} & -- & -- & \xmark & \pmark\textsuperscript{u} & \pmark & \cmark & \cmark & \xmark & -- & \pmark\textsuperscript{i} & \cmark & 1{,}490 \\
\addlinespace[3pt]
\multicolumn{16}{@{}l}{\emph{Live real-world services}} \\[1pt]
MCP-Universe~\citeyearpar{luo2026mcpuniverse} & \cmark & 11 & 133 & -- & -- & \xmark & \pmark & \pmark & \xmark & \cmark\textsuperscript{l} & \xmark & -- & \pmark\textsuperscript{i} & \xmark & 231 \\
ToolBench~\citeyearpar{qin2024toolllm} & \cmark & 3{,}451 & 16{,}464 & --\textsuperscript{m} & -- & \xmark & \xmark & \cmark\textsuperscript{$\ddagger$} & \xmark & \xmark & \xmark & -- & \cmark & \pmark & 1{,}200 \\
LiveMCPBench~\citeyearpar{mo2026livemcpbench} & \cmark & 70 & 527 & --\textsuperscript{m} & -- & \xmark & \xmark & \xmark & \xmark & \xmark & \xmark\textsuperscript{n} & -- & \cmark & \pmark & 95 \\
Toolathlon~\citeyearpar{li2026tool} & \cmark & 32 & 604 & 69.9 & 3.4 & \xmark & \pmark\textsuperscript{u} & \xmark & \xmark & \cmark & \xmark\textsuperscript{o} & -- & \pmark\textsuperscript{i} & \pmark & 108 \\
MCP-Atlas~\citeyearpar{bandi2026mcpatlas} & \cmark & 36 & 220 & 15.2 & -- & \xmark & \xmark & \xmark & \cmark & \xmark\textsuperscript{p} & \xmark & -- & \pmark\textsuperscript{i} & \pmark & 1{,}000 \\
\addlinespace[3pt]
\multicolumn{16}{@{}l}{\emph{Simulated services}} \\[1pt]
BFCL v4~\citeyearpar{patil2025berkeley} & \pmark & -- & 3{,}417\textsuperscript{q} & 3\textsuperscript{q} & -- & \xmark & \pmark\textsuperscript{u} & \xmark & \xmark & \cmark & \xmark & -- & \xmark & \xmark & 5{,}090 \\
API-Bank~\citeyearpar{li2023apibank} & \pmark & -- & 73 & --\textsuperscript{m} & -- & \xmark & \xmark & \xmark & \xmark & \xmark & \cmark & -- & \xmark & \xmark & 314 \\
BFCL v3~\citeyearpar{patil2025berkeley} & \xmark & 8 & 128 & -- & -- & \xmark & \pmark\textsuperscript{u} & \xmark & \xmark & \cmark & \cmark & -- & \xmark & \xmark & 1{,}000 \\
UltraTool~\citeyearpar{huang2024planning} & \xmark & -- & 2{,}032 & -- & -- & \xmark & \pmark\textsuperscript{u} & \xmark & \xmark & \xmark & \cmark & -- & \cmark & -- & 5{,}824 \\
CCTU~\citeyearpar{ye2026cctubenchmarktooluse} & \xmark & -- & -- & 9.3 & -- & \xmark & \cmark & \xmark & \xmark & \xmark\textsuperscript{r} & \cmark & -- & \pmark\textsuperscript{i} & \pmark & 200 \\
WorkBench~\citeyearpar{styles2024workbench} & \xmark & 5 & 26 & 26 & 5 & \xmark & \pmark\textsuperscript{u} & \xmark & \xmark & \cmark & \cmark & -- & \xmark & \pmark & 690 \\
$\tau$-bench~\citeyearpar{yao2025taubench} & \xmark & 2 & 28 & 13--15 & 1 & \cmark & \pmark\textsuperscript{u} & \pmark\textsuperscript{$\ddagger$} & \xmark & \cmark & \cmark & \xmark & \xmark & \xmark & 165 \\
ToolSandbox~\citeyearpar{lu2025toolsandbox} & \pmark & 11 & 34 & 7 & -- & \cmark & \pmark\textsuperscript{u} & \xmark & \xmark & \pmark & \xmark & \cmark & \pmark\textsuperscript{i} & \xmark & 1{,}032 \\
AppWorld~\citeyearpar{trivedi2024appworld} & \cmark & 9 & 457 & --\textsuperscript{m} & 9 & \xmark & \pmark\textsuperscript{u} & \pmark\textsuperscript{$\ddagger$} & \pmark\textsuperscript{z} & \cmark & \cmark & -- & \pmark & \xmark & 750 \\
CRMArena-Pro~\citeyearpar{huang2026crmarenapro} & \cmark & 1 & 2 & 2 & 1 & \cmark & \pmark\textsuperscript{w} & \xmark & \xmark & \xmark & \xmark & \cmark & \xmark & \xmark & 4{,}280 \\
$\tau^2$-bench~\citeyearpar{barres2025tau2bench} & \xmark & 3 & 42\textsuperscript{t} & 14\textsuperscript{t} & 1 & \cmark & \pmark\textsuperscript{u} & \xmark & \xmark & \cmark & \cmark & \cmark & \xmark & \pmark & 279 \\
$\tau^3$-bench~\citeyearpar{sierra2026tau3} & \xmark & 4 & 105\textsuperscript{v} & 16\textsuperscript{v} & 1 & \cmark & \pmark\textsuperscript{u} & \pmark & \xmark & \cmark & \cmark & \cmark\textsuperscript{v} & \pmark\textsuperscript{i} & \pmark & 375\textsuperscript{v} \\
HANDBOOK.md~\citeyearpar{surge2026handbook} & \cmark & 5 & 82 & 82 & 5 & \xmark & \pmark\textsuperscript{u} & \xmark & \xmark & \cmark\textsuperscript{x} & \cmark & -- & \pmark\textsuperscript{i} & \cmark & 65 \\
AutomationBench~\citeyearpar{shepard2026automationbench} & \cmark & 47 & $\sim$500 & --\textsuperscript{m} & --\textsuperscript{m} & \xmark & \pmark\textsuperscript{u} & \xmark & \cmark & \cmark & \cmark & -- & \xmark & \pmark & 600 \\
GAIA2~\citeyearpar{froger2025are} & \xmark & 12 & 101 & 101 & 12 & \xmark & \pmark\textsuperscript{u} & \pmark\textsuperscript{$\ddagger$} & \cmark & \pmark & \cmark & -- & \cmark & \pmark & 1{,}120 \\
\midrule
\textbf{\MV{} (ours)} & \cmark & 35 & 809 & 148.5\textsuperscript{y} & 6 & \cmark & \cmark & \cmark & \cmark & \cmark & \cmark & \cmark & \cmark & \cmark & 796 \\
\bottomrule
\end{tabular}
\par\vspace{2pt}
\begin{minipage}{\textwidth}
\footnotesize
\textsuperscript{$\dagger$}A task set should be large enough to separate similarly performing models, yet small enough that every task can undergo manual QC and the full benchmark run with a reasonable budget (\S\ref{sec:dimensions}).
\end{minipage}
\end{table*}

\begin{table*}[p]
\ContinuedFloat
\caption*{\textbf{Notes for Table~\ref{tab:comparison}.}}
\footnotesize
\newcommand{\tnoteline}[2]{\par\noindent\hangindent=1.4em\hangafter=1 \textsuperscript{#1}\hspace{0.35em}#2}
\begin{flushleft}
\par\noindent\textbf{Real-world tools} \cmark: tools are the live services, or simulations implemented against a real service's documented interface (its public API specification, or its implementation documentation where no public API exists) that preserve its operations and argument schemas. \pmark: a mixture, or simulations that follow a real service's interface loosely. \xmark: benchmark-defined synthetic backends whose interfaces are reduced relative to the real service. \MV{}: service references, tool counts, and omissions or interface adaptations are documented individually in Table~\ref{tab:service-grounding}. Marks follow each benchmark's own description of its tools.
\par\noindent\textbf{Category-wise steering evaluation} Steering concerns explicit instructions about how an agent carries out a task or presents its response. \cmark: at least two distinct behavioral categories, identifiable adherence checks for each, and quantitative adherence results with the evaluated population stated for each category. \pmark: at least one instruction is evaluated, but the requirements for \cmark{} are not all met. \xmark: no such instruction is evaluated. The presence of a parser alone does not count. Explicit formatting instructions count when compliance is separately assessed or analyzed. Limits enforced by the environment do not count. Combining adherence with task completion in the headline score does not itself limit the mark. Benchmark-provided reference prompts count; third-party harness text does not.
\par\noindent\textbf{Service-state verification} \cmark: at least some tasks check the resulting service or environment state at completion. \pmark: checks intermediate state or matches executed write actions to an oracle without establishing that the required state holds at completion. \xmark: neither. Mixed task coverage does not reduce credit. This column records the verification mechanism, not its accuracy or whether it checks every unwanted change.
\par\noindent\textbf{Self-contained} \cmark: task services and required task content can be fixed and reset; \xmark: task-time dependencies on uncontrolled external services or content can change outcomes. Hosted model calls and ordinary installation dependencies do not by themselves disqualify a benchmark.
\par\noindent\textbf{AutomationBench population} The displayed task count covers 600 public tasks; its official leaderboard uses a separate unreleased set of 600+ private tasks.
\tnoteline{a}{\cmark: every evaluated model is run under every evaluated harness (full model$\times$harness grid); \pmark: multiple harnesses are evaluated, but the full model$\times$harness grid is not covered, including matched comparisons for selected models or unmatched submissions. A bare harness plus one end-user harness can receive partial credit. \xmark: no harness variation. A harness implements agent orchestration; browsing access, plugins, prompting, or sampling alone do not establish a distinct harness.}
\tnoteline{b}{\cmark: reports a numerical measure of simulator correctness or errors against human judgments. \pmark: reports human assessment that does not measure correctness, such as helpfulness or human-likeness; these assessments may also be quantitative. \xmark: no human assessment is reported. --: no user simulator.}
\tnoteline{c}{\cmark: reports a numerical measure of evaluator correctness against human judgments or independently established reference labels, such as accuracy, agreement, precision/recall, or correlation with human scores. \pmark: reports checks of evaluator correctness but no numerical result. \xmark: reports neither. Agreement between LLM judges alone does not establish correctness. This criterion applies to deterministic checks and exact-match scoring as well as LLM judges; it is never inapplicable.}
\tnoteline{d}{The evaluation harness classifies infrastructure failures (API, environment, quota) apart from agent errors, keeps them out of reported scores, and supports re-running errored tasks after the infrastructure is fixed. Errors returned by the simulated tools themselves, including residual defects, are part of the task and remain scored. \pmark: some but not all of these provisions, e.g., errors recorded but scored as agent failures, or crash reruns without error classification. --: no official rollout harness (or static evaluation), so infrastructure handling falls to the submitter.}
\end{flushleft}
\end{table*}

\begin{table*}[p]
\ContinuedFloat
\caption*{\textbf{Notes for Table~\ref{tab:comparison} (continued).}}
\footnotesize
\newcommand{\tnoteline}[2]{\par\noindent\hangindent=1.4em\hangafter=1 \textsuperscript{#1}\hspace{0.35em}#2}
\begin{flushleft}
\tnoteline{e}{The agent brings its own tools (browser, code interpreter, native product features); the benchmark provides none.}
\tnoteline{f}{Questions are public; answers to the 300 test questions are withheld.}
\tnoteline{g}{Graded against 1--10 expert-written pass/fail criteria and gold outputs per task, not by verifying service state.}
\tnoteline{h}{Provides generic action primitives (e.g., UI control, terminal, code execution, desktop actions) rather than a fixed set of tools; counts are not directly comparable.}
\tnoteline{i}{Evaluator checks are reported without a numerical measure of correctness against human judgments or reference labels. Checks performed during benchmark construction count; their timing does not determine the mark.}
\tnoteline{l}{State-based verification on a subset of tasks.}
\tnoteline{m}{Tools are exposed through run-time retrieval or routing over the full tool set (search-based access), not as a fixed per-task set.}
\tnoteline{n}{Servers are deployable without API keys, but tasks operate on live, time-varying data.}
\tnoteline{o}{Mostly self-hosted, but some tasks depend on live cloud services.}
\tnoteline{p}{Read-only tasks by design; grading is a claim-level rubric over the final answer.}
\tnoteline{q}{BFCL v4: 5{,}090 scored cases, counting web-search and memory conditions separately. The agent initially receives a median of three tool definitions per case. We count 3{,}417 distinct function definitions by comparing their full schemas, including description text. Differently worded descriptions of the same operation can therefore be counted separately. No uniform service grouping is available.}
\tnoteline{r}{Executable per-step constraint validation, not verification of the final service state.}
\tnoteline{s}{JobBench reports 130 tasks (65 main and 65 easy); the inspected public release contains 128 (65 main and 63 easy). Some tasks check final database or spreadsheet contents through an LLM judge over extracted artifacts, rather than a running multi-service application.}
\tnoteline{t}{$\tau^2$-bench uses release v0.1.1 (\texttt{2692867}). Its exposed agent toolkits contain 14 airline, 15 retail, and 13 telecom tools: 42 total, with a median of 14 per task (range 13--15). The paper reports 38; our code-based count includes the calculator and human-handoff tools exposed in both airline and retail. The task count retains the paper-reported 279; this release contains 278.}
\tnoteline{u}{Policy or behavioral constraints affect scored outcomes or individual assertions, but the results we inspected do not report quantitative adherence across at least two distinct steering categories. Individually recorded checks count even when combined with task completion in the headline score. HANDBOOK.md records individual policy-check results across multiple behaviors; it receives partial credit because no adherence analysis by category is reported.}
\tnoteline{v}{$\tau^3$-bench uses release v1.0.1 (\texttt{b7ea9074}): 50 airline, 114 retail, 114 telecom, and 97 banking tasks. We count 100 domain-specific agent operations and five document-search tools across the available configurations, giving 105 in total. These counts exclude tools available only to the user simulator and an operation included as an implementation example. Banking discovers additional operations at run time. Under the default configurations, the agent initially receives a median of 16 tools per task (range 13--17). The banking domain provides a shell for searching its document files, while its service APIs remain synthetic. Simulator-validation credit is supported by the banking-domain study reported in $\tau$-Knowledge~\citep{shi2026tauknowledge}.}
\tnoteline{w}{Confidentiality refusal is scored separately and compared with a prompt lacking the confidentiality instruction. This is one steering category, so it receives partial credit under the multi-category criterion.}
\tnoteline{x}{Code checks the final files and service data for required changes and changes forbidden by the task.}
\tnoteline{y}{Median across validation tasks: except in OpenClaw, which retrieves tool schemas through discovery, the agent receives all tools of the services initialized for the task (maximum 324 tools across 9 services) in context.}
\tnoteline{z}{Test-set setup and solution programs are withheld, but evaluation programs are released and run locally. The authors describe a completely hidden test set as future work.}
\par\noindent\textbf{Numeric buckets in Table~\ref{tab:comparison-compact}.} Low (\bktL), mid (\bktM), and high (\bktH) are defined as \#~Services $<6$ / 6--15 / $\geq16$; \#~Tools $<100$ / 100--500 / $>500$; \#~Tools/task $<20$ / 20--50 / $>50$; \#~Services/task $=1$ / 2--5 / $\geq6$; \#~Tasks $<300$ / 300--999 / $\geq1{,}000$.
\tnoteline{$\ddagger$}{Cross-harness variation here is limited to reasoning strategies (e.g., ReAct, DFSDT) or research agent scaffolds, not distinct production agent harnesses.}
\end{flushleft}
\end{table*}

\clearpage

\subsection{Dimension-by-dimension audit}
\label{sec:full-audit}

We assess the benchmarks against each design dimension in \S\ref{sec:dimensions}. Table~\ref{tab:comparison} records the assessments for the tabulated benchmarks, including those with no fixed set of services or tools (marked --); the discussion also covers related instruction-following and interaction studies.

\paragraph{Real tools versus reproducibility} Benchmarks using live services depend on factors outside the evaluator's control. ToolBench~\citep{qin2024toolllm} draws on 16{,}464 live APIs, and StableToolBench's subsequent scan obtained successful responses for 44.4\% of them~\citep{guo2024stabletoolbench}. The failures include connectivity, authorization, parameter, and local documentation-parsing errors, so this is not a direct estimate of service attrition. MCP-Universe~\citep{luo2026mcpuniverse} retrieves real-time ground truth from live servers, so upstream changes can affect results. LiveMCPBench~\citep{mo2026livemcpbench} bundles 70 key-free servers for easy deployment, yet its tasks operate on live, time-varying data. MCP-Atlas~\citep{bandi2026mcpatlas} pins server implementations but still queries live endpoints; its tasks are read-only and single-turn. Score comparability over time therefore depends partly on external services. Local implementations vary in coverage and dependencies: WorkBench~\citep{styles2024workbench} provides 26 tools over five databases; ToolSandbox~\citep{lu2025toolsandbox} provides 34 tools across 11 tool domains, combining local state with live RapidAPI-backed tools; and GAIA2/ARE~\citep{froger2025are} implements simulated applications rather than real services. AppWorld~\citep{trivedi2024appworld} and Toolathlon~\citep{li2026tool} provide 457 and 604 real-world tools, respectively, but AppWorld spans nine services and Toolathlon is not self-contained. GAIA~\citep{mialon2024gaia}, BrowseComp~\citep{wei2025browsecomp}, and GDPval~\citep{patwardhan2025gdpval} instead let each agent bring its own browser, interpreter, or product stack. Their scores reflect both the model and its tooling, without a standardized, shared, distractor-filled tool set or resettable cross-service state.

JobBench~\citep{li2026jobbench} evaluates professional tasks through reference files and rubric-graded deliverables, using tools supplied by the selected agent harness. Some tasks require live web sources, and execution is non-interactive. Its evaluator checks final SQLite database contents on some tasks and scores explicit constraints, including restrictions on disclosing internal references, but the reported results do not break down adherence by steering category. The paper compares Claude Code, Codex CLI, OpenCode, and OpenClaw, including comparisons with the model held fixed, without covering the full model-by-harness grid.

$\tau^3$-bench~\citep{sierra2026tau3} combines revised airline, retail, and telecom tasks with the banking domain introduced in $\tau$-Knowledge~\citep{shi2026tauknowledge} and the full-duplex voice extension $\tau$-Voice~\citep{ray2026tauvoice}. We assess v1.0.1 separately from the original $\tau^2$-bench release. Its 375 standard tasks use local service databases; banking adds a corpus of 698 documents and tools discovered through that corpus. Final-state verification and responsive user simulation are shared with \MV{}, but the service APIs are synthetic and each task stays within one domain.

\paragraph{Tool-schema complexity}
\label{sec:tool-schema-comparison}
AppWorld's published tools~\citep{trivedi2024appworld} average 3.2 leaf input parameters, with a maximum of 17, and use only scalar or scalar-list inputs.\footnote{We use the 457 tools published in AppWorld's \href{https://appworld.dev/appworld/api-explorer}{API Explorer}, including its supervisor and documentation tools.} The 30 tool schemas in $\tau$-bench~\citep{yao2025taubench}, including its two reasoning tools, average 2.7 leaf input parameters and nest at most two levels. In comparison, \MV{} averages 5.9 leaf input parameters per tool, with a maximum of 148 (\texttt{google\_slides.create\_presentation}) and nesting up to eight levels. We count required and optional inputs, expanding object fields and counting each array's element schema once. Flat inputs have depth one; array containers do not add a level.

\paragraph{Multi-service scope and service selection} Even benchmarks with many tools often expose a small task-level action space: 26 tools in WorkBench, a median of 7 in ToolSandbox, a median of 14 agent tools in the original $\tau^2$-bench release (13--15 per domain), a mean of 15.2 in MCP-Atlas, and a median initial exposure of 3 across BFCL v4's 5{,}090 scored cases. ToolSandbox and MCP-Atlas include distractors rather than only the relevant tools. Other benchmarks provide access to many tools through retrieval: ToolBench retrieves top-$k$ APIs, AppWorld fetches API documentation on demand, and LiveMCPBench and AutomationBench~\citep{shepard2026automationbench} route tool access through search over their servers and endpoints. Even when the query omits the service names, a harness can restrict service selection and make the evaluation easier; MCP-Universe supplies only the servers relevant to each task in its main evaluation~\citep{luo2026mcpuniverse}. UltraTool supplies a per-task tool list~\citep{huang2024planning}. \MV{} combines implicit service selection with a large in-context action space: validation tasks present a median of 148.5 tools spanning a median of six services, with all tools of the initialized services available rather than a per-task gold tool subset. Tasks do not name all of the services their solutions require, so the agent must infer where to find the referenced information.

\paragraph{Interaction with users and developers} Many benchmarks provide no user who responds to clarification questions. API-Bank~\citep{li2023apibank} and BFCL~\citep{patil2025berkeley} use predefined user turns, while AgentBench~\citep{liu2024agentbench} provides action-dependent environment feedback rather than a simulated delegating user. GAIA2 schedules scripted user and environment events rather than generating responses to arbitrary clarification questions. ToolBench, the MCP benchmarks, AppWorld, and Toolathlon evaluate without a simulated user in the loop. Reactive user simulators appear in $\tau$-bench, $\tau^2$-bench, MINT~\citep{wang2024mint}, ToolSandbox, CRMArena-Pro~\citep{huang2026crmarenapro}, and TheAgentCompany~\citep{xu2025theagentcompany}. Validation differs across releases: the original $\tau$-bench discusses simulator errors without reporting an accuracy study, whereas $\tau^2$-bench reports errors in 40\% of 50 retail conversations and 47\% of 100 airline conversations, including both benign and task-blocking errors~\citep{barres2025tau2bench}. In $\tau^3$-bench's banking domain, introduced in $\tau$-Knowledge, the simulated user reveals private details in response to targeted questions. The $\tau$-Knowledge study reports that two domain-expert annotators reviewed 194 banking trajectories, of which four contained task-critical simulator errors~\citep{shi2026tauknowledge}. This result concerns banking and task-critical errors, not all simulator mistakes across $\tau^3$-bench.

\paragraph{Category-wise steering evaluation} We apply the steering criteria defined in Table~\ref{tab:comparison}. CCTU~\citep{ye2026cctubenchmarktooluse} meets this criterion through its constraint categories and category-specific violation rates. IFEval~\citep{zhou2023ifeval} also meets it through explicit response constraints, programmatic checks, and category-level results. AgentIF~\citep{qi2025agentif} extends instruction evaluation to agent specifications, including semantic and tool-related constraints, with results by behavioral category; its released evaluation scores the next response rather than executing service operations. Other benchmarks meet only part of this criterion: TheAgentCompany~\citep{xu2025theagentcompany} has confidentiality and escalation-timing checkpoints, Terminal-Bench~\citep{merrill2026terminalbench} tests method and edit restrictions, and CRMArena-Pro~\citep{huang2026crmarenapro} reports a separate confidentiality-refusal rate. HANDBOOK.md~\citep{surge2026handbook} and AutomationBench~\citep{shepard2026automationbench} have individual policy-related checks, but the analyses we inspected do not report adherence rates across distinct steering categories. MCP-Universe and MINT~\citep{wang2024mint} separately analyze output-format compliance. MINT reports how often models fail to mark whether their output is code to execute or an answer to evaluate; neither analysis covers multiple steering categories. In $\tau^3$-bench, policy checks affect its scores, while the published failure categories and voice-interaction metrics do not constitute adherence results for multiple explicitly instructed steering categories~\citep{shi2026tauknowledge,ray2026tauvoice}. LiveMCPBench requires tool-derived information in grading, while its agent prompt requests tool use only ``whenever possible,'' so we do not count it as an evaluated steering instruction. Comparisons with and without behavioral steering instructions, such as $\tau$-bench's policy-removal ablation~\citep{yao2025taubench}, test changes from default behavior. They neither count toward this criterion nor are required for full credit.

\paragraph{Cross-harness evaluation} Many tool-use benchmarks evaluate a single fixed harness. ToolBench~\citep{qin2024toolllm} is an exception: it evaluates its models with both ReAct and DFSDT, providing a full comparison of two research scaffolds. Terminal-Bench~\citep{merrill2026terminalbench} and OSWorld-Verified~\citep{xie2024osworld,osworldverified2025} accept agent--model pairs on their leaderboards; Terminal-Bench also compares selected models across harnesses. Agents' Last Exam~\citep{sun2026agents} reports fixed-harness model sweeps and fixed-model harness comparisons, but does not evaluate and analyze its full model-by-harness grid. GAIA includes AutoGPT alongside direct model execution. GDPval evaluates models with tools supplied by their respective products, such as web search, code execution, and file creation. These differences in available tools are not a comparison of the same model running under different harnesses. MCP-Universe~\citep{luo2026mcpuniverse} compares ReAct with Cursor Agent for Claude-4.0-Sonnet and with OpenAI Agents SDK for o3 across six domains. Cursor trails ReAct overall (26.41\% versus 29.44\%), whereas OpenAI Agents SDK improves on ReAct (31.60\% versus 26.41\%). These selected comparisons do not form a full model-by-harness grid. The $\tau^3$-bench leaderboard includes a tool-mentor agent with an additional model that can block state-changing tool calls or transfers to a human and require the agent to continue after a blocked transfer. It also includes cascaded voice pipelines~\citep{sierra2026tau3}. These submissions do not cover a full model-by-harness grid; the mentor entry also changes simulator and audio settings, so it is not a controlled harness comparison. HAL~\citep{stroebl2026holistic} compares specialized and generalist agents across existing benchmarks and analyzes model-dependent scaffold effects, although its evaluation matrix is incomplete. A common model-call limit does not itself equalize monetary or token costs. \MV{}'s protocol runs every evaluated model under a bare harness and each of several widely used end-user harnesses on identical tasks, making model and harness explicit experimental dimensions.

\paragraph{Hidden test set} Released tasks can enter training data. A held-out portion evaluated only through submission, as in SWE-Bench Pro~\citep{deng2025swebench}, GDPval~\citep{patwardhan2025gdpval}, GAIA2, and Agents' Last Exam~\citep{sun2026agents}, limits contamination. \MV{} withholds a hidden test set of 400 of its 796 tasks for this purpose and plan to support a model-submission-based leaderboard.

\paragraph{Reliable verification} Trustworthy scores require verifying what happened to the services. Verification targets vary: ToolBench judges solution paths~\citep{qin2024toolllm}; WebArena combines answer, URL, and program-state checks~\citep{zhou2024webarena}; CRMArena-Pro evaluates answers and confidentiality refusals~\citep{huang2026crmarenapro}. Other approaches include milestone matching against intermediate database snapshots (ToolSandbox), LLM-judge scoring over the answer and tool-call trace (LiveMCPBench~\citep{mo2026livemcpbench}), claim-level rubrics over answers (MCP-Atlas), or expert-written pass/fail criteria over deliverables rather than resulting state (APEX-Agents~\citep{vidgen2026apex}). State-based verification exists in $\tau$-bench, $\tau^2$-bench, AppWorld, WorkBench, BFCL's synthetic backends, and Toolathlon, though it must still credit valid alternative outcomes. WebArena, AgentBench, TheAgentCompany, and MINT's ALFWorld tasks also check resulting environment or service state. GAIA2 instead checks state-changing tool calls against reference actions and their ordering requirements. It permits independent actions to occur in either order and is designed to accept semantically equivalent arguments, but does not accept every trajectory that reaches the same final state. Its verifier reaches 0.98 agreement with human labels over 450 trajectories~\citep{froger2025are}. OSWorld-Verified~\citep{xie2024osworld,osworldverified2025} subjects its execution-based evaluation scripts to review by independent domain experts, while Terminal-Bench~\citep{merrill2026terminalbench} and SWE-Bench Pro~\citep{deng2025swebench} manually verify their tests during construction.

\paragraph{Infrastructure-error isolation} We inspected the released evaluation code of benchmarks in Table~\ref{tab:comparison} for error classification, score exclusion, and retry support. Several score an inference-time exception as the agent's response or as a zero reward~\citep{patil2025berkeley,yao2025taubench,lu2025toolsandbox}, treat errored tasks as completed, so resuming does not rerun them~\citep{bandi2026mcpatlas,huang2026crmarenapro}, or drop crashed tasks from the reported denominator without noting it~\citep{zhou2024webarena}. Wall-clock budgets can also confound scores when provider retries consume the time available for agent decisions. Toolathlon kills a rollout at its time limit and records the timeout as a failed task~\citep{li2026tool}. The $\tau^2$-bench v0.1.1 runner can retry interrupted runs but does not record typed infrastructure errors or explicitly exclude them from scores~\citep{barres2025tau2bench}. The $\tau^3$-bench runner classifies, excludes, and retries infrastructure errors, but some published leaderboard scores still count them as failures~\citep{sierra2026tau3}. GAIA2 distinguishes errors and retries missing or errored scenarios, although its reported pass rate still includes errors in the denominator~\citep{froger2025are}. Agents' Last Exam records a six-category infrastructure taxonomy and leaves failed units unscored rather than assigning them zero~\citep{sun2026agents}. \MV{} uses model-call budgets and reruns detected infrastructure failures rather than scoring them as task failures.

\paragraph{Right-sized task sets} Task-set size trades evaluation precision against the cost of QC for individual tasks and repeated runs. Large sets such as UltraTool (5{,}824 tasks) and CRMArena-Pro (4{,}280) use templating or synthetic generation; evaluating thousands of tasks across a full model$\times$harness grid with LLM judges also increases cost. Smaller sets, such as Terminal-Bench (89) and LiveMCPBench (95), provide less precision for distinguishing similarly performing models. \MV{}'s manually created task set contains 396 validation tasks and 400 hidden test tasks, distributed approximately evenly across the four settings. Its size balances the cost of QC for individual tasks and repeated evaluation against comparison precision. Task-sampling intervals report the precision of the resulting comparisons (\S\ref{sec:setup}).

\section{Benchmark construction and task details}
\label{sec:construction}

\begin{figure}[tbp]
\centering
\begingroup
\setlength{\unitlength}{0.01\linewidth}
\setlength{\fboxsep}{0pt}
\setlength{\fboxrule}{0.4pt}
\fontsize{9}{10}\selectfont
\newcommand{\workflowbox}[4]{%
  \makebox(0,0)[lb]{%
    \fcolorbox{black!55}{black!3}{%
      \parbox[c][#2\unitlength][c]{#1\unitlength}{%
        \centering\textbf{#3}\\[2pt]#4}}}}
\begin{picture}(100,127)
  \put(0,16){\color{black!30}\framebox(100,110){}}
  \put(3,69){\color{black!25}\line(1,0){94}}

  \put(2,121.5){\makebox(96,3)[l]{\textbf{Task construction}}}
  \put(2,116.5){\makebox(96,3)[l]{\fontsize{8}{9}\selectfont
    Task batches progressed concurrently: construction, QC, and correction period overlapped.}}

  \put(2,92.5){\workflowbox{16}{11}{Authors}{Specify task and\\evaluation\\requirements}}
  \put(23,92.5){\workflowbox{19}{11}{Vendor A}{Construct tasks;\\three internal\\QC passes}}
  \put(50,104.5){\workflowbox{24}{9}{Vendor A}{Internal QC;\\repeat after each fix}}
  \put(50,93.5){\workflowbox{24}{9}{Vendor B}{Task-QC team:\\independent QC}}
  \put(50,82.5){\workflowbox{24}{9}{Authors}{Informal QC of\\task subsets}}
  \put(83,92.5){\workflowbox{15}{11}{Vendor A}{Correct task\\defects}}

  \thicklines
  \put(18.2,98){\vector(1,0){4.5}}
  \put(42.2,98){\line(1,0){3.8}}
  \put(46,87){\line(0,1){22}}
  \put(46,109){\vector(1,0){3.8}}
  \put(46,98){\vector(1,0){3.8}}
  \put(46,87){\vector(1,0){3.8}}
  \put(74.2,109){\line(1,0){0.6}}
  \put(74.2,98){\line(1,0){0.6}}
  \put(74.2,87){\line(1,0){0.6}}
  \put(74.8,87){\line(0,1){22}}
  \put(74.8,98){\vector(1,0){8}}
  \put(74.8,98.6){\makebox(8,3){\fontsize{7}{8}\selectfont Feedback}}
  \put(90.5,92.3){\line(0,-1){15.3}}
  \put(46,77){\line(1,0){44.5}}
  \put(46,77){\vector(0,1){10}}
  \put(50,77.8){\makebox(37,3){\scriptsize Revised tasks}}
  \put(43,70.8){\makebox(52,6){\parbox{52\unitlength}{\centering
    \emph{Authors mediated QC rounds\\and resolved disagreements.}}}}

  \put(3,64.2){\makebox(94,3)[l]{\textbf{Service implementation}}}
  \put(3,41){\workflowbox{25}{11}{Vendor B}{Engineering team:\\build and test services}}
  \put(39,54.5){\workflowbox{28}{7.5}{Vendor B}{Internal QC;\\repeat after each fix}}
  \put(39,46){\workflowbox{28}{7.5}{Vendor C}{Independent service QC}}
  \put(39,37.5){\workflowbox{28}{7.5}{Vendor A}{Informal QC of API subsets\\during task construction}}
  \put(39,29){\workflowbox{28}{7.5}{Authors}{Informal QC of\\API subsets}}
  \put(78,41){\workflowbox{19}{11}{Vendor B}{Engineering team:\\correct service\\defects}}
  \put(28.2,46.5){\line(1,0){5.8}}
  \put(34,32.75){\line(0,1){25.5}}
  \put(34,58.25){\vector(1,0){4.8}}
  \put(34,49.75){\vector(1,0){4.8}}
  \put(34,41.25){\vector(1,0){4.8}}
  \put(34,32.75){\vector(1,0){4.8}}
  \put(67.2,58.25){\line(1,0){1.3}}
  \put(67.2,49.75){\line(1,0){1.3}}
  \put(67.2,41.25){\line(1,0){1.3}}
  \put(67.2,32.75){\line(1,0){1.3}}
  \put(68.5,32.75){\line(0,1){25.5}}
  \put(68.5,46.5){\vector(1,0){9.3}}
  \put(68.5,47.1){\makebox(9.3,3){\fontsize{7}{8}\selectfont Feedback}}
  \put(87.5,40.8){\line(0,-1){15.8}}
  \put(34,25){\line(1,0){53.5}}
  \put(34,25){\vector(0,1){7.75}}
  \put(39,25.8){\makebox(46,3){\scriptsize Revised services}}
  \put(34,21.5){\makebox(63,3){\emph{Authors mediated between engineering and QC teams.}}}

  \put(50,16){\vector(0,-1){5.5}}
  \put(52,12){\makebox(46,3)[l]{\scriptsize After both vendor workflows}}
  \put(10,0){\workflowbox{80}{10}{Authors}{%
    Correct task and API defects found during experiments and analyses\\
    after the vendor QC workflows.}}
\end{picture}
\endgroup
\caption{Overall workflow for task construction and service implementation. Task batches progressed concurrently, with construction, QC, and revision periods overlapping. Authors conducted QC on subsets of tasks and APIs. Vendor A conducted QC on API subsets during task creation. Vendor B's task-QC and service-engineering teams are separate.}
\label{fig:vendor-workflow}
\end{figure}
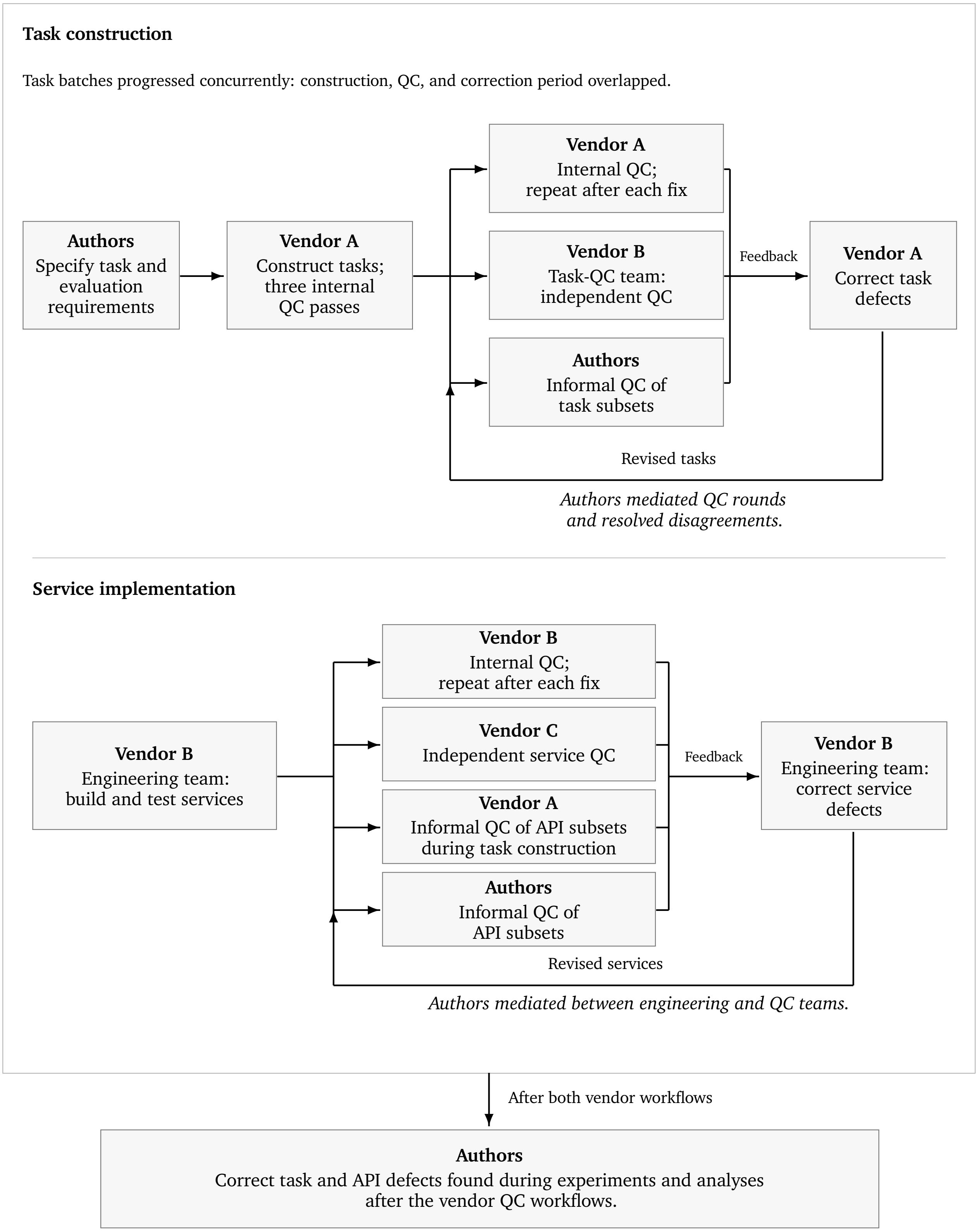

The authors first specified the task and evaluation requirements. Task construction had two stages: constructing the task and gold solution, then building the verifiers and testing the task with reference models. Three QC passes at Vendor~A checked task validity, verifier correctness, and consistency between the two, and Vendor~B performed independent QC on every task (\S\ref{sec:qc}). After the vendor QC workflows, authors continued correcting defects in both tasks and services. Figure~\ref{fig:vendor-workflow} summarizes the workflow for task construction and service implementation.
We describe the construction of each component and the diversity controls used across tasks.

\subsection{Service implementation}
\label{sec:env-gen}

The 35 services draw on public API documentation, documented tool interfaces, and implementation documentation, as listed in Table~\ref{tab:service-grounding}. WhatsApp models the consumer application, while GitHub Actions follows an MCP-server interface. The 11 phone and device functions follow implementation documentation. SAP Concur exposes Concur's Travel Profile and Itinerary schemas, with booking inventory and policy seeded from $\tau^2$-bench's airline domain. Shopify exposes Shopify's customer, order, and product resources, with order support implemented locally. The services were built by an engineering team at Vendor~B, separate from Vendor~B's task-QC team (\S\ref{sec:qc}). Vendor~B performed internal QC on the services, Vendor~C performed independent QC over several rounds, and the authors conducted QC on API subsets. Vendor~A also conducted QC on API subsets when issues arose during task creation. Vendor~B's engineering team fixed the reported defects, with two authors mediating between the vendors. Tool bugs that the task-QC team or we found were also reported to the engineering team. The scope of each service was set explicitly. Operations that concern transport and authentication (HTTP, OAuth scopes, webhooks), media bytes, asynchronous machinery (timers, triggers, delivery retries), and provider quotas are omitted, and the omissions are enumerated per service together with the service-specific behaviors that were reproduced, such as the query semantics of a search endpoint or the atomicity of a batch update (Table~\ref{tab:service-grounding}). These include the parts a user configures once when connecting a service to a harness. Behavior was matched to the specification and, for Google Calendar, to captures from live accounts at four permission levels. Semantic search uses a deterministic local encoder, so the simulated services do not depend on a live embedding API.

{\footnotesize
\newcommand{\serviceseparator}{\arrayrulecolor{black!35}\midrule\arrayrulecolor{black}}
\setlength{\tabcolsep}{4pt}
\renewcommand{\arraystretch}{1.08}
\begin{longtable}{@{}>{\raggedright\arraybackslash}p{1.9cm}r>{\raggedright\arraybackslash}p{4.2cm}>{\raggedright\arraybackslash}p{6.2cm}@{}}
\caption{Grounding of the 35 simulated services, the number of tools each exposes, and their simulation scope and interface adaptations. Services are listed under their original names (Table~\ref{tab:service-mapping}). The phone and device functions were implemented from implementation documentation; where the table lists nothing omitted, the documentation was implemented as specified.}
\label{tab:service-grounding} \\
\toprule
\textbf{Service} & \textbf{Tools} & \textbf{Grounding} & \textbf{Simulation scope and interface adaptations} \\
\midrule
\endfirsthead
\multicolumn{4}{@{}l}{\emph{Table~\ref{tab:service-grounding}, continued}} \\
\toprule
\textbf{Service} & \textbf{Tools} & \textbf{Grounding} & \textbf{Simulation scope and interface adaptations} \\
\midrule
\endhead
\bottomrule
\endlastfoot
Gmail & 60 & Gmail REST API v1 & No mail transport: delivery is local and unknown recipients are skipped; history cursors partly ignored; adds recipient, subject, and body convenience fields for sending \\
\serviceseparator
Google Calendar & 32 & Calendar API v3 (discovery revision 20260218), checked against live accounts at four permission levels & No quickAdd or incremental sync tokens; watch channels stored but no callbacks \\
\serviceseparator
Google Drive & 47 & Drive REST API v3 & No revision update/delete, OCR, export conversion, thumbnails, client-side encryption, or approvals \\
\serviceseparator
Google Docs & 3 & Docs REST API v1 & 5 of 40 batchUpdate request types; simplified content model; no suggestions or tabs \\
\serviceseparator
Google Slides & 5 & Slides REST API v1 & 11 of 44 batchUpdate request types; no thumbnails, media fetch, or chart refresh; creation additionally accepts initial slide content \\
\serviceseparator
Google Sheets & 15 & Sheets REST API v4 & 4 of 69 batchUpdate operations; cell formulas are stored, but their results are not computed; no pivots, charts, or live imports \\
\serviceseparator
Google Chat & 31 & Chat REST API v1 & 31 of 54 methods; no custom emoji; OAuth replaced by a stored current user; no binary attachments \\
\serviceseparator
Google Meet & 16 & Meet REST API v2 & Spaces are mutable; conference history and artifacts read from fixtures; no caller authorization \\
\serviceseparator
Slack & 64 & Slack Web API (chat, conversations, files, reminders, search, users, usergroups) & No OAuth scopes or rate limits; message formatting and unfurl controls dropped; some cross-resource side effects omitted \\
\serviceseparator
WhatsApp & 14 & Consumer WhatsApp app (not the Cloud/Business API) & No encryption, servers, presence, status, payments, or group creation; media stored as placeholders \\
\serviceseparator
Messages & 11 & Implementation documentation (SMS and chat) & No rich media streaming, MMS attachments, or RCS applets; delivery is instantaneous and synchronous, without carrier retries \\
\serviceseparator
Zendesk & 22 & Support REST API v2 (organizations, users, tickets, comments, attachments, audits, search) & No inbound email parsing, triggers, SLA timers, webhooks, or binary storage; macros and custom fields are placeholders \\
\serviceseparator
Salesforce & 51 & Event and Task objects plus the REST Query resource (SOQL), version-neutral subset & No Apex, triggers, flows, sharing rules, or OAuth; SOQL over seven sObjects; the related-record field (WhatId) can reference only Account, Opportunity, or Case records \\
\serviceseparator
HubSpot & 37 & HubSpot REST APIs v2 & No OAuth or tenancy; email sends recorded without delivery; campaigns omit links to marketing assets, revenue attribution, and business-unit information; requests cannot freely select which fields to return \\
\serviceseparator
Stripe & 23 & Payments and Billing REST (customers, products, prices, payment intents, payment links, invoices, refunds, subscriptions, coupons, disputes, balance) & No confirm/capture, webhooks, or invoice pay/void; refunds succeed immediately; totals exclude tax \\
\serviceseparator
Shopify & 29 & Shopify customer, order, and product resources; order support implemented locally & Payments, refunds, and inventory are database operations; transaction receipts are synthetic \\
\serviceseparator
SAP Concur & 16 & Travel Profile v2 and Itinerary APIs (v1.1/v4); booking inventory and policy seeded from $\tau^2$-bench's airline domain & Expense and invoice suites omitted; no live distribution system; static seat inventory \\
\serviceseparator
Jira & 88 & REST API v2 (Jira 6.1 reference) with JQL & No OAuth, quotas, marketplace apps, mail, or automation; JQL supports comparisons, logical combinations, result ordering, and pagination, but only a subset of Jira's query fields and functions; local attachments \\
\serviceseparator
Confluence & 39 & REST API v1 (Confluence 5.7.1 reference) with CQL & No OAuth, licensing, plugins, or collaborative editing; CQL supports selected query fields and operators, with functions for the current time, current user, and start of the year; body conversion changes markup tags and wrappers but does not render macros \\
\serviceseparator
GitHub & 32 & REST API, reference version 2022-11-28 & Files, commits, and branches as records; no Git transport, CI runners, or webhooks; merge commits use the files from the pull-request branch rather than combining changes from both branches \\
\serviceseparator
GitHub Actions & 9 & GitHub Actions MCP server (nine operations) & Same nine operations; no live GitHub calls, token permissions, or rate limits \\
\serviceseparator
LinkedIn & 18 & REST/Rest.li APIs & No OAuth or versioning headers; media metadata only; organization lookup by vanity name only \\
\serviceseparator
Instagram & 10 & Platform API v25.0 (professional accounts) & No OAuth or permission scoping; posts created immediately from a URL; local fixture identities \\
\serviceseparator
TikTok & 4 & Business API v1.3 (profile data, video publishing) & No metric windows or latency; synchronous publishing; custom post metadata \\
\serviceseparator
Spotify & 41 & Web API (users/social, artists/albums, browse, playlists) & No OAuth scopes; list tools auto-paginate; batch mutations validate all IDs up front \\
\serviceseparator
Clock & 18 & Implementation documentation (OpenAPI contract, provider-neutral) & No OS alarm registration, sound, vibration, notification, or lock-screen integration; no background ticking, state derived at call time; recurrence is a weekly weekday set \\
\serviceseparator
Reminders & 5 & Implementation documentation (OpenAPI contract, IANA time zones) & No notifications, device alarms, cloud sync, or background scheduler; recurrence limited to supported patterns; log stores before-images for undo, not delivery events \\
\serviceseparator
Phone & 4 & Implementation documentation (device function) & None; implemented as specified \\
\serviceseparator
Contacts & 9 & Implementation documentation (device function) & None; implemented as specified \\
\serviceseparator
Notifications & 3 & Implementation documentation (mobile notification tray) & No creation or push of notifications during evaluation (tray is seeded); permission checks always pass; no binary media or interactive actions \\
\serviceseparator
Media control & 11 & Implementation documentation (playback control) & Local player state only; no audio or video output, downloads, or streaming accounts; position does not advance with time \\
\serviceseparator
Device actions & 12 & Implementation documentation (Android-style system and intent actions) & State-only app dispatch and navigation; media capture records metadata, not files; no network fetch or page rendering \\
\serviceseparator
Device settings & 14 & Implementation documentation (device settings and system control) & Single in-memory device; Wi-Fi scan returns seven fixture networks; volumes are independent 0--100 values; health metrics from stored snapshots \\
\serviceseparator
Notes and lists & 14 & Implementation documentation (personal notes and checklists) & Revision history without depth caps; undo snapshots only for title updates; no cloud sync, reminder alarms, sharing, or deep links \\
\serviceseparator
Generic media & 2 & Implementation documentation (media search and selection across providers) & Local content only; no provider calls or account, subscription, or region checks; no audio, streaming, or playback queue; search returns up to 100 matches \\
\end{longtable}
}

\paragraph{Implementation validation} Unit tests check request validation, state transitions, pagination, and error responses, with 88\% statement coverage across the service implementations and shared utilities. These tests assess conformance to documented interfaces, not equivalence to live-service behavior (\S\ref{sec:limitations}).

\subsubsection{Service name mapping}
\label{sec:service-mapping}

Table~\ref{tab:service-mapping} maps the original service names used in this paper to the pseudo service names used in the released benchmark. Services without a brand retain their generic names. Renaming changes service names and tool-name prefixes, not tool schemas.

\begin{table}[H]
\centering
\footnotesize
\setlength{\tabcolsep}{5pt}
\renewcommand{\arraystretch}{1.05}
\caption{Original service names (as used in this paper) and their pseudo service names in the released \MV{} benchmark. Renaming changes only service names and tool-name prefixes; tool schemas are unchanged.}
\label{tab:service-mapping}
\begin{tabular}{@{}ll@{\hspace{2em}}ll@{}}
\toprule
\textbf{Original} & \textbf{Released} & \textbf{Original} & \textbf{Released} \\
\midrule
\texttt{clock} & \texttt{clock} & \texttt{instagram} & \texttt{photogram} \\
\texttt{confluence} & \texttt{docfluence} & \texttt{jira} & \texttt{zira} \\
\texttt{contacts} & \texttt{contacts} & \texttt{linkedin} & \texttt{linkedup} \\
\texttt{device\_actions} & \texttt{device\_actions} & \texttt{media\_control} & \texttt{media\_control} \\
\texttt{device\_setting} & \texttt{device\_setting} & \texttt{messages} & \texttt{messages} \\
\texttt{gdrive} & \texttt{gridwell\_drive} & \texttt{notes\_and\_lists} & \texttt{notes\_and\_lists} \\
\texttt{generic\_media} & \texttt{generic\_media} & \texttt{notifications} & \texttt{notifications} \\
\texttt{generic\_reminders} & \texttt{generic\_reminders} & \texttt{phone} & \texttt{phone} \\
\texttt{github} & \texttt{githut} & \texttt{salesforce} & \texttt{salespower} \\
\texttt{github\_actions} & \texttt{githut\_actions} & \texttt{sapconcur} & \texttt{sapspend} \\
\texttt{gmail} & \texttt{gridwell\_mail} & \texttt{shopify} & \texttt{shopster} \\
\texttt{google\_calendar} & \texttt{gridwell\_calendar} & \texttt{slack} & \texttt{slick} \\
\texttt{google\_chat} & \texttt{gridwell\_chat} & \texttt{spotify} & \texttt{streamify} \\
\texttt{google\_docs} & \texttt{gridwell\_docs} & \texttt{stripe} & \texttt{strope} \\
\texttt{google\_meet} & \texttt{gridwell\_meet} & \texttt{tiktok} & \texttt{cliptok} \\
\texttt{google\_sheets} & \texttt{gridwell\_sheets} & \texttt{whatsapp} & \texttt{whatsup} \\
\texttt{google\_slides} & \texttt{gridwell\_slides} & \texttt{zendesk} & \texttt{zenhelp} \\
\texttt{hubspot} & \texttt{hubscape} &  &  \\
\bottomrule
\end{tabular}
\end{table}

\subsection{Graphical interfaces}
\label{sec:gui}

\begin{figure}[!tp]
\centering
\includegraphics[width=\linewidth,height=0.41\textheight,keepaspectratio]{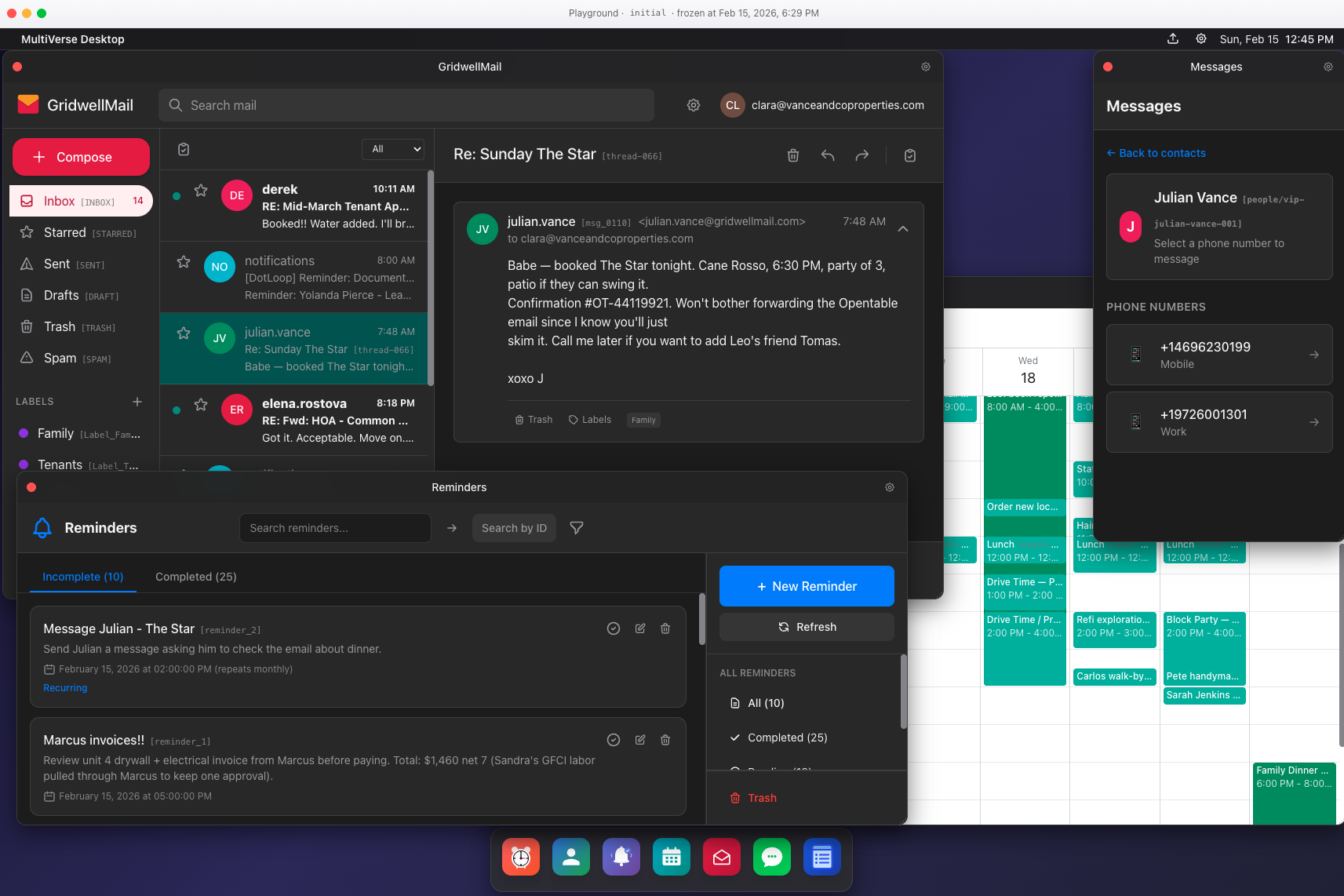}
\caption{Initial state of a family-dinner task, viewed through the desktop GUI. Julian's email and the open reminder are visible alongside the calendar and his messaging contact.}
\label{fig:gui-initial}
\medskip
\includegraphics[width=\linewidth,height=0.41\textheight,keepaspectratio]{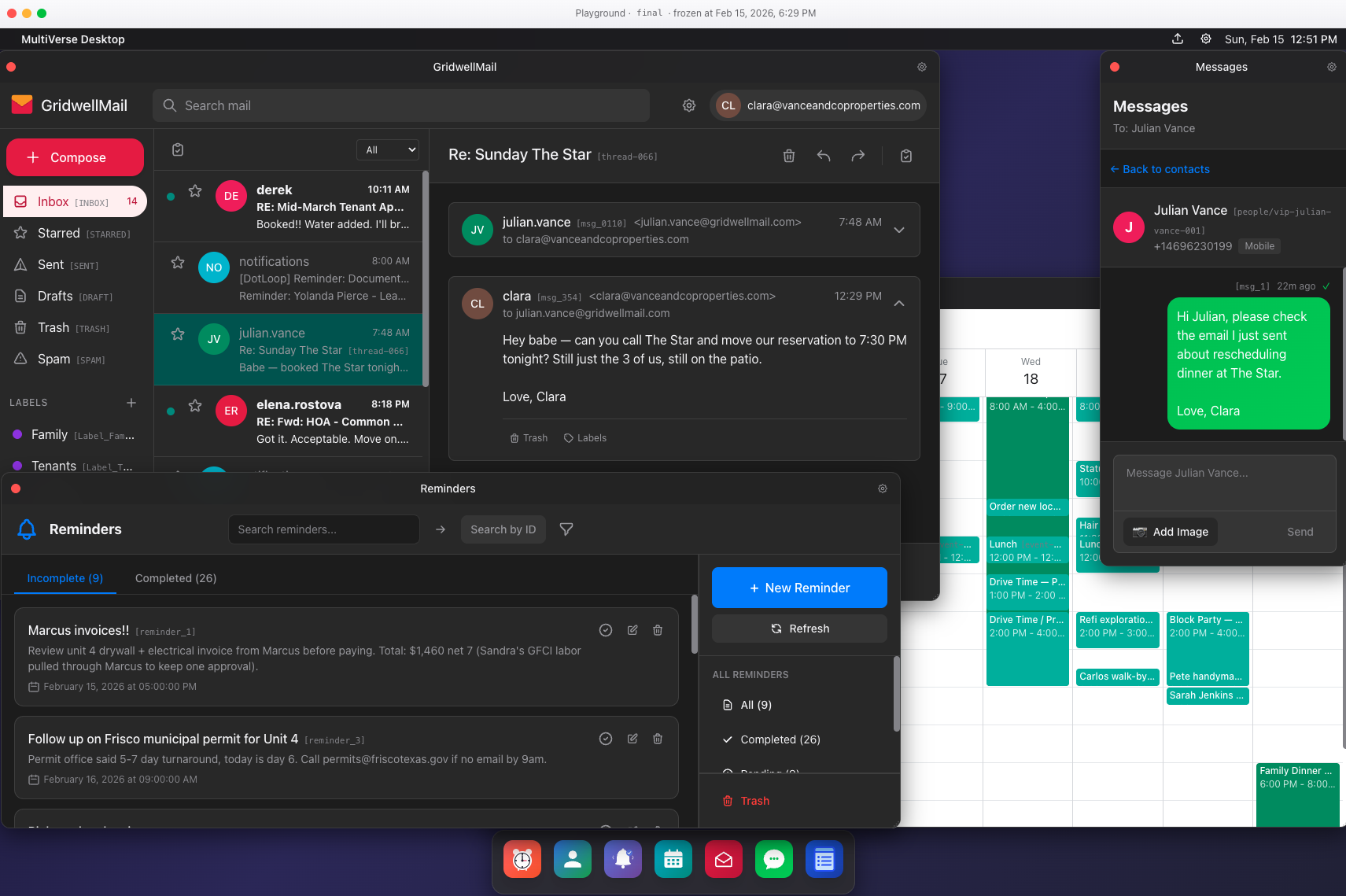}
\caption{Final state of the same family-dinner task. The email reply requests a 7:30~PM reservation, the text asks Julian to check his email, and the reminder no longer appears among incomplete reminders.}
\label{fig:gui-final}
\end{figure}

Vendor~A built mock graphical interfaces for the simulated services; their visual design differs from that of the real services. A task visualizer displays the initial and final database states, their differences, and the tool calls and responses in a trajectory. Its desktop interface provides applications for all 35 simulated services. Users can browse records and perform actions, such as sending a message or editing a calendar event, through the same simulated APIs used by agents. Each playground operates on a separate copy of the task state and can reload the initial or final snapshot. Figures~\ref{fig:gui-initial} and~\ref{fig:gui-final} illustrate this interface with a family-dinner task. We release these interfaces with the tools. They could support annotation tools that let annotators solve tasks through the GUI and record the corresponding tool calls as gold trajectories. They could also support a computer-use version of \MV{}, where agents operate the same simulated services through the GUI.

The interfaces are supplementary and are not used in the evaluation: agents use the simulated services only through tool calls, so a defect in an interface does not affect benchmark results. They have therefore not undergone thorough QC, may contain defects, and leave room for improvement.

\subsection{Initial state generation}
\label{sec:workspace-gen}

For each task, an annotator specified the services, the intended task, and constraints on the initial service state. Gemini~3.1 Pro first wrote a blueprint: a structured list of the entities the task needs (people, messages, events, orders) and the relationships between them, which an independent judge reviewed. Service databases were then generated from this blueprint, with foundational entities such as contacts generated before the records that refer to them, so the same entities appeared consistently across the task's services. Schema checks and a second judge checked the resulting initial state for validity and consistency; failed checks triggered revisions. The annotator inspected and corrected the initial state before task verification.

\subsection{Task construction}
\label{sec:task-construction}

In Stage 1, the annotator designed the task end to end: name and motivation, service and tool selection, narrative context, developer/user instructions where applicable, target capabilities, the gold trajectory, and the expected outcome. Tasks were designed to challenge strong reference models, following adversarial data collection with models in the loop~\citep{nie2020adversarial,kiela2021dynabench}. For each task, one of three reference models (GPT~5.4, Claude Opus~4.6, and Gemini~3.1 Pro, the first two with a 1M-token context window) was chosen at random, and the task was admitted only if that model, run under the bare harness, failed it without hints or guidance from the annotator. Gold trajectories were produced by running the task's sampled reference model on the task under the annotator's guidance, with hints or corrections at the steps where it went wrong, and recording the successful execution. We chose this over having annotators solve tasks by hand because handwritten solutions often diverged from what an agent would do: annotators already held a model of the task and its expected approach, and their solutions relied on judgment and priors that the agents could not reproduce, which made them a poor reference for the verifiers and for difficulty calibration. For the instruction-bearing settings, the annotator started from a base task that the chosen model completed (a single-turn task for IF, a multi-turn task for MTIF) and added developer/user instructions until it failed. This construction filter made the instructions the intended source of the reference-model failure, but the evaluation settings, which consist of different tasks, do not identify a general causal instruction effect.
In Stage 2, reference-model runs checked that the task met the admission criterion. Gold solutions establish a successful solution path through the services (\S\ref{sec:tasks}).

\subsubsection{Instruction rubric categories}
\label{sec:if-types}

The IF and MTIF evaluation assesses 1{,}046 rubrics from 731 instructions across 97 IF and 99 MTIF tasks (median 3 instructions and 4 rubrics per task; 1--10 instructions), grouped into the nine categories in Table~\ref{tab:if-groups}. A check is an observable requirement used by the instruction verifier; a compound instruction can contain checks from several categories.

\begin{table}[t]
\centering
\scriptsize
\setlength{\tabcolsep}{3pt}
\renewcommand{\arraystretch}{1.02}
\caption{Instruction rubric categories, definitions, examples paraphrased from validation checks, and the number of checks in each setting. Categories refer to the behavior checked, not necessarily every requirement in the parent instruction.}
\label{tab:if-groups}
\begin{tabular}{@{}>{\raggedright\arraybackslash}p{2.4cm}>{\raggedright\arraybackslash}p{4.2cm}>{\raggedright\arraybackslash}p{3.9cm}cc@{}}
\toprule
\textbf{Instruction rubric category} & \textbf{Definition} & \textbf{Example} & \textbf{IF} & \textbf{MTIF} \\
\midrule
Argument and outcome constraints & Requirements specifying argument values, written content, targets, scope, or final-state properties & Include the specified recipient in the calendar event's attendees & 167 & 179 \\
\tablerowseparator
Response content and presentation & User-facing messages' content, wording, structure, or formatting & Put the created event's title on its own line in the final response & 148 & 66 \\
\tablerowseparator
Ordering and preconditions & A named action must precede another or satisfy a prerequisite & Read the meeting transcript before updating the spreadsheet & 85 & 117 \\
\tablerowseparator
Tool-output grounding & Claims, arguments, and decisions must faithfully reflect retrieved evidence & Ground the BBQ limit in the written reply, not an inaccessible voice note & 45 & 51 \\
\tablerowseparator
Lookup and provenance & Obtain information from an authoritative source or preserve its provenance & Include the source draft's title or version in the reminder & 21 & 44 \\
\tablerowseparator
Specific-tool workflows & Use or avoid a specified tool or service in preference to an alternative & Read the document body with Docs, not a Drive export & 24 & 28 \\
\tablerowseparator
Clarification and assumptions & Requirements on when and how to ask for missing information or resolve ambiguity, including when to proceed without asking & Ask when the teams should bring their robots onto the stage & 4 & 40 \\
\tablerowseparator
Process constraints & Restrictions on redundant calls or prescribed procedures & Avoid repeating a tool call once its result is already known & 8 & 8 \\
\tablerowseparator
Unsupported-action and recovery requirements & Handle unavailable operations or tool failures as instructed & State that engagement metrics are unavailable when none are retrieved & 7 & 4 \\
\bottomrule
\end{tabular}
\end{table}

\subsection{Agent capability annotations}
\label{sec:capabilities}

Annotators labeled the capabilities required to solve each task. Table~\ref{tab:capability-labels} briefly defines these labels.

\begin{table}[!htbp]
\centering
\footnotesize
\setlength{\tabcolsep}{5pt}
\renewcommand{\arraystretch}{1.12}
\caption{Capability labels, grouped by top-level category.}
\label{tab:capability-labels}
\begin{tabular}{@{}>{\raggedright\arraybackslash}p{0.43\linewidth}>{\raggedright\arraybackslash}p{\dimexpr0.57\linewidth-2\tabcolsep\relax}@{}}
\arrayrulecolor{black}
\toprule
\textbf{Capability} & \textbf{Description} \\
\midrule
\multicolumn{2}{@{}l}{\textbf{Planning \& Execution}} \\[2pt]
Search Synthesis (Mental Joins) & Find and connect relevant information across sources. \\
Order of Operations (Logical Dependencies) & Respect action dependencies and API prerequisites. \\
Temporal \& Interval Logic & Reason about times, durations, overlaps, and schedules. \\
Parallel Execution \& Batching & Group independent actions into a single response turn. \\
Quantitative/Formal Logic & Perform arithmetic or symbolic reasoning under formal rules. \\
Tool Definition Matching \& Schema Compliance & Select suitable tools and supply schema-compliant arguments. \\
Data Marshalling (Glue Work) & Convert tool outputs into formats required by subsequent tools. \\
Search Iteration \& Reformulation & Change strategy when searches return unhelpful results. \\
Hypothesis Generation (Abductive) & Propose testable explanations for unexpected tool outputs. \\
Technical Error Recovery & Recover from tool errors using retries, fixes, or fallbacks. \\
Human Feedback Adaptation & Revise beliefs or plans following human corrections. \\
Sustained Execution & Execute accurately through long workflows and large datasets. \\
User Confirmation & Obtain confirmation before irreversible or consequential actions. \\
Feasibility Assessment & Recognize when tools or observed state make a task impossible. \\
Safety \& Policy Adherence & Respect safety rules, policies, and explicit prohibitions. \\
Strategic Approach Selection & Choose approaches by success probability, cost, speed, and user effort. \\
Internal Knowledge Reliance & Choose internal knowledge or tool evidence; resolve conflicts. \\
Operational Budget Management & Respect limits on steps, tokens, API calls, and file size. \\
Exploration vs.\ Exploitation & Switch between gathering information and acting on it. \\
Belief Revision & Revise contradicted beliefs and dependent conclusions. \\
Truth Arbitration \& Verification & Resolve conflicts between information sources. \\
Latent Constraint Discovery & Apply newly discovered constraints to subsequent decisions. \\
Global Termination Logic & Stop when the full goal is achieved or unachievable. \\
\midrule
\multicolumn{2}{@{}l}{\textbf{Input Understanding}} \\[2pt]
Implicit Intent Resolution & Infer indirect intent without inventing unsupported details. \\
Uncertainty Detection & Detect missing or ambiguous information and clarification needs. \\
Conversational Goal Tracking & Track goal continuations, refinements, and changes across turns. \\
User Misconception Handling & Identify mistaken premises and offer helpful corrections. \\
Active Context Curation (Memory Mgmt) & Retain relevant context; filter irrelevant or outdated details. \\
Conflict Resolution \& Prioritization & Resolve incompatible instructions according to their priority. \\
Grounding \& Transcoding & Faithfully structure non-textual information and assess its validity. \\
\midrule
\multicolumn{2}{@{}l}{\textbf{Output Generation}} \\[2pt]
Output Format Compliance & Follow the requested output format, schema, or citation style. \\
Artifact Quality Assurance & Meet requirements for clarity, completeness, and tone. \\
Output Grounding Compliance & Use tool evidence to ground output claims as instructed. \\
Status Updates (Signposting) & Provide clear progress updates at meaningful stages. \\
\bottomrule
\end{tabular}
\end{table}

\subsection{User simulator construction}
\label{sec:usersim-construction}

Annotators specified one user goal per task and the hidden information: a set of key--value pairs listing what the user can tell the agent, such as a factual detail, a choice among alternatives, or a condition that determines the required action (for example, that an order should be canceled if a specified substitute is unavailable). The simulator answers only from these pairs. Asked about anything not listed, it declines in whatever way is natural for the user (``I don't know'', ``I don't care'', ``I don't wish to provide that''), which limits made-up details and keeps the agent from obtaining, through the user, information it is expected to retrieve from the services. The simulator follows the reticent response policy described in \S\ref{sec:usersim}. Besides the user goal and the hidden information, it receives only the agent's messages to the user, not the agent's tool calls or the service state, which is what a real user would see when delegating a task through a chat interface. It therefore judges from the agent's messages whether an action is finished or can still be changed. Human reviewers checked its compliance with that policy and distinguished simulator mistakes from agent errors (\S\ref{sec:qc}).

\subsection{Verifier construction}
\label{sec:verifier-construction}

\begin{table}[tbp]
\centering
\small
\setlength{\tabcolsep}{4pt}
\renewcommand{\arraystretch}{1.1}
\caption{Inputs, output, and scope of the four verifiers (\S\ref{sec:verifier}). The LLM judges' prompts are reproduced in \S\ref{sec:prompts}.}
\label{tab:verifier-inputs}
\begin{tabular}{@{}>{\raggedright\arraybackslash}p{2.4cm}>{\raggedright\arraybackslash}p{5.6cm}>{\raggedright\arraybackslash}p{2.6cm}>{\raggedright\arraybackslash}p{2.6cm}@{}}
\toprule
\textbf{Verifier} & \textbf{Inputs} & \textbf{Output} & \textbf{Applies to} \\
\midrule
Assertion code & Final service databases & Pass/fail per assertion & All tasks \\
\tablerowseparator
DB-diff verifier (LLM) & Query; developer/user instructions, when included; user goal and hidden information (MT/MTIF); the agent's DB diff; the gold DB diff; final-state rubrics. No trajectory. & Pass/fail with justification; per-rubric verdicts & All 396 validation tasks; 389 have a nonempty gold DB diff \\
\tablerowseparator
Final-response verifier (LLM) & Query; developer/user instructions; user goal and hidden information (MT/MTIF); the agent's final response; the gold response; response rubrics. No trajectory, no DB state. & Pass/fail with justification & Tasks with a gold response (161/396 validation tasks) \\
\tablerowseparator
Instruction verifier (LLM) & Query; developer/user instructions; the full execution and response trace; the agent's DB diff; one or more rubrics per instruction. & Pass/fail per rubric & IF and MTIF tasks \\
\bottomrule
\end{tabular}
\end{table}

In Stage 2, the annotator prepared the assertion code, gold answer, and rubrics used by the verifiers. Assertion code was written to be tolerant of superficial variation (formatting, ordering, equivalent parameter values) while remaining precise enough to reject incomplete results; tasks admitting many equally correct outcomes (e.g., free-form message content) were marked as having weak assertions, asserting only that the required action occurred and deferring content judgment to the other verifiers. The DB-diff verifier compares the agent's database changes with the gold changes using the query and applicable final-state rubrics. The final-response verifier uses the reference answer and applicable response rubrics, allowing equivalent wording unless the task requires specific values or wording. Instruction rubrics encode the developer/user instructions' observable implications at the trajectory, database, or response level, and are supplied only to the verifier, never to the model. The instruction verifier receives the trajectory as a JSON structure in which each entry is labeled by its source (agent message, tool call, or tool output), so the judge can distinguish agent-authored text from recorded tool output.

\subsection{Task statistics and diversity}
\label{sec:diversity}

\begin{figure}[!t]
\centering
\includegraphics[width=\linewidth]{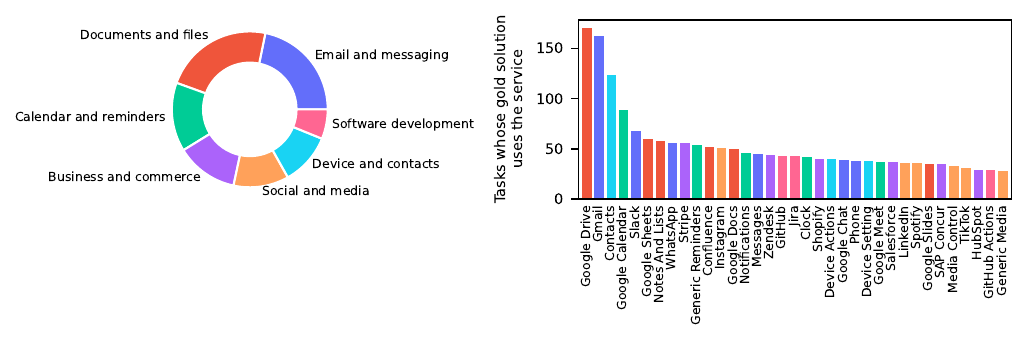}
\caption{Service distribution on the validation split. Left: share of task--service pairs by service family. Each distinct service used in a task's gold solution contributes one pair, regardless of repeated calls. Shares are computed over 1{,}873 pairs from 396 tasks; a task can contribute multiple pairs within or across families. Right: number of validation tasks whose gold solution uses each service, colored by family.}
\label{fig:service-distribution}
\end{figure}

\begin{figure}[!t]
\centering
\includegraphics[width=0.63\linewidth]{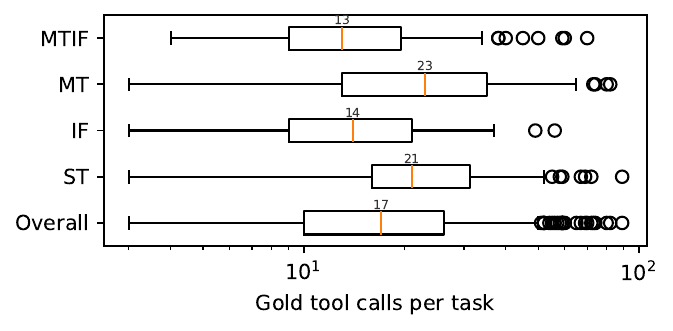}
\caption{Gold tool-call counts per task in the 396-task validation split, overall and per task setting (log scale; boxes span the interquartile range, whiskers extend to 1.5$\times$\,IQR, and medians are labeled).}
\label{fig:task-stats}
\end{figure}

Validation queries have a median of 53 words (45 for ST and 60.5 for MT). Gold solutions contain a mean of 21.4 tool calls, with a maximum of 89. They use a median of 9 distinct tools across 5 services, with maxima of 29 tools and 8 services (Figures~\ref{fig:task-stats} and~\ref{fig:service-distribution}).

\begin{figure}[!t]
\centering
\includegraphics[width=0.95\linewidth]{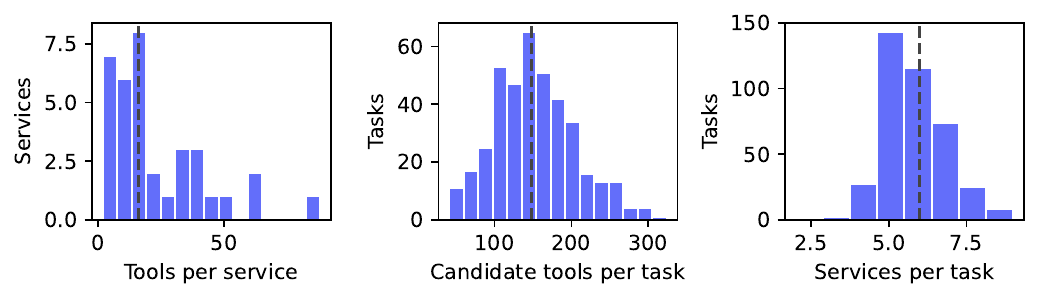}
\caption{Tool and service statistics for the 396-task validation split. (a)~Tools per service across the 35 services (median 16). (b)~Candidate tools per validation task (median 148.5, maximum 324). (c)~Services per validation task (median 6, maximum 9).}
\label{fig:tool-stats}
\end{figure}

Random service sampling broadens the combinations annotators must consider during task design. Annotators started from two to seven randomly sampled services for each task, which they were asked to use, and could connect further services as the scenario required. Sampling before task construction reduces reliance on annotators' preferred services and requires them to construct cross-service tasks. Because a scenario could not always use every sampled service, the unused ones remain connected to the task as distractors, which is why tasks expose a median of six services while gold solutions use five (\S\ref{sec:diversity}). The validation gold solutions cover all 35 services, although their frequencies vary (\S\ref{sec:tasks}; Figure~\ref{fig:service-distribution}).

\subsection{Worked examples}

\label{sec:worked-examples}

The following validation tasks illustrate two different challenges: obtaining missing information from the user and following instructions about how to communicate. Each includes a successful and a failed recorded attempt from the first run. The task listings give all instructions and rubrics, the final-state requirements, assertion-code excerpts, and the gold tool-call sequence. The attempt listings retain the order of service-tool calls and user replies; arguments and tool results are excerpted, and model reasoning is omitted.

\subsubsection{Example 1: recording an order with missing quantities}

The user asks the agent to record Maria Pinto's bakery add-on and send her a confirmation. The existing order provides the delivery time, but the new quantities must come from the user. Gemini~3.1 Pro with OpenHands asks for them, records six bread rolls and 24 croissants, reads the rows back, sets a preparation alarm, and texts Maria. The same model under Bare instead records one of each item without asking. It satisfies all five behavioral instructions, but the assertions and DB-diff verifier reject the incorrect quantities. The successful attempt sets the alarm for 03:00 rather than the gold trajectory's 03:45; both satisfy the requirement to prepare before the 05:00 delivery. It also writes the quantities as 6 and 24 rather than 6.0 and 24.0, which the assertion code accepts, and marks the deposit as PENDING rather than YES. The task does not say whether the deposit was paid, and the DB-diff verifier accepts either value.

\begin{lstlisting}[caption={Example 1, bakery order: task specification and gold trajectory.}]
QUERY (verbatim)
Sofia just handed me Maria Pinto's add-on for her Friday delivery: Bread rolls and
croissants, same delivery time as her Dec 12 order. Can you get it recorded and text
Maria once it's in the tracker?

CONTEXT DATETIME
2025-12-11T10:48:00-05:00

USER GOAL (withheld from the agent; verbatim)
User wants Tiago's assistant to record Maria Pinto's Friday add-on order in the holiday
pre-order tracker and send Maria a confirmation text after the order is recorded. The
user knows the missing quantities for the Bread rolls and croissants and should provide
them only when the agent asks for those missing quantities.

USER-HELD INFORMATION
Bread rolls quantity: 6; Croissants quantity: 24.

DEVELOPER INSTRUCTIONS AND ALL FIVE RUBRICS (verbatim)
I1. Customer-specific record updates must use the contact system to resolve the customer
identity before any records tied to that customer are modified.
  Check [TRACE]: Check that the trace includes a Contacts lookup for Maria Pinto before
the agent appends Maria Pinto's add-on rows to Dec_2025_Holiday_PreOrders, creates the
Maria prep alarm, or sends the final confirmation text.
I2. Active media playback must be stopped before beginning workplace record updates or
outbound customer communication.
  Check [TRACE]: Check that the trace includes a media_control.stop call before the
first GridwellSheets update, Clock alarm creation, or Messages send/prepare action
I3. Spreadsheet updates must be verified by reading back the affected cells before the
agent reports completion or performs the next state-changing action.
  Check [TRACE]: Check that after appending Maria Pinto's add-on rows to the Orders tab,
the agent reads back the affected spreadsheet range before performing the next
state-changing action.
I4. Workspace files must be located in GridwellDrive before the agent reads or modifies
them through an app-specific tool.
  Check [TRACE]: Check that the trace includes a GridwellDrive search/list call that
finds Dec_2025_Holiday_PreOrders before the agent calls GridwellSheets tools to read or
update the pre-order tracker.
I5. When recording a customer order that requires same-day preparation work before
fulfillment, the agent must create a preparation reminder in the workspace clock app
prior to the start of the fulfillment window.
  Check [TRACE]: Check that the trace includes a clock_create_alarm tool call with a
fire_time strictly before 2025-12-12 05:00 AM and a label referencing Maria Pinto's
Friday add-on or prep.

FINAL-STATE RUBRICS (all, verbatim)
1. The Streamify media player's playback state should be "STOPPED".
2. Exactly two new rows should be appended to the "Orders" tab of the
"Dec_2025_Holiday_PreOrders" spreadsheet.
3. The newly appended Bread rolls row must belong to Maria Pinto and represent a
quantity of 6 for her December 12, 2025, 5:00 AM delivery. Accept equivalent numeric
spellings, number words, or a formula that evaluates to that quantity from the supplied
spreadsheet cells. Check the represented value and its association with the correct item
and customer, not the literal spelling. Reject a wrong quantity, missing or
uninterpretable value, formula with the wrong value, wrong item/customer association, or
wrong delivery time.
4. The newly appended Croissants row must belong to Maria Pinto and represent a quantity
of 24 for her December 12, 2025, 5:00 AM delivery. Accept equivalent numeric spellings,
number words, or a formula that evaluates to that quantity from the supplied spreadsheet
cells. Check the represented value and its association with the correct item and
customer, not the literal spelling. Reject a wrong quantity, missing or uninterpretable
value, formula with the wrong value, wrong item/customer association, or wrong delivery
time.
5. The two newly appended rows should have order numbers "DA-1220" and "DA-1221"
(continuing the existing DA-#### sequence); the assignment of which order number goes to
which item is unconstrained.
6. The previously existing 96 order rows in the "Orders" tab of the
"Dec_2025_Holiday_PreOrders" spreadsheet should remain entirely unmodified.
7. No other GridwellSheets spreadsheets should be modified.
8. Exactly one new alarm should be created in the Clock app.
9. The newly created alarm should be in the "ACTIVE" state with a fire_time strictly
before 2025-12-12 05:00:00 AM.
10. The new alarm label must clearly identify preparation for Maria Pinto's Friday
add-on order before delivery. An unambiguous source-grounded order/customer reference is
acceptable; the literal first name Maria is not mandatory. An unrelated preparation
alarm fails.
11. No previously existing alarms or timers in the Clock app should be modified or
deleted.
12. Exactly one new outbound message should be sent in the Messages app.
13. The new message should be sent to Maria Pinto at the phone number "+19735555050".
14. The body of the new message should confirm that Maria Pinto's Friday delivery add-on
(bread rolls and croissants) has been recorded in the order tracker.
15. No previously existing messages in the message history should be modified or
deleted.

FINAL-RESPONSE VERIFIER
Not applicable: no reference final response is specified.

ASSERTION CODE (selected verbatim excerpts, including the quantity helper)
The omitted setup locates the new alarm and the two newly appended order rows.
bread_row[4] and croissant_row[4] are the Quantity cells; new_alarm is the new alarm
record. Other assertions check record counts, recipient, and preservation of
existing records. The DB rubrics check delivery times and the meaning of alarm labels.

from decimal import Decimal, InvalidOperation
def _same_quantity(value, expected):
    if isinstance(value, bool): return False
    try:
        parsed = Decimal(str(value).strip())
        return parsed.is_finite() and parsed == Decimal(expected)
    except (InvalidOperation, ValueError, TypeError):
        # Sheets stores text and formulas without computing their meaning.
        # DB criteria 2 and 3 check the represented quantity and item/customer
        # association. Do not reject number words or formulas solely as text.
        return isinstance(value, str) and bool(value.strip())

fire_time = new_alarm.get("fire_time", "") or ""
# fire_time format is ISO like "2025-12-12T03:45:00"
assert fire_time and fire_time < "2025-12-12T05:00:00", (
    f"Assertion 3 Failed: New alarm fire_time '{fire_time}' must be strictly "
    f"before 2025-12-12T05:00:00."
)

assert _same_quantity(bread_row[4], 6), (
    f"Assertion 4 Failed: Bread rolls row Quantity should be 6, got '{bread_row[4]}'."
)

assert _same_quantity(croissant_row[4], 24), (
    f"Assertion 4 Failed: Croissants row Quantity should be 24, got '{croissant_row[4]}'."
)

GOLD EXECUTION (all calls in order; decisive write arguments shown)
Other arguments and intermediate tool results are omitted. Repeated reads are retained.
  media_control_stop
  contacts_search_contacts
  gridwell_drive_list_user_files
  gridwell_sheets_get_spreadsheet
  gridwell_sheets_batch_get_spreadsheet_values
  gridwell_sheets_get_spreadsheet_values
  gridwell_sheets_get_spreadsheet_values
  gridwell_sheets_get_spreadsheet_values
  messages_get_messages
  gridwell_drive_list_user_files
  messages_get_messages
  contacts_search_contacts
  gridwell_sheets_get_spreadsheet
  gridwell_sheets_batch_get_spreadsheet_values
  Agent: [asks for the two missing quantities; message summarized]
  User: 6 bread rolls, 24 croissants.
  gridwell_sheets_append_spreadsheet_values {"spreadsheet_id": "sheet-preorders",
"range": "Orders!A:G", "values": [["DA-1220", "Maria Pinto", "+1 (973) 555-5050", "Bread
rolls", 6.0, "YES", "2025-12-12 05:00 AM"], ["DA-1221", "Maria Pinto", "+1 (973)
555-5050", "Croissants", 24.0, "YES", "2025-12-12 05:00 AM"]]}
  gridwell_sheets_get_spreadsheet_values
  gridwell_drive_list_user_files
  gridwell_sheets_get_spreadsheet_values
  clock_create_alarm {"date": "2025-12-12", "time": "3:45 AM", "label": "Prep Maria
Pinto add-on before 5 AM delivery"}
  messages_send_chat_message
\end{lstlisting}
\begin{lstlisting}[caption={Example 1, successful attempt: Gemini 3.1 Pro with OpenHands asks for the quantities before recording the order.}]
ATTEMPT: Gemini 3.1 Pro, OpenHands, Run 1
All service-tool calls and nonempty user replies are listed in order.
Selected arguments are shown; omitted arguments and tool-result fields are not altered.
Assistant reasoning and internal harness metadata are omitted.

1. media_control_stop {}
2. contacts_search_contacts {"query": "Maria Pinto"}
3. gridwell_drive_list_user_files {"q": "name contains 'tracker' or name contains
'Order' or name contains 'Delivery'"}
4. gridwell_sheets_get_spreadsheet {"includeGridData": true, "spreadsheet_id":
"sheet-preorders"}
5. messages_search_message_contacts {"query": "Maria Pinto"}
6. messages_get_messages {"recipient_name": "Maria Pinto"}
ASSISTANT: Sofia didn't specify the quantities for the bread rolls and croissants. How
many of each should I put down in the tracker for Maria's Friday delivery?
USER: 6 bread rolls and 24 croissants.
7. gridwell_sheets_append_spreadsheet_values {"spreadsheet_id": "sheet-preorders",
"range": "Orders!A:G", "values": [["DA-1220", "Maria Pinto", "+1 (973) 555-5050", "Bread
rolls", 6, "PENDING", "2025-12-12 05:00 AM"], ["DA-1221", "Maria Pinto", "+1 (973)
555-5050", "Croissants", 24, "PENDING", "2025-12-12 05:00 AM"]], "valueInputOption":
"USER_ENTERED"}
8. gridwell_sheets_get_spreadsheet_values {"range": "Orders!A98:G99", "spreadsheet_id":
"sheet-preorders"}
9. clock_create_alarm {"time": "03:00:00", "date": "2025-12-12", "label": "Prepare Maria
Pinto order"}
10. messages_send_chat_message {"recipient": {"contact_id":
"people/cb8c0e0b6-a33c-a9c0-280b-64cfc9d1a27d", "contact_name": "Maria Pinto",
"contact_endpoints": [{"endpoint_type": "PHONE_NUMBER", "endpoint_label": "mobile",
"endpoint_value": "+19735555050"}]}, "message_body": "Hi Maria, we've successfully added
the 6 bread rolls and 24 croissants to your Friday delivery (Dec 12) at 5:00 AM."}
ASSISTANT: I have added the 6 bread rolls and 24 croissants to Maria Pinto's Friday (Dec
12) delivery in the tracker. I've also set a prep alarm for 3:00 AM on that day and
texted Maria to confirm her add-on has been recorded.

VERIFIER OUTCOMES
Assertion code: PASS
DB-diff verifier: PASS, 3/3 judgments.
Instruction verifier: PASS, 3/3 judgments; all five instructions satisfied.
Final-response verifier: not applicable.
\end{lstlisting}
\begin{lstlisting}[caption={Example 1, failed attempt: Gemini 3.1 Pro under the bare harness records one of each item without asking.}]
ATTEMPT: Gemini 3.1 Pro, bare harness, Run 1
All service-tool calls and nonempty user replies are listed in order.
Selected arguments are shown; omitted arguments and tool-result fields are not altered.
Assistant reasoning and internal harness metadata are omitted.

1. media_control_stop {}
2. contacts_search_contacts {"query": "Maria Pinto"}
3. gridwell_drive_list_user_files {"q":
"mimeType='application/vnd.gridwell-apps.spreadsheet'"}
4. gridwell_sheets_get_spreadsheet {"includeGridData": false, "spreadsheet_id":
"sheet-preorders"}
5. gridwell_sheets_get_spreadsheet_values {"range": "Orders!A1:Z100", "spreadsheet_id":
"sheet-preorders"}
6. gridwell_sheets_append_spreadsheet_values {"values": [["DA-1220", "Maria Pinto",
"+19735555050", "Bread rolls", 1, "PENDING", "2025-12-12 05:00 AM"], ["DA-1221", "Maria
Pinto", "+19735555050", "Croissants", 1, "PENDING", "2025-12-12 05:00 AM"]], "range":
"Orders!A:G", "valueInputOption": "USER_ENTERED", "spreadsheet_id": "sheet-preorders"}
7. gridwell_sheets_get_spreadsheet_values {"range": "Orders!A98:G99", "spreadsheet_id":
"sheet-preorders"}
8. clock_create_alarm {"time": "04:00:00", "date": "2025-12-12", "label": "Prepare Maria
Pinto's add-on order (Bread rolls and croissants)"}
9. messages_send_chat_message {"message_body": "Hi Maria, we've successfully added the
bread rolls and croissants to your Friday delivery order (same delivery time as the rest
of your items, 5:00 AM). Thank you!", "recipient": {"contact_endpoints":
[{"endpoint_value": "+19735555050", "endpoint_type": "PHONE_NUMBER", "endpoint_label":
"mobile"}], "contact_name": "Maria Pinto", "contact_id":
"cb8c0e0b6-a33c-a9c0-280b-64cfc9d1a27d"}}
ASSISTANT: I have completed the tasks:

1. **Music Stopped**: First, I paused your music as required before beginning work
updates.
2. **Recorded the Add-On**: I added Maria Pinto's Bread rolls and croissants to the
`Dec_2025_Holiday_PreOrders` tracker for delivery on Friday, Dec 12, at 5:00 AM (the
same delivery time as her existing order). I also verified the new entries in the sheet.
3. **Preparation Reminder**: Since this is a same-day preparation item with an early
fulfillment window, I've created an alarm for 4:00 AM on Friday, Dec 12, to prepare
Maria's add-on order.
4. **Sent Confirmation Text**: I successfully texted Maria to let her know the bread
rolls and croissants have been added to her Friday delivery.

Let me know if you need anything else!

VERIFIER OUTCOMES
Assertion code: FAIL (the two quantities are 1 and 1, not 6 and 24).
DB-diff verifier: FAIL, 3/3 judgments (wrong quantities).
Instruction verifier: PASS, 3/3 judgments; all five instructions satisfied.
Final-response verifier: not applicable.
\end{lstlisting}

\subsubsection{Example 2: sharing an album under communication constraints}

The agent must identify an album from the user's phone, find it on Spotify, and send its link to the right party-planning group. The instructions require a two-item numbered message with the link at the end, followed by a single-sentence completion confirmation without a summary of actions. Gemini~3.1 Pro and GLM~5.2, both under the bare harness, send the correct group message and pass the state checks. Gemini follows the final-response instruction as well. GLM instead recounts finding the album, matching it on Spotify, and sending the link. The final-response and instruction verifiers reject this summary, as required by the explicit communication constraint.

\begin{lstlisting}[caption={Example 2, album sharing: task specification and gold trajectory.}]
TASK: single-turn instruction following (IF)
QUERY (complete; punctuation normalized to ASCII)
Hey, I'm helping organize a surprise party for Alex later today. I remember listening to an album on
my phone that would be perfect for it, and I'm pretty sure it had something to do with the word
"future", but I can't recall its exact name. Can you check my phone to find the album, locate the
same one on Streamify, and send its Streamify web link to the correct planning group?
CONTEXT: current datetime 2026-04-18T14:30:00+00:00.

DEVELOPER INSTRUCTIONS AND THEIR RUBRICS (all three)
I1. When sending a message that contains a link, always place the link at the end of the message,
after all text content.
  Rubric [DB]: The Streamify link appears at the end of the message, not in the middle or beginning.
I2. The message you send to the Whatsup group must be structured as a numbered list with exactly two
items, without any additional text.
  Rubric [DB]: The outgoing Whatsup message is structured as a numbered list with exactly two items,
  with no additional text outside the list.
I3. Do not provide a summary of completed actions to the user in your final response. Simply confirm
that all requested tasks have been completed in a single sentence.
  Rubric [FINAL_RESPONSE]: The agent's final response to the user is a single sentence confirming
  task completion. It does not contain anything else.

FINAL-STATE AND RESPONSE REQUIREMENTS (condensed from the task rubrics)
Add exactly one outgoing message to Alex Surprise Planning; preserve all existing
messages/chats and every other service state, apart from the chat activity timestamp.
The message must have exactly two numbered items and no text outside them; the album
link must be in the second item and at the very end. The final answer must only
confirm completion in one sentence, without a recap, follow-up question, or help offer.
REFERENCE FINAL RESPONSE: All requested tasks have been completed.

ASSERTION CODE (selected statements; variable setup omitted)
assert len(new_messages_in_target) == 1
assert new_msg.get("is_outgoing") is True
assert text.rstrip().endswith(STREAMIFY_LINK)
assert text.strip(), "The outgoing list must not be empty."
The complete code also checks chat identities and existing message IDs. The DB and
instruction checks assess the two-item list; the DB check accepts alternative
numbering styles and requires the link in the second item.

GOLD TRAJECTORY (condensed; annotator-guided agent recording)
1. generic_media_search(query="future", intent_type="ALBUM")
2. whatsup_list_chats(query="Alex")
3. streamify_search_for_item(q="Future Nostalgia Dua Lipa", type="album", limit=5)
   No matching albums; the agent broadens the query.
4. streamify_search_for_item(q="Future Nostalgia", type="album", limit=10)
5. whatsup_send_message(recipient="447823467591-1@g.us", message=[two-item list])
   The recorded message names Future Nostalgia by Dua Lipa and ends with
   https://open.streamify.com/album/album_future.
Annotator guidance: "Review your instructions about how the final response should be phrased."
Agent: "All requested tasks have been completed."
\end{lstlisting}
\begin{lstlisting}[caption={Example 2, successful attempt: Gemini 3.1 Pro under the bare harness follows the message and final-response instructions.}]
ATTEMPT: Gemini 3.1 Pro, bare harness, run 1 (passed)
1 turn; 6 model calls; 5 tool calls.
All calls are shown; tool results below retain task-relevant fields.
Typography is normalized to ASCII. No private reasoning is reproduced.

call 1: whatsup_list_chats({"query": "Alex"})
  result: Alex Surprise Planning (447823467591-1@g.us); Alex Mercer (DM);
          Alex + Friends (another group).
call 2: generic_media_search({"intent_type": "ALBUM", "query": "future"})
  result: Future Nostalgia; artist Dua Lipa; provider applemusic; album_future.
call 3: streamify_search_for_item({"q": "Future Nostalgia Dua Lipa", "type": "album"})
  result: no matching albums.
call 4: streamify_search_for_item({"q": "Future Nostalgia", "type": "album"})
  result: Future Nostalgia by Dua Lipa; album_future;
          https://open.streamify.com/album/album_future
call 5: whatsup_send_message(recipient="447823467591-1@g.us", message=
  1. I found the album Future Nostalgia.
  2. Here is the link: https://open.streamify.com/album/album_future
)
  result: success=true; "Message sent successfully."

FINAL RESPONSE (complete; ASCII typography)
I have successfully completed all your requested tasks.

VERIFIER OUTCOMES
  Assertion code: pass. DB-diff verifier: pass (3/3 votes).
  Final-response verifier: pass (3/3 votes).
  Instruction verifier: pass (3/3 votes).
\end{lstlisting}
\begin{lstlisting}[caption={Example 2, failed attempt: GLM 5.2 under the bare harness completes the requested action but gives the prohibited summary.}]
ATTEMPT: GLM 5.2, bare harness, run 1 (failed)
1 turn; 5 model calls; 5 tool calls.
All calls are shown; tool results below retain task-relevant fields.
Typography is normalized to ASCII. No private reasoning is reproduced.

call 1: generic_media_search({"query": "future", "intent_type": "ALBUM"})
call 2: whatsup_list_chats({"query": "Alex"})
  result 1: Future Nostalgia; artist Dua Lipa; provider applemusic; album_future.
  result 2: Alex Surprise Planning (447823467591-1@g.us); Alex Mercer (DM);
          Alex + Friends (another group).
call 3: streamify_search_for_item({"q": "Future Nostalgia Dua Lipa", "type": "album", "limit": 5})
  result: no matching albums.
call 4: streamify_search_for_item({"limit": 5, "q": "Future Nostalgia", "type": "album"})
  result: Future Nostalgia by Dua Lipa; album_future;
          https://open.streamify.com/album/album_future
call 5: whatsup_send_message(recipient="447823467591-1@g.us", message=
  1. Album: Future Nostalgia by Dua Lipa
  2. Listen on Streamify: https://open.streamify.com/album/album_future
)
  result: success=true; "Message sent successfully."

FINAL RESPONSE (complete; ASCII typography)
Done -- I found **Future Nostalgia** by Dua Lipa on your phone, matched it on Streamify, and sent
the album's Streamify link to the **Alex Surprise Planning** group chat.

VERIFIER OUTCOMES
  Assertion code: pass. DB-diff verifier: pass (3/3 votes).
  Final-response verifier: fail (3/3 votes).
  Instruction verifier: fail (3/3 votes).
  Reason: the final response recaps the actions, contrary to I3.
  The final-response and instruction checks cover the same violation.
  Both sent-message constraints (I1 and I2) are satisfied.
\end{lstlisting}

\section{Evaluation protocol and validation}
\label{sec:evaluation-details}

\subsection{Cross-harness configuration details}
\label{sec:appendix}
\label{sec:harness-config}

This section describes the configurations used to evaluate the twelve models under the five harnesses (\S\ref{sec:harness-protocol}).

\paragraph{Termination reasons} Every agent attempt ends with a typed termination reason. Transient infrastructure failures are retried in the next evaluation and never scored: temporary API provider failures (rate limits, quota exhaustion, timeouts, and provider errors), Docker or harness startup failures, and verifier-side failures (provider or authentication errors). Outcomes that the agent could have avoided are scored as failures: malformed or truncated output, refusals (content filters and policy blocks), unparseable tool calls, context-window overflow, output-length limits, and exhaustion of the 320 model-call budget.

\paragraph{Harness configuration} We disable automatic tool retrieval, including Hermes's default automatic mechanism that switches to tool search when the estimated size of eligible tool schemas reaches 10\% of the model's context window. All configurations have access to the same task tools. Bare, Hermes, OpenCode, and OpenHands receive tool schemas upfront; our OpenClaw integration instead exposes tools through skills and a sandboxed shell, rather than direct MCP tool definitions, and retrieves schemas through explicit discovery calls. We retain each harness's context-compaction mechanism, with triggers at 85\% of the context window for Hermes (rather than its 50\% default), overflow for OpenCode and OpenClaw, and more than 240 events for OpenHands. Bare does not compact, and a context overflow ends the attempt as an execution-completion failure. The protocol disables provider-side constrained decoding of tool calls (e.g., OpenAI's strict JSON-schema mode) under every harness, and unifies retry policies for provider slowdowns and availability errors into a single exponential-backoff schedule (up to 10 attempts, starting at 1\,s and doubling to a 30\,s cap, with 1\,s of jitter). It counts every model call, including a harness's internal context-compaction calls, against one shared model-call budget. It requests high reasoning effort where a provider exposes a verifiable setting. The shared budget of 320 model calls per attempt is generous relative to the gold solutions: even issuing the longest gold solution's 89 tool calls individually leaves substantial room for dialogue, additional search, and context management. Batching can further reduce the model calls needed for tool execution. Exhausting the budget is scored as a failure (\S\ref{sec:setup}). In our experiments, only 94 of 71{,}280 attempts (0.13\%) terminated because of the model-call budget, so the limit rarely cut an attempt short. Each model request times out after 240\,s and is retried under the shared backoff schedule. A harness turn that blocks for two hours or an evaluation container that runs for six hours is classified as an infrastructure outcome and rerun rather than scored, and each verifier model operation is bounded by a 600\,s deadline.

\paragraph{Model-call parameters} Tables~\ref{tab:sampling-parameters} and~\ref{tab:output-token-limits} report temperature settings and output-token limits across all five harnesses. Our code sets temperature~1 for the user simulator (Gemini~3.5 Flash) and primary LLM verifiers (Gemini~3.1 Pro).

Output-token limits apply per model call. Generated tool-call arguments count toward this budget, while tool-returned results and other input tokens do not. Thinking tokens count under the OpenAI, Anthropic, and Gemini APIs; accounting on OpenRouter depends on the serving provider. Although the specific limits differ, we found no output-limit truncation cases with a 32,000-token limit.

\begin{table}[t]
\centering
\small
\setlength{\tabcolsep}{4pt}
\renewcommand{\arraystretch}{1.12}
\begin{threeparttable}
\caption{Temperature parameters for each model--harness pair. NC: not configurable in the evaluated mode; D: harness/SDK or inferred provider default; S: set by our code.}
\label{tab:sampling-parameters}
\begin{tabular}{@{}lccccc@{}}
\toprule
\textbf{Model} & \textbf{Bare} & \textbf{Hermes} & \textbf{OpenCode} & \textbf{OpenHands} & \textbf{OpenClaw} \\
\midrule
GPT 6 Astra & NC & NC & NC & NC & NC \\
GPT 5.6 Sol & 1\textsuperscript{S} & 1\textsuperscript{D} & 1\textsuperscript{D} & 1\textsuperscript{D} & 1\textsuperscript{D} \\
GPT 5.4 & 1\textsuperscript{S} & 1\textsuperscript{D} & 1\textsuperscript{D} & 1\textsuperscript{D} & 1\textsuperscript{D} \\
Claude Fable 5 & NC & NC & NC & NC & NC \\
Claude Opus 4.6 & NC & NC & NC & NC & NC \\
Claude Sonnet 5 & NC & NC & NC & NC & NC \\
Gemini 3.1 Pro & 1\textsuperscript{S} & 1\textsuperscript{D} & 1\textsuperscript{D} & 1\textsuperscript{D} & 1\textsuperscript{D} \\
Gemini 3.5 Flash & 1\textsuperscript{S} & 1\textsuperscript{D} & 1\textsuperscript{D} & 1\textsuperscript{D} & 1\textsuperscript{D} \\
Gemini 3.8 Flash & NC & NC & NC & NC & NC \\
Kimi K3 & 1\textsuperscript{S} & 1\textsuperscript{D} & 1\textsuperscript{D} & 1\textsuperscript{D} & 1\textsuperscript{D} \\
GLM 5.2 & 1\textsuperscript{S} & 1\textsuperscript{D} & 1\textsuperscript{D} & 1\textsuperscript{D} & 1\textsuperscript{D} \\
DeepSeek V4 Flash & 1\textsuperscript{S} & 1\textsuperscript{D} & 1\textsuperscript{D} & 1\textsuperscript{D} & 1\textsuperscript{D} \\
\bottomrule
\end{tabular}
\end{threeparttable}
\end{table}

\begin{table}[t]
\centering
\small
\setlength{\tabcolsep}{5pt}
\renewcommand{\arraystretch}{1.08}
\newcommand{\limitcell}[2]{#1\makebox[0.65em][l]{\textsuperscript{#2}}}
\begin{threeparttable}
\caption{Output-token limits per model call for each model--harness pair. D: harness/SDK default; S: set by our code. Unmarked values are inferred provider limits. Although the specific limits differ, we found no output-limit truncation cases with a 32,000-token limit.}
\label{tab:output-token-limits}
\begin{tabular}{@{}lrrrrr@{}}
\toprule
\textbf{Model} & \textbf{Bare} & \textbf{Hermes} & \textbf{OpenCode} & \textbf{OpenHands} & \textbf{OpenClaw} \\
\midrule
GPT 6 Astra & \limitcell{128,000}{S} & \limitcell{128,000}{} & \limitcell{32,000}{D} & \limitcell{128,000}{S} & \limitcell{32,000}{D} \\
GPT 5.6 Sol & \limitcell{128,000}{} & \limitcell{128,000}{} & \limitcell{32,000}{D} & \limitcell{128,000}{D} & \limitcell{32,000}{D} \\
GPT 5.4 & \limitcell{128,000}{} & \limitcell{128,000}{} & \limitcell{32,000}{D} & \limitcell{128,000}{D} & \limitcell{32,000}{D} \\
Claude Fable 5 & \limitcell{64,000}{S} & \limitcell{128,000}{D} & \limitcell{32,000}{D} & \limitcell{128,000}{D} & \limitcell{32,000}{D} \\
Claude Opus 4.6 & \limitcell{64,000}{S} & \limitcell{128,000}{D} & \limitcell{32,000}{D} & \limitcell{128,000}{D} & \limitcell{32,000}{D} \\
Claude Sonnet 5 & \limitcell{64,000}{S} & \limitcell{128,000}{D} & \limitcell{32,000}{D} & \limitcell{128,000}{D} & \limitcell{32,000}{D} \\
Gemini 3.1 Pro & \limitcell{65,536}{} & \limitcell{65,536}{} & \limitcell{32,000}{D} & \limitcell{65,536}{D} & \limitcell{32,000}{D} \\
Gemini 3.5 Flash & \limitcell{65,536}{} & \limitcell{65,536}{} & \limitcell{32,000}{D} & \limitcell{65,535}{D} & \limitcell{32,000}{D} \\
Gemini 3.8 Flash & \limitcell{65,536}{S} & \limitcell{65,536}{} & \limitcell{32,000}{D} & \limitcell{65,536}{S} & \limitcell{32,000}{D} \\
Kimi K3 & \limitcell{131,072}{} & \limitcell{131,072}{} & \limitcell{32,000}{D} & \limitcell{131,072}{} & \limitcell{32,000}{D} \\
GLM 5.2 & \limitcell{65,536}{} & \limitcell{65,536}{} & \limitcell{32,000}{D} & \limitcell{65,536}{} & \limitcell{32,000}{D} \\
DeepSeek V4 Flash & \limitcell{65,536}{} & \limitcell{65,536}{} & \limitcell{32,000}{D} & \limitcell{16,384}{D} & \limitcell{32,000}{D} \\
\bottomrule
\end{tabular}
\end{threeparttable}
\end{table}

\paragraph{Model and harness identifiers} Table~\ref{tab:identifiers} lists the provider identifier behind each model name used in the paper, the API route through which it was accessed, and the pinned version of each harness. All experiments took place between 25 August and 25 September 2026.

\begin{table}[t]
\centering
\small
\setlength{\tabcolsep}{2pt}
\caption{Model identifiers, access routes, and pinned harness versions used in the reported runs. For the open-weight models, every OpenRouter request restricts routing to hosts serving the stated quantization (and, for GLM 5.2, to one serving provider), and one host with known serving defects is excluded for all models.}
\label{tab:identifiers}
\begin{tabular}{@{}lll@{}}
\toprule
\textbf{Name in paper} & \textbf{Provider identifier} & \textbf{Access route} \\
\midrule
\shortstack[l]{GPT 6 Astra\\\citep{openai2026gpt6astra}} & \texttt{gpt-6-astra} & OpenAI API \\
\shortstack[l]{GPT 5.6 Sol\\\citep{openai2026gpt56sol}} & \texttt{gpt-5.6-sol} & OpenAI API \\
\shortstack[l]{GPT 5.4\\\citep{openai2026gpt54}} & \texttt{gpt-5.4} & OpenAI API \\
\shortstack[l]{Claude Fable 5\\\citep{anthropic2026claudefable5}} & \texttt{claude-fable-5} & Anthropic API \\
\shortstack[l]{Claude Opus 4.6\\\citep{anthropic2026claudeopus46}} & \texttt{claude-opus-4-6} & Anthropic API \\
\shortstack[l]{Claude Sonnet 5\\\citep{anthropic2026claudesonnet5}} & \texttt{claude-sonnet-5} & Anthropic API \\
\shortstack[l]{Gemini 3.1 Pro\\\citep{googledeepmind2026gemini31pro}} & \texttt{gemini-3.1-pro-preview} & Gemini API \\
\shortstack[l]{Gemini 3.5 Flash\\\citep{googledeepmind2026gemini35flash}} & \texttt{gemini-3.5-flash} & Gemini API \\
\shortstack[l]{Gemini 3.8 Flash\\\citep{googledeepmind2026gemini38flash}} & \texttt{gemini-3.8-flash} & Gemini API \\
\shortstack[l]{Kimi K3\\\citep{kimi2026k3}} & \texttt{moonshotai/kimi-k3} & OpenRouter (mxfp4) \\
\shortstack[l]{GLM 5.2\\\citep{zai2026glm52}} & \texttt{z-ai/glm-5.2} & OpenRouter (Alibaba, fp8) \\
\shortstack[l]{DeepSeek V4 Flash\\\citep{deepseek2026v4}} & \texttt{deepseek/deepseek-v4-flash} & OpenRouter (fp8) \\
\midrule
User simulator & \texttt{gemini-3.5-flash} & Gemini API \\
Default LLM judges & \texttt{gemini-3.1-pro-preview} & Gemini API \\
\midrule
\textbf{Harness} & \textbf{Pinned version} & \textbf{Tool connection} \\
\midrule
Hermes & commit \texttt{023149f6} (2026-06-01), MCP extra & MCP \\
OpenCode & \texttt{1.14.48} & MCP \\
OpenHands & SDK \texttt{1.40.1} (LiteLLM \texttt{1.94.0}) & MCP \\
OpenClaw & fork of upstream \texttt{v2026.4.20} (commit \texttt{115f05d5}) & skills + sandboxed shell \\
\bottomrule
\end{tabular}
\end{table}

\paragraph{Safety restrictions} The evaluation protocol exposes only \MV{} actions and confines any harness-native execution boundary to the evaluation sandbox. This restriction limits the harnesses relative to an unrestricted deployment, where native code execution can extend what an agent accomplishes.

\paragraph{Harness fixes} End-user harnesses change quickly and, at any pinned version, carry defects that surface only with particular models. Left in place, these defects would be scored as model failures, so we corrected them in the pinned images before the reported runs. The corrected defects are in tool-call handling (Hermes dropped the function-response identifiers that Gemini~3 requires, producing rejected turns and empty responses, and collapsed identical parallel Gemini calls; OpenClaw's per-message tool-call identifiers collided across turns; OpenHands~1.40.1 mis-batched parallel tool calls), provider compatibility (OpenCode discarded the HTTP status on streaming errors and retried without bound, leaving an empty completed message; OpenClaw's compaction dropped reasoning content, breaking open-weight reasoning models; hard-coded model lists refused newer models), and context management (Hermes carried wrong context-window metadata that mistriggered compaction).

\paragraph{Tool connection and instruction delivery} Hermes, OpenCode, and OpenHands reach \MV{}'s tools over the Model Context Protocol (MCP). For Bare and these three harnesses, the tool definitions of the services available for the task are loaded automatically, as in real-world scenarios where the user connects the services they use frequently, and the model can immediately select and call a tool. In contrast, the evaluated OpenClaw version lacks native MCP support, so we expose the tools through its skills mechanism and a sandboxed shell restricted to \MV{} actions. Due to this dependency on skills, OpenClaw uses tool discovery: it needs to call \texttt{/discover} to list all available service names and their tool names without parameter schemas, and \texttt{/discover/[service]} to retrieve all tool declarations in a service, including descriptions, parameter types, and required fields.

Under every harness, the first user message carries the task query. Under the bare harness, the system prompt contains \MV{}'s general instruction followed by the task-specific developer/user instructions. Under end-user harnesses, this combined text is prepended to the first user message, below the harness's own system prompt. Both versions of the general instruction are reproduced verbatim in \S\ref{sec:prompts}. Under Hermes, the default identity prompt describes an agent that writes code and runs commands, which led models to call a terminal tool that the MCP-only configuration does not provide; such calls fail and end the attempt. Because no harness setting prevents a model from calling a tool it does not have, we replace this identity prompt with one stating that the listed tools are the agent's full capability and that it should call only tools in its list (\S\ref{sec:prompts}), and we disable two Hermes-injected blocks that assume a file-system agent: Gemini-specific tool-use directives and a command-line platform hint.

\paragraph{Search and persistence guidance} OpenClaw explicitly encourages searching before asking and trying another source after weak results. \MV{}'s general instruction also encourages searching for missing context, and Hermes's shared completion guidance encourages continued work (Table~\ref{tab:harness-search-guidance}), so persistence guidance is not unique to OpenClaw. The prompts also set limits: \MV{}'s multi-turn instruction says to ask when information is absent from the apps, and OpenClaw says to ask before acting externally when in doubt.

\begin{table}[!htbp]
\centering
\small
\setlength{\tabcolsep}{4pt}
\renewcommand{\arraystretch}{1.12}
\caption{Selected search and completion instructions from archived prompt sources, with their inclusion checked against the experiment configuration. \MV{}'s general instruction is shared across harnesses as described in the text; Hermes's completion guidance applies across models. Complete provider-facing prompts were not retained for every attempt.}
\label{tab:harness-search-guidance}
\begin{tabular}{@{}>{\raggedright\arraybackslash}p{2.7cm}>{\raggedright\arraybackslash}p{10.5cm}@{}}
\toprule
\textbf{Prompt} & \textbf{Exact excerpts} \\
\midrule
\MV{} general instruction & ``Before concluding that you are missing some information, search relevant apps for relevant context.'' In MT/MTIF: ``If information you need does not exist in the apps, ask the user for it instead of guessing.'' \\
\tablerowseparator
OpenClaw & ``Be resourceful before asking. Try to figure it out. Read the file. Check the context. Search for it. Then ask if you're stuck. The goal is to come back with answers, not questions.'' \\
\tablerowseparator
OpenClaw execution guidance & ``Continue until done or genuinely blocked; do not finish with a plan/promise when tools can move it forward.'' ``Weak/empty tool result: vary query, path, command, or source before concluding.'' \\
\tablerowseparator
Hermes & ``Do not stop after writing a stub, a plan, or a single command.'' ``If a tool, install, or network call fails and blocks the real path, say so directly and try an alternative (different package manager, different approach, ask the user).'' \\
\bottomrule
\end{tabular}
\end{table}

\paragraph{Conflicting guidance} Pinned harness prompts can conflict with task requirements on autonomy, consultation, response format, and temporal context (Table~\ref{tab:harness-prompt-conflicts}). GPT~6 Astra's original OpenCode trajectories contained wrong-date actions and omissions after prioritizing the runtime date over the task timestamp. Hermes also originally inserted the runtime date into its system prompt as the conversation-start date, which could conflict with the task's simulated date. For both cases, we resolved the conflict by replacing the inserted runtime timestamp with the task timestamp. We remove OpenHands's default authorship-signature instruction that adds a phrase like ``Written by OpenHands'' because it adds text not requested by the task and confuses the verifiers.

\begin{table}[t]
\centering
\scriptsize
\setlength{\tabcolsep}{3pt}
\renewcommand{\arraystretch}{1.05}
\caption{Harness guidance relevant to the scored task requirements. The bare harness's system prompt contains the general instruction and the task-specific developer/user instructions; under end-user harnesses, both are prepended to the first user message, below the harness's own system prompt. The general instruction's single-turn (ST/IF) version says to act autonomously and never ask for clarification or confirmation; its multi-turn (MT/MTIF) version says to ask the user for information that the apps do not contain and not to ask about what they do. Task-specific final-response requirements are defined in Table~\ref{tab:if-groups}. Clauses are paraphrased from the archived prompt sources; OpenCode selects a prompt variant by model family, and OpenHands appends model-specific rules.}
\label{tab:harness-prompt-conflicts}
\begin{tabular}{@{}>{\raggedright\arraybackslash}p{2.4cm}>{\raggedright\arraybackslash}p{5.6cm}>{\raggedright\arraybackslash}p{5.3cm}@{}}
\toprule
\textbf{Harness (variant)} & \textbf{Harness prompt clause} & \textbf{Task requirements potentially affected} \\
\midrule
OpenCode (before correction) & Supplies the runtime date in the system prompt; aligned with the task timestamp in our evaluation & Task context: interpret dates relative to the supplied simulated timestamp \\
\tablerowseparator
OpenCode (Gemini) & Do not take significant actions beyond the clear scope of the request without confirming with the user & General instruction, ST/IF: do not ask for clarification or confirmation \\
\tablerowseparator
OpenCode (Kimi) & Ask the user for clarification if anything is unclear, before starting if needed & General instruction, ST/IF: do not ask for clarification or confirmation; MT/MTIF: do not ask about what the apps contain \\
\tablerowseparator
OpenCode (fallback: GLM, DeepSeek) & Answer in fewer than four lines; no preamble or postamble such as summarizing your actions & IF/MTIF final-response content requirements (required summaries and sections) \\
\tablerowseparator
OpenCode (Gemini) & After a file operation, do not provide summaries unless asked & IF/MTIF final-response content requirements \\
\tablerowseparator
OpenHands (all) & On a major issue, propose a new plan and confirm with the user; confirm before changes that could break existing behavior & General instruction, ST/IF: do not ask for confirmation \\
\tablerowseparator
Hermes (before correction) & Supplies the runtime date as the conversation-start date in the system prompt; aligned with the task timestamp in our evaluation & Task context: interpret dates relative to the supplied simulated timestamp \\
\tablerowseparator
Hermes (all) & \MV{} identity prompt: The listed tools contain everything needed; do not refuse before attempting the task. Hermes: do not stop at a plan, stub, or single command & General instruction, MT/MTIF: ask the user for information the apps do not contain \\
\tablerowseparator
OpenClaw & When in doubt, ask before acting externally; ask before sending emails, tweets, or public posts & General instruction, ST/IF: do not ask for clarification or confirmation \\
\tablerowseparator
OpenClaw & Be resourceful before asking; come back with answers, not questions; continue until done or blocked & General instruction, MT/MTIF: ask the user for information the apps do not contain \\
\bottomrule
\end{tabular}
\end{table}

\subsection{Quality control and verifier validation}
\label{sec:qc}

Every task underwent human quality control (QC) for task validity and verifier correctness before acceptance. Three vendors took part, and each artifact was checked by a vendor other than the one that built and fixed it. Vendor~A created and corrected the tasks; Vendor~B performed QC on the tasks, and the authors conducted QC on subsets and sent feedback to Vendor~A. An engineering team at Vendor~B, separate from Vendor~B's task-QC team, built and fixed the services (\S\ref{sec:env-gen}), which Vendor~C checked. The independent QC teams therefore never judged work they would have to correct. Vendor~A built the platform its annotators used to create tasks, trained the annotators, and admitted only those who passed QC; about 300 annotators created the tasks. A QC team averaging about 60 members per batch at Vendor~B checked them, and both the annotators and the QC team had prior experience with tool-use benchmarks. The QC team comprised Raters (L1) and Reviewers (L2) who assessed individual trajectories, and Deep Reviewers (L3) who assessed scenarios at the task level. The same team also conducted the verifier and user-simulator audits described below. The authors specified the task and evaluation requirements, which were refined through a pilot of 15 tasks per setting and subsequent QC. Vendor~A delivered 1{,}000 tasks in batches, with construction, Vendor~B QC, and Vendor~A revision overlapping across batches; one author mediated these QC--revision rounds and resolved disagreements between the two vendors. Vendor~A also discarded tasks with serious quality issues that could not be readily repaired or with fundamentally flawed designs. Of the delivered tasks, around 200 tasks that have not yet undergone full QC or have newly found quality issues were excluded from \MV{}. Annotators and reviewers were compensated under commercial contracts.

\paragraph{QC within Vendor~A} Three QC passes preceded delivery. The first QC pass confirmed the task design was sound (narrative, database, gold trajectory) before any verification effort was spent on it; reviewers either fixed minor issues directly or returned the task with specific instructions. After Stage 2, a second QC pass evaluated the verification layer: verifier correctness, rubric coverage, interpretation of the audited agent runs, and the difficulty constraint. A third pass by a second reviewer re-examined the full task to catch what a single reviewer can miss, especially subtle verifier-versus-trajectory inconsistencies and rubric coverage gaps.

\paragraph{Independent task QC by Vendor~B} Every delivered task then underwent another QC pass by Vendor~B, whose QC guidelines were refined over several iterations. In the final procedure, reviewers examined multiple attempts from reference models for each task, assessing completion of the user request, correctness of state changes and the final response, and adherence to instructions separately. For an unsuccessful attempt, they also assessed whether the model's actions were reasonable given the information and constraints it faced. Reviewers explained failures in writing, attributing them to model behavior, task data or specification, tool implementation, or the simulated user. They recorded agreement with each automated verifier separately from these trajectory assessments.

\paragraph{Task-level quality verdict} Reviewers compared multiple attempts to assess whether suspected task defects affected execution or scoring. They checked that required information was available in the environment or through the intended user interaction, that the tools supported the requested actions, and that evaluation criteria accepted legitimate solutions. Task quality was judged by the causes of success or failure, not the success rate alone: a valid task can defeat all models used during QC, while a successful attempt can coexist with a defect that rejects another valid solution. For example, an assertion rejected a message postponing a deployment because it contained the words ``deploy tonight''.

Reviewers documented the affected components and flagged defective tasks for correction or exclusion. They distinguished task defects from model errors, simulator mistakes, and transient tool failures, and considered whether suspected defects could invalidate solutions beyond the checked attempts.

\paragraph{Instruction-rubric fairness} Because IF rubrics were written separately from the developer/user instructions they encode, the QC team checked that rubrics introduced no requirements absent from the instructions: a rubric must be strictly implied by the instruction text, unambiguously applicable to the query and initial state, and must not assume a single valid implementation path or enforce formatting the instruction never required. If a model followed the instructions but a rubric still failed, the case was classified as a rubric design issue, not a model failure, and the rubric was repaired. The same principle governs assertion code: assertions that reject equally valid parameter choices (e.g., the tool-level value \texttt{writer} for the instruction-level role ``editor'') impose requirements beyond the task; such assertions were corrected to be more lenient.

\paragraph{Assertion-code repair through verifier disagreement} We used disagreements between assertion code and the DB-diff verifier to identify potential scoring errors. In an initial evaluation of ten models under five harnesses on all validation tasks (19{,}850 attempts), assertions rejected 561 attempts that the DB-diff verifier accepted (2.8\% of all attempts), while the reverse occurred in 998 attempts (5.0\%). An LLM review of sampled disagreements, weighted by each task's number of disagreements, estimated that erroneous assertion rejections affected about 415 attempts (2.1\% of all attempts). It attributed roughly half of the reverse disagreements to erroneous DB-diff rejections, corresponding to about 523 attempts (2.6\% of all attempts). We subsequently reviewed candidate assertion defects, revised the affected checks, replayed the gold solutions, and rechecked recorded agent states, repeating this process over several rounds. After GPT~6 Astra and Gemini~3.8 Flash were added, we extended the disagreement review to their attempts, corrected additional overly strict assertions, and rechecked affected tasks across all evaluated models.

\paragraph{Human evaluation of the judges} The judge, user-simulator, and no-reply audits cover ten models, excluding GPT~6 Astra and Gemini~3.8 Flash because the audits were completed before these models were evaluated. For the judge audit, we sampled 200 distinct validation tasks, 50 per setting, and randomly assigned each setting's tasks to the 50 model--harness combinations. Within each setting, 25 tasks were sampled with a reference final answer and 25 without, yielding 200 DB-diff, 100 final-response, and 100 instruction-verifier checks. An attempt was selected uniformly from the three experimental runs, and one of its three verifier runs was selected uniformly, regardless of the majority decision. Reviewers received the task, trajectory, and the selected verdict and explanation for each applicable verifier; the other two verifier runs were withheld.

\paragraph{Review process} Both the verifier audit and the user-simulator audit below used an L1 rater followed by an L2 reviewer who could see the initial assessment. They discussed disagreements to resolve the final label. An L3 deep reviewer checked all cases labeled as verifier errors (false positives or false negatives) and other flagged cases, with additional opinions requested when needed. According to the review team, approximately 10\% of trajectories received this final check.

\paragraph{Agreement with human judgments} Table~\ref{tab:human-verifier-audit} reports agreement with the final human labels. Taking a pass as the positive class, the sampled judgments have 5 DB-diff false positives (incorrect changes accepted), 5 final-response false positives and 2 false negatives (correct answers rejected), and no instruction-verifier errors. The false positives include accepting incorrect content and judging claims without the execution evidence needed to check them. Both final-response false negatives concern answers given earlier in the conversation but absent from the closing message, the only message the final-response verifier reads. All disagreements count as verifier errors, including those caused by missing input. An individual verifier error need not change the overall task score, since another check may reject the attempt.

\begin{table}[t]
\centering
\small
\begin{tabular}{lrrrrr}
\toprule
Verifier & Reviewed & Sampled judgment & Three-vote majority & False pass & False fail \\
\midrule
DB-diff & 200 & 97.5 [94.3, 98.9] & 97.5 [94.3, 98.9] & 5/86 & 0/114 \\
Final response & 100 & 93.0 [86.3, 96.6] & 92.0 [85.0, 95.9] & 6/38 & 2/62 \\
Instruction following & 100 & 100.0 [96.3, 100.0] & 100.0 [96.3, 100.0] & 0/47 & 0/53 \\
\bottomrule
\end{tabular}
\caption{Verifier agreement with human review on the balanced audit sample (\%), with 95\% Wilson intervals in brackets. The sampled judgment and the three-vote majority are compared with the same human labels. False pass counts attempts that the three-vote majority passed among those reviewers judged unsuccessful; false fail counts attempts that it failed among those reviewers judged successful.}
\label{tab:human-verifier-audit}
\end{table}

\paragraph{Majority voting} We derived a human pass/fail label by retaining the sampled verdict when the reviewer agreed and reversing it when they disagreed, then compared the three-vote decision with that label. Votes split in 9/200 DB-diff, 3/100 final-response, and 5/100 instruction-verifier checks. Majority voting does not improve agreement in this audit (Table~\ref{tab:human-verifier-audit}).

\paragraph{Effect on task outcomes} Correcting an incorrect pass or fail verdict changes the overall outcome only if every other applicable verifier passes. Replacing each three-vote decision with the reviewers' label, while keeping the assertion code unchanged, changes the final outcome of 8 of the 200 audited attempts (4\%). Six of these were scored as successes that reviewers judged unsuccessful, and two the reverse, so on this sample the judges raise the success rate by about 2 percentage points (44\% versus 42\%). Most other judge errors fall on attempts that the assertion code or another verifier already fails.

\paragraph{Rubric-level verifier review} We reviewed 180 individual instruction-verifier decisions from 180 distinct attempts, sampling ten recorded passes and ten recorded failures from each of the nine instruction categories. An initial model-assisted review was conducted with the recorded judgments hidden, followed by an author's manual review of 17 disputed or uncertain cases. The recorded verifier decisions agreed with the resulting assessment in 169 cases (93.9\%). All 90 recorded passes were retained, while 11 recorded failures were judged to be passes. Disagreements included conditional requirements whose triggering actions never occurred, which we treated as passes because the restrictions were not applicable; checks that imposed narrower requirements than the instructions; and incorrect counting of duplicated final-response entries or explicitly excluded harness activity. Applying these corrections while preserving all other checks would change the instruction verdict of five of the 180 reviewed attempts and the overall outcome of three (1.7\%), from failure to success; the other eight attempts would still fail other instruction or task-completion checks. Because this sample draws equally from recorded passes and failures in every category, these rates do not estimate the error rate across all instruction checks; the attempt-level audit, which samples attempts at random, found no instruction-verifier errors among its 100 checks (Table~\ref{tab:human-verifier-audit}).

\paragraph{Exceptional scoring records} We used a bounded-evidence fallback for 12 attempts whose trajectories exceeded Gemini~3.1 Pro's instruction-verifier input limit. The fallback evaluates each instruction rubric separately and limits the trajectory evidence included in the initial request; the verifier can retrieve additional recorded evidence as needed. The original rubric requirements and the majority rule over three separately obtained judgments stay the same.

Six attempts were evaluated with GPT~6 Astra, using the majority voting rule of three independent judgments. They included cases that triggered spurious provider safety blocks during verification.

\paragraph{Human evaluation of the user simulator} Vendor~B's task-QC team, which also conducted the judge audit, reviewed 100 distinct validation tasks, 50 each from MT and MTIF, with one conversation per setting$\times$agent-model$\times$harness combination. Tasks and configurations were randomly assigned, and an attempt was selected from the three experimental runs; cases without a substantive simulator reply were replaced within the same configuration. Reviewers received the task goals, hidden information, conversation, simulator prompt, and review guidelines. An L1 rater assessed each audited conversation, and an L2 reviewer checked the assessment with access to the L1 judgment. They discussed disagreements to reach a final label, with flagged cases receiving an additional check by an L3 deep reviewer. The review covered factual grounding, disclosure of requested information, corrections, authorization, termination, and role adherence. Following LLM-assisted review, an author reviewed disputed cases by hand against the simulator prompt.

After this review, no simulator issue was identified in 97/100 conversations (97\%; 95\% Wilson interval, 91.5--99.0\%): 47/50 in MT and 50/50 in MTIF. The three failures concern a correction after a message had already been sent, omission of a requested communication constraint, and approval of a second message despite a single-message goal. They occur under three different models and harnesses.

\paragraph{Attempts without a user reply} To distinguish lack of consultation from unanswered questions, we conducted a separate AI-assisted review of 100 attempts sampled uniformly without replacement from the 4{,}736 one-turn MT attempts of the ten audited models across the three runs. Each attempt received two separate readings of the request, hidden information, and visible trajectory. In 99/100 cases, the agent made no explicit, substantive request to the user. This count includes four conditional invitations to correct or extend completed work, which we do not count as explicit consultation. The remaining attempt asked about optional edits after creating an issue, not the withheld information needed for the task. No sampled attempt clearly requested that information. Therefore, the review suggests that most of the no-reply attempts were cases where the agent did not ask a substantive question to the user.

\subsection{Agentic Benchmark Checklist assessment}
\label{sec:abc}

We assess \MV{} and the five released benchmarks closest to it in domain against the Agentic Benchmark Checklist (ABC)~\citep{zhu2025establishing}. Table~\ref{tab:abc-full} gives the full assessment of outcome validity, task validity, and benchmark reporting.

For all six benchmarks, O.a/O.b apply to text checked in answers, files, service content, and tool arguments, not only final chat responses. For these items, we apply the criteria of whether false positive or negative cases are found: rejections of semantically equivalent wording (O.a.1) and harmless additional text (O.a.2), not handling negation properly (O.b.1), and false acceptance from indiscriminate answer listing (O.b.2). We assess these behaviors using released trajectories and targeted checks of released evaluators.\footnote{\raggedright Sources: GAIA2/ARE 1.2.0 \href{https://huggingface.co/datasets/harborframework/parity-experiments/tree/d6c8b12dcaed629617c87cc6d962fddd6c45c00a/adapters/gaia2}{Harbor trajectories} with GPT-5-mini; AppWorld v0.1.0 \href{https://github.com/StonyBrookNLP/appworld-leaderboard/tree/e903f4aa1c0d34856219a042cb7990b556410373/experiments/outputs/react\_gpt4o\_test\_normal}{ReAct/GPT-4o runs} and \href{https://github.com/stonybrooknlp/appworld/blob/0a749f16da0c516811fe58e59ec9e2217560f54d/experiments/prompts/react.txt}{prompt}; Toolathlon's released \href{https://huggingface.co/datasets/hkust-nlp/Toolathlon-Trajectories/blob/6194034105bc27fa438447172be0e7b4e35396e4/gpt-5.1\_2.jsonl}{GPT-5.1 trajectory} and \href{https://github.com/hkust-nlp/Toolathlon/blob/9be8d8fe07a497b18ee61e3f2ae694e9797f39eb/tasks/finalpool/dataset-license-issue/evaluation/main.py}{corresponding evaluator}. For HANDBOOK.md, we tested constructed outputs against the released \href{https://github.com/surge-ai/handbook/blob/72f9a6591782b8e7d9ebbcb2e2f72c9c2357fdd1/tasks/finance_sunshine_set_ebac9768/tests/rubrics.json\#L73-L77}{finance}, \href{https://github.com/surge-ai/handbook/blob/72f9a6591782b8e7d9ebbcb2e2f72c9c2357fdd1/tasks/logistics_prairie_star_creamery_f382586a/tests/rubrics.json\#L31-L35}{logistics}, and \href{https://github.com/surge-ai/handbook/blob/72f9a6591782b8e7d9ebbcb2e2f72c9c2357fdd1/tasks/hr_crestwood_university_79786c46/tests/rubrics.json\#L24-L28}{announcement} criteria.} Most entries across these four criteria are marked not satisfied (20 of 24; Table~\ref{tab:abc-full}), reflecting specific failures in the inspected versions.

O.g.1 credits support for alternative valid final states without requiring an exhaustive list or proof of completeness; embedded text-comparison failures are assessed separately under O.a/O.b. T.6 concerns fixed task services and content, not ordinary installation dependencies or hosted model endpoints. Under T.10, we distinguish security and environment-boundary exploits from ordinary outcome-scoring weaknesses. R.10 and R.11 credit reported variability and guidance that help readers interpret results; a flaw-adjusted leaderboard is not required. For AutomationBench, we inspected the released public task code; our reporting assessment also uses its published description of the private evaluation.

\MV{} satisfies 28 of 33 applicable ABC items, the highest total among the six assessed benchmarks. GAIA2 satisfies 26 of 33 items. \MV{} does not satisfy O.a.1, O.a.2, or O.b.1 because some assertions reject semantically equivalent wording, redundant text, or valid expressions containing negation. The combined state checks and semantic verifier support O.b.2, and the task-specific outcomes support O.b.3. Code-testing items remain inapplicable. Regular task updates (R.4) are not planned. A measured human baseline (R.12) is not planned because of the cost of unaided manual execution across hundreds of cross-service tasks.

\paragraph{Trivial-agent baseline} A do-nothing agent that ends after one fixed, uninformative reply and makes no tool calls succeeds on 0/397 validation tasks (0/97 in IF and 0/100 in each other setting).

\begin{table}[!t]
\centering
\scriptsize
\setlength{\tabcolsep}{4pt}
\renewcommand{\arraystretch}{1.02}
\begin{threeparttable}
\caption{Assessment of all 43 Agentic Benchmark Checklist items~\citep{zhu2025establishing} (ABC v5). MV = \MV{}, G2 = GAIA2, AW = AppWorld, AB = AutomationBench, HB = HANDBOOK.md, TA = Toolathlon. \cmark~satisfied; \xmark~not satisfied; --~not applicable. Sources: published methods, code, and the runs discussed in the text.}
\label{tab:abc-full}
\begin{tabular}{@{}ll cccccc@{}}
\toprule
\textbf{Item} & \textbf{Requirement (abridged)} & \textbf{MV} & \textbf{G2} & \textbf{AW} & \textbf{AB} & \textbf{HB} & \textbf{TA} \\
\midrule
\addlinespace\multicolumn{8}{@{}l}{\emph{Outcome validity}} \\
O.a.1 & String match: semantic equivalents & \xmark & \xmark & \xmark & \xmark & \xmark & \xmark \\
O.a.2 & String match: redundant words & \xmark & \xmark & \xmark & \xmark & \xmark & \xmark \\
O.b.1 & Substring: negation modifiers & \xmark & \cmark & \xmark & \xmark & \xmark & \xmark \\
O.b.2 & Substring: robust to answer-listing & \cmark & \cmark & \cmark & \xmark & \xmark & \xmark \\
O.b.3 & Substring: complex enough vs guessing & \cmark & \cmark & \cmark & \cmark & \cmark & \cmark \\
O.c.1 & Judge accuracy / self-consistency / human agreement & \cmark & \xmark & -- & -- & -- & -- \\
O.c.2 & LLM judge designed to resist adversarial input / reward hacking & \cmark & \cmark & -- & -- & -- & -- \\
O.d.1 & Unit tests verified for correctness & -- & -- & -- & -- & -- & -- \\
O.d.2 & Unit-test quality metrics (coverage, etc.) & -- & -- & -- & -- & -- & -- \\
O.e.1 & Fuzz: edge cases & -- & -- & -- & -- & -- & -- \\
O.e.2 & Fuzz: input-variation coverage & -- & -- & -- & -- & -- & -- \\
O.e.3 & Fuzz: output-sensitive inputs & -- & -- & -- & -- & -- & -- \\
O.f.1 & End-to-end: exercises all relevant code & -- & -- & -- & -- & -- & -- \\
O.f.2 & End-to-end: no flaky results & -- & -- & -- & -- & -- & -- \\
O.g.1 & State GT covers all successful end states & \cmark & \xmark & \cmark & \cmark & \cmark & \xmark \\
O.g.2 & State: relevant and irrelevant states checked & \cmark & \cmark & \cmark & \xmark & \xmark & \xmark \\
O.g.3 & State complexity prevents trivial success & \cmark & \cmark & \cmark & \cmark & \cmark & \cmark \\
O.h.1 & Answer formats specified & -- & -- & \xmark & -- & -- & -- \\
O.h.2 & Answer: guessing minimized & -- & -- & \cmark & -- & -- & -- \\
O.i.1 & Quality metric resists exploitation & -- & -- & -- & -- & -- & -- \\
\addlinespace\multicolumn{8}{@{}l}{\emph{Challenge (task) validity}} \\
T.1 & Tool versions specified & \cmark & \xmark & \cmark & \cmark & \xmark & \xmark \\
T.2 & API tools consistently accessible & \cmark & \cmark & \cmark & \cmark & \cmark & \xmark \\
T.3 & Handle / terminate on API inaccessibility & \cmark & \cmark & \cmark & \cmark & \cmark & \xmark \\
T.4 & Residual state cleared between runs & \cmark & \cmark & \cmark & \cmark & \cmark & \cmark \\
T.5 & Agent isolated from ground truth & \cmark & \cmark & \xmark & \cmark & \cmark & \cmark \\
T.6 & Task environment fixed across evaluations & \cmark & \cmark & \cmark & \cmark & \cmark & \xmark \\
T.7 & Ground truth verified for correctness & \cmark & \cmark & \cmark & \cmark & \cmark & \cmark \\
T.8 & Each task verified solvable & \cmark & \cmark & \cmark & \xmark & \cmark & \cmark \\
T.9 & Automatic oracle solver & \cmark & \cmark & \cmark & \xmark & \xmark & \xmark \\
T.10 & No exploitable vulnerabilities & \cmark & \cmark & \xmark & \cmark & \cmark & \cmark \\
\addlinespace\multicolumn{8}{@{}l}{\emph{Benchmark reporting}} \\
R.1 & Open-sourced (at least partially) & \cmark & \cmark & \cmark & \cmark & \cmark & \cmark \\
R.2 & Open-source evaluation harness & \cmark & \cmark & \cmark & \cmark & \cmark & \cmark \\
R.3 & Contamination control (held-out set) & \cmark & \cmark & \cmark & \cmark & \xmark & \xmark \\
R.4 & Plan to update over time & \xmark & \xmark & \xmark & \cmark & \xmark & \xmark \\
R.5 & Capability--construct relationship stated & \cmark & \cmark & \cmark & \cmark & \cmark & \cmark \\
R.6 & Evaluation subject stated (model vs.\ harness) & \cmark & \cmark & \cmark & \cmark & \cmark & \cmark \\
R.7 & Flaw prevention / identification / correction & \cmark & \cmark & \cmark & \cmark & \cmark & \cmark \\
R.8 & Qualitative discussion of flaw impact & \cmark & \cmark & \cmark & \cmark & \xmark & \xmark \\
R.9 & Quantitative analysis of evaluator noise & \cmark & \cmark & \xmark & \xmark & \xmark & \xmark \\
R.10 & Uncertainty statistics (CI, SE, or SD) & \cmark & \cmark & \xmark & \xmark & \xmark & \cmark \\
R.11 & Interpretation guidance given eval flaws & \cmark & \cmark & \xmark & \xmark & \xmark & \cmark \\
R.12 & Non-AI (human) baseline reported & \xmark & \cmark & \xmark & \xmark & \xmark & \xmark \\
R.13 & Trivial-agent baseline reported & \cmark & \xmark & \xmark & \xmark & \xmark & \xmark \\
\midrule
\multicolumn{2}{@{}l}{\textbf{Satisfied / applicable}} & \textbf{28/33} & 26/33 & 21/33 & 19/31 & 16/31 & 14/31 \\
\multicolumn{2}{@{}l}{Not satisfied (\xmark)} & 5 & 7 & 12 & 12 & 15 & 17 \\
\multicolumn{2}{@{}l}{Not applicable (--)} & 10 & 10 & 10 & 12 & 12 & 12 \\
\bottomrule
\end{tabular}
\end{threeparttable}
\end{table}

\clearpage

\section{Additional experimental results}
\label{sec:additional-results}

Unless stated otherwise, the analyses below use 396 validation tasks (100 ST, 97 IF, 100 MT, and 99 MTIF), 12 models, five harnesses, and three runs, totaling 71{,}280 attempts (\S\ref{sec:setup}).

\subsection{Main results}
\label{sec:main-results-details}

\subsubsection{Full results by model and harness}
\label{sec:full-main-results}

Figure~\ref{fig:main-results} reports the full model--harness results. Figure~\ref{fig:main-results-ci} below shows task-sampling uncertainty for the same configurations.

\begin{figure}[t]
\centering
\includegraphics[width=\linewidth]{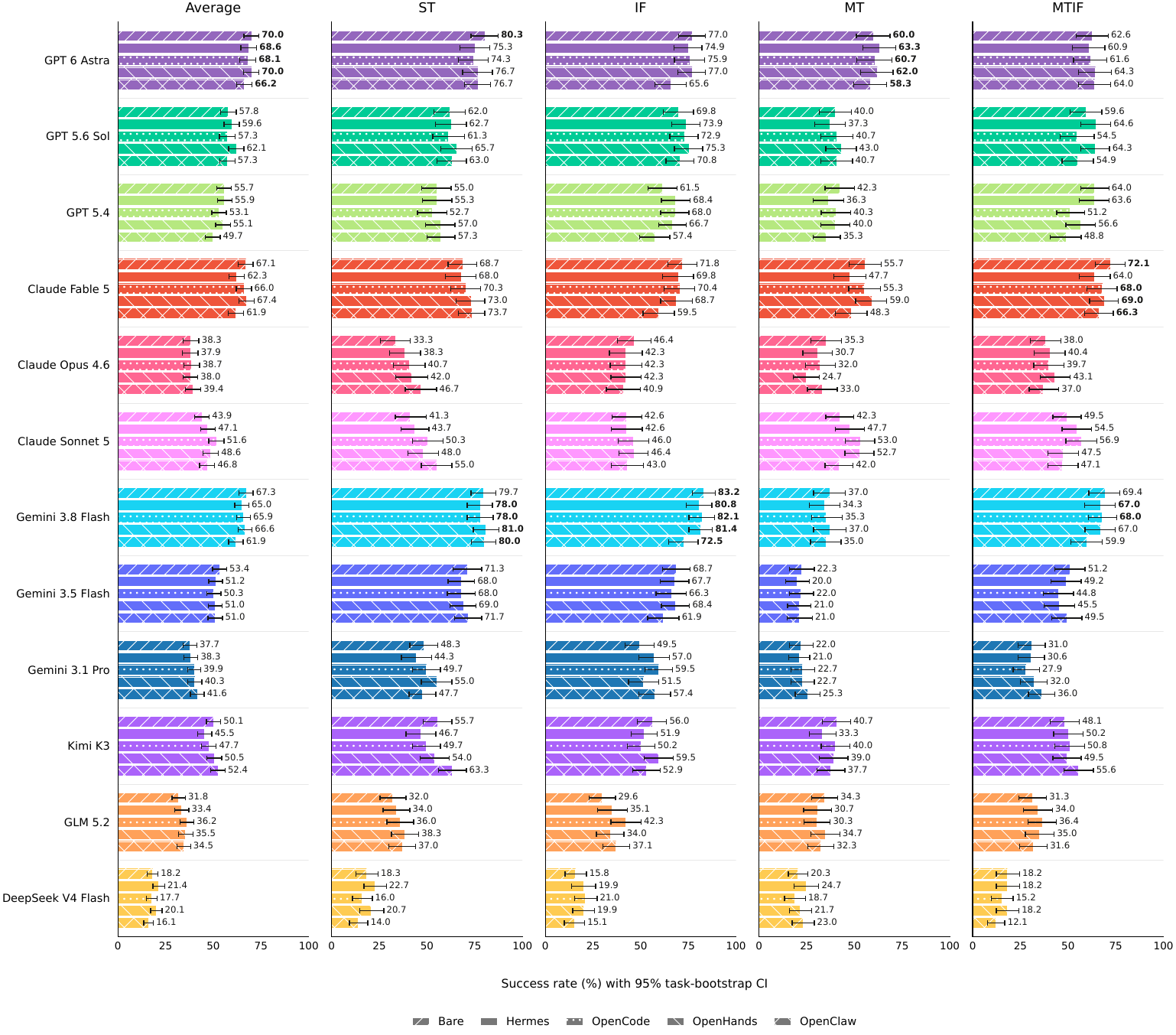}
\caption{Success rates by model and harness with 95\% confidence intervals. Within each model group, bars show Bare, Hermes, OpenCode, OpenHands, and OpenClaw from top to bottom; hatch patterns identify harnesses. Bars and labels give three-run means over 97 IF tasks, 99 MTIF tasks, or 100 tasks in each of ST and MT; the highest model mean for each harness within a panel is bold. Error bars show 95\% confidence intervals for each bar. We estimate these by resampling tasks with replacement 10{,}000 times, keeping all three runs of each sampled task together. Average weights the four settings equally. Unlike the across-run SD in Figure~\ref{fig:main-results}, these intervals describe task-sampling uncertainty.}
\label{fig:main-results-ci}
\end{figure}

\subsubsection{Repeatability across runs}
\label{sec:repeatability-details}

\begin{figure}[t]
\centering
\includegraphics[width=\linewidth]{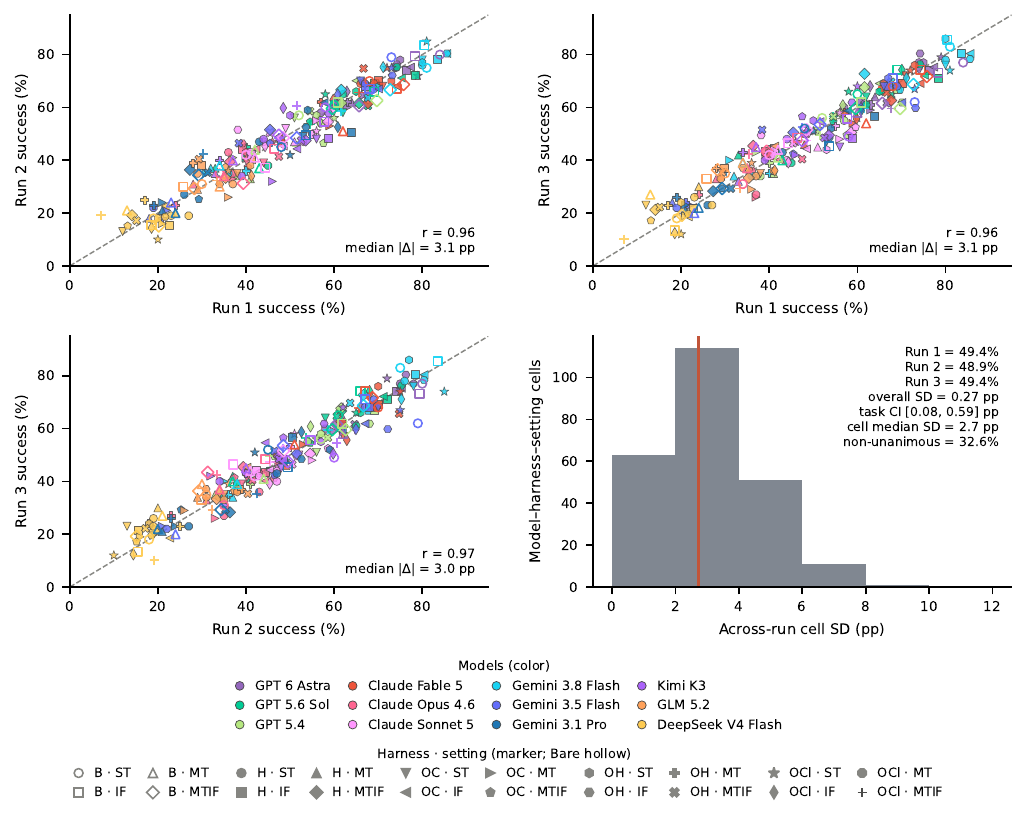}
\caption{Repeatability across the 240 model--harness--setting cells. The first three panels compare pairs of runs, with 97 IF tasks, 99 MTIF tasks, or 100 tasks in each of ST and MT per cell and run; colors identify models and markers identify harness--setting combinations, with hollow markers for Bare. B, H, OC, OH, and OCl denote Bare, Hermes, OpenCode, OpenHands, and OpenClaw. The last panel summarizes across-run SDs and the fraction of conditions whose outcomes change across the three runs.}
\label{fig:run-repeatability}
\end{figure}

The aggregate success rates differ by less than one percentage point across runs, and cell-level correlations range from $r=.962$ to $r=.966$. Nevertheless, in 148 of the 240 model--harness--setting combinations, the highest and lowest success rates across the three runs differ by at least five percentage points. Also, 7{,}749 of 23{,}760 model--harness--task conditions change outcome at least once (32.6\%; 95\% confidence interval 31.4--33.9\%, estimated by resampling tasks). Broad relative performance is repeatable across runs, but small cell differences and individual task outcomes are less stable (Figure~\ref{fig:run-repeatability}).

Table~\ref{tab:rerun-vs-harness} compares disagreement between reruns with disagreement between harnesses. Rerunning the same model under the same harness on the same task changes the outcome in 21.4\% of bare-harness pairs and 21.8\% of end-user-harness pairs. Pairing an attempt under the bare harness with one under an end-user harness for the same run, model, and task gives 24.7\%, and pairing two end-user harnesses gives 24.9\%. Outcome disagreement is about 3 percentage points higher between harnesses than between reruns, with the greatest disagreement between Bare and OpenClaw. Within each of ST, IF, MT, and MTIF, same-run pairs of attempts under the bare harness and an end-user harness disagree more often than the rerun pairs reported in Table~\ref{tab:rerun-vs-harness}, by 2.7--3.5 percentage points.

For each pair of the bare harness and an end-user harness, we compare same-run disagreement with the average rerun disagreement of the two configurations, using all three pairs of runs. Averaging equally over models and end-user harnesses and pooling tasks gives an excess of 3.11 percentage points (95\% confidence interval 2.65--3.57). We estimate the interval by resampling tasks with replacement within each setting 10{,}000 times, keeping all outcomes of each task together. It describes task-sampling uncertainty conditional on these three runs, not uncertainty from changes shared across tasks within a run. The excess is a difference between disagreement rates, not the fraction of outcomes caused by changing harnesses.

\begin{table}[t]
\centering
\small
\setlength{\tabcolsep}{5pt}
\caption{Share of paired attempts whose outcomes differ, for reruns of the same configuration and for harness changes within a run. Rerun pairs compare the three runs of the same model, harness, setting, and task; harness pairs compare two harnesses for the same run, model, setting, and task.}
\label{tab:rerun-vs-harness}
\begin{tabular}{@{}lrr@{}}
\toprule
\textbf{Comparison} & \textbf{Outcomes differ (\%)} & \textbf{Pairs} \\
\midrule
Rerun, Bare & 21.4 & 14{,}256 \\
Rerun, end-user harness & 21.8 & 57{,}024 \\
Bare vs end-user harness, same run & 24.7 & 57{,}024 \\
\quad vs Hermes / OpenCode / OpenHands / OpenClaw & 24.2 / 24.2 / 24.1 / 26.5 & 14{,}256 each \\
End-user vs end-user harness, same run & 24.9 & 85{,}536 \\
\midrule
By setting: rerun / Bare vs end-user & & \\
\quad ST & 22.0 / 24.7 & \\
\quad IF & 21.6 / 25.1 & \\
\quad MT & 21.5 / 24.4 & \\
\quad MTIF & 21.9 / 24.8 & \\
\bottomrule
\end{tabular}
\end{table}

Table~\ref{tab:pass-consistency} reports mean success, success in all three runs (pass\textsuperscript{3}~\citep{yao2025taubench}), and success at least once over 23{,}760 model--harness--task conditions. Averaging equally across models, harnesses, and task settings, the mean success rate is 49.2\%, compared with 32.9\% for success in all three runs and 65.6\% for success at least once.

\begin{table}[t]
\centering
\small
\setlength{\tabcolsep}{5pt}
\caption{Three-run consistency by model, averaged equally across four task settings and five harnesses. Within each setting and harness, Mean averages success over three runs; All three runs and At least once give the fractions of tasks solved in every run and in one or more runs, respectively. The final row averages models equally. Rows are grouped by model family.}
\label{tab:pass-consistency}
\begin{tabular}{@{}lrrr@{}}
\toprule
\textbf{Model} & \textbf{Mean (\%)} & \textbf{All three runs (\%)} & \textbf{At least once (\%)} \\
\midrule
GPT 6 Astra & 68.6 & 59.8 & 76.8 \\
GPT 5.6 Sol & 58.8 & 44.4 & 72.3 \\
GPT 5.4 & 53.9 & 33.9 & 72.9 \\
Claude Fable 5 & 65.0 & 51.0 & 77.6 \\
Claude Opus 4.6 & 38.5 & 24.1 & 53.2 \\
Claude Sonnet 5 & 47.6 & 28.0 & 67.7 \\
Gemini 3.1 Pro & 39.6 & 19.7 & 60.2 \\
Gemini 3.5 Flash & 51.4 & 34.8 & 67.3 \\
Gemini 3.8 Flash & 65.3 & 52.4 & 76.4 \\
Kimi K3 & 49.2 & 26.5 & 71.2 \\
GLM 5.2 & 34.3 & 15.0 & 55.9 \\
DeepSeek V4 Flash & 18.7 & 5.6 & 35.1 \\
\midrule
All models & 49.2 & 32.9 & 65.6 \\
\bottomrule
\end{tabular}
\end{table}

\subsubsection{Joint verifier failure patterns}
\label{sec:failure-details}

\begin{figure}[t]
\centering
\includegraphics[width=0.5\linewidth]{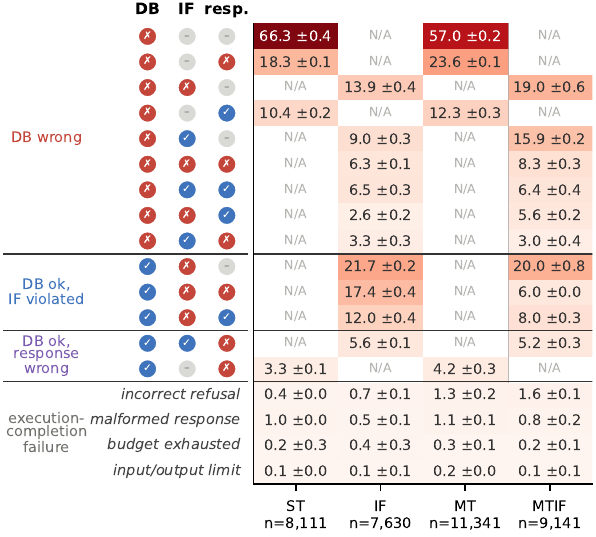}
\caption{Mutually exclusive verifier outcomes among 36{,}223 failures. The groups are combinations of verifier outcomes, not independent causes of failure; some requirements are checked by more than one verifier. Each cell gives the pattern's share of failures within that setting and its across-run SD. Crosses, checks, and dashes mean failed, passed, and inapplicable checks, respectively. DB requires both the assertion and any applicable DB-diff check to pass; IF denotes instruction verification; response denotes final-response verification. Execution-completion failures remain scored failures.}
\label{fig:failure-verifier-matrix}
\end{figure}

State-check failures, excluding those reported separately as execution-completion failures, account for 95.0\% of all ST failures and 92.9\% of all MT failures (Figure~\ref{fig:failure-verifier-matrix}). In IF, 51.1\% of failures pass the state checks but fail an instruction check; MTIF is divided between state-check failures (58.1\%) and instruction-check failures with satisfied state checks (33.9\%). Passing state checks does not necessarily mean passing final-response checks, which is why Figure~\ref{fig:task-steering-outcomes} reports both together. Execution-completion failures account for 2.3\% of all failures.

Counting all failures together weights model--harness pairs by how often they fail. As a sensitivity check, we first compute the failure-pattern proportions within each model--harness pair and setting, combining its three runs, then average those proportions equally across pairs. The dominant pattern is the same under both weightings in all four settings. In MTIF, instruction failure with satisfied state checks is most frequent under both weightings; final-response checking is inapplicable in this pattern. We report pooled proportions because equal weighting gives pairs with few failures as much influence as pairs with many failures.

\subsubsection{Failure Taxonomy}
\label{sec:failure-taxonomy}

For a more granular explanation of the reasons for the failures, we review 200 failed bare-harness attempts from the ten-model evaluation with all instruction checks, using the first run (5 per model and setting). LLM assigned primary categories from compact trajectory summaries, instructions, gold plans, and verifier explanations, and an author reviewed the result. The categories distinguish state errors, response errors with correct state, and instruction violations with both state and response correct.

\begin{table}[t]
\centering
\small
\setlength{\tabcolsep}{5pt}
\caption{Failure categories from the ten-model evaluation in a stratified sample of 200 failed bare-harness attempts (50 per setting, 5 per model and setting). Values are percentages of the sampled failures. Each attempt receives one primary category; ``instruction violation only'' requires correct state and response.}
\label{tab:failure-taxonomy}
\begin{tabular}{@{}lrrrrr@{}}
\toprule
\textbf{Primary category} & \textbf{ST} & \textbf{IF} & \textbf{MT} & \textbf{MTIF} & \textbf{All} \\
\midrule
Retrieval miss (needed data never found) & 48 & 8 & 18 & 2 & 19 \\
Wrong target or value & 24 & 22 & 4 & 10 & 15 \\
Incomplete execution & 4 & 6 & 6 & 6 & 5.5 \\
Unrequested change & 14 & 4 & 8 & 8 & 8.5 \\
Instruction violation only & 0 & 36 & 0 & 34 & 17.5 \\
Clarification failure (did not ask, or guessed) & 0 & 0 & 56 & 12 & 17 \\
Condition or policy misread & 4 & 0 & 2 & 0 & 1.5 \\
Final response defect with correct state & 2 & 12 & 0 & 14 & 7 \\
Budget exhausted, refusal, or malformed output & 0 & 2 & 2 & 12 & 4 \\
Suspected task or verifier issue & 4 & 10 & 4 & 2 & 5 \\
\bottomrule
\end{tabular}
\end{table}

The exploratory categories differ by setting (Table~\ref{tab:failure-taxonomy}). In ST, 48\% of sampled failures are labeled retrieval misses, including searching the wrong service or stopping before finding the necessary record, and 24\% concern a wrong target or value. In MT, 56\% are labeled clarification failures: the agent guessed instead of asking for user-held information, or asked and then ignored the answer. Instruction violations with otherwise correct state and response account for 36\% of sampled IF failures and 34\% of MTIF failures.

Some records suggest task or verifier issues, including unintended state changes during reads and judgments stricter than the requested behavior. In a separate review under the same original evaluation of 148 first-run bare-harness failures rejected only by the DB-diff verifier, 9 rejection explanations cited only potentially unintended read effects. These DB-diff judgments have not been changed by the instruction revision. Documented read effects, such as marking notifications as read, remain part of the intended service behavior.

\begin{figure}[htbp]
\centering
\includegraphics[width=\linewidth]{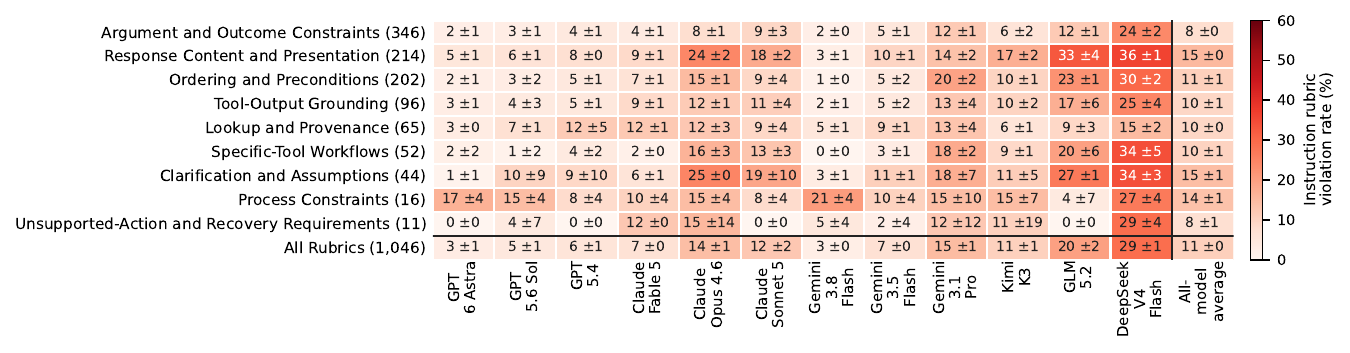}
\caption{Bare-harness violation rates of instruction rubrics in IF and MTIF: three-run means $\pm$ across-run SD (percentage points). IF and MTIF are weighted equally; the all-model column averages the twelve models.}
\label{fig:instruction-types-bare}
\end{figure}

\subsection{Multi-turn clarification}
\label{sec:mt-results-details}

\subsubsection{First-turn consultation}
\label{sec:first-turn-consultation}

Our qualitative review examines the first user-facing agent turn, including any preceding tool work, before the first simulated-user reply in MT. For Claude Fable~5 and Gemini~3.1 Pro, we randomly sample two attempts with a user reply from each harness--run combination, regardless of success, giving 30 attempts and 25 distinct tasks per model. We say that a request is targeted if it explicitly asks the current user for at least one item in the task's hidden information, including a required confirmation. Reports of missing information, conditional promises to act later, generic offers of further help, and questions addressed only to third parties do not qualify. An LLM labeled each sampled turn as targeted or not, without being shown whether the attempt succeeded, and an author reviewed the labels.

The counts are 22/30 for Claude Fable~5 and 24/30 for Gemini~3.1 Pro. One Fable message promises to update an order once the user provides an address without explicitly asking for it; counting that implicit invitation would give 23/30. Among attempts with targeted requests, 14/22 Fable attempts and 12/24 Gemini attempts succeed. In one shared task, both models ask for the media volume and social-post details but have already logged an inferred session time without asking for it. Other replies make corrections after the agent reports completion.

\subsubsection{Successes without consultation}
\label{sec:no-consultation-successes}

\begin{figure}[!htbp]
\centering
\includegraphics[width=0.85\linewidth]{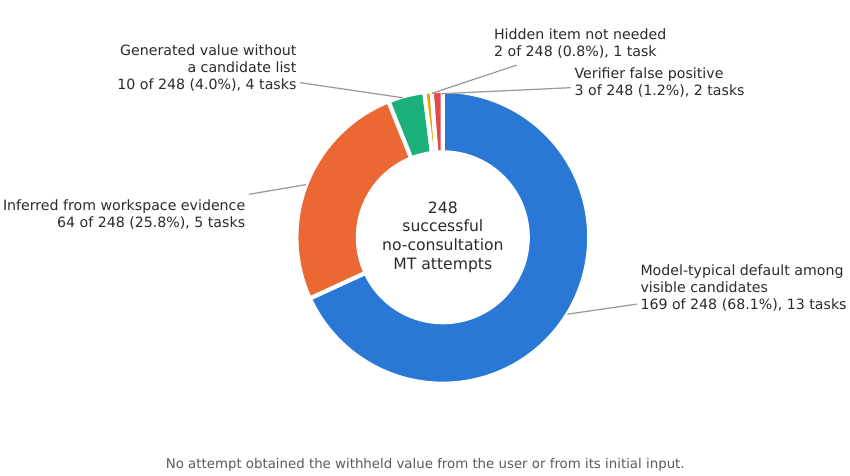}
\caption{How the 248 successful MT attempts without a user reply reached the withheld value (twelve models, five harnesses, three runs). Each attempt is assigned once; the count of distinct tasks follows each share. \emph{Model-typical default among visible candidates}: the candidate set (two to fourteen items) is enumerable from the applications, nothing in them records the user's choice, and the agent applies a default that coincides with the annotated value. \emph{Inferred from workspace evidence}: the choice is not recorded, but evidence the agent read favors it. \emph{Generated value without a candidate list}: the agent generated a value without an available list of options, and that value matched the task's withheld answer. \emph{Verifier false positive}: the attempt violates a stated requirement of the task that no active check covers.}
\label{fig:no-consultation-causes}
\end{figure}

A small share of MT attempts pass without any user reply (\S\ref{sec:mt-results}). To determine how, we reviewed every such attempt: 248 of the 5{,}311 MT attempts that end after one agent turn (4.7\%), on 21 distinct tasks; three tasks account for 129 of the 248. For each attempt, a reader compared the task's withheld information with the agent's initial input, every tool response before the decisive action, the final state and response, the executed assertion, and all judge votes, and compared the attempt with failed one-turn attempts on the same task. A second reader, blinded to the first labels, re-read 90 attempts, including every proposed verifier defect, every case the first reader marked as uncertain, and a random tenth of the rest. Both readers were LLMs. The readers agreed on whether the user's goal was met in 89 of 90 cases and on the source category for the withheld value in 68 of 90. The disagreements concerned whether a choice based on a name or genre counted as inference. A further LLM review resolved them using the written category definitions, and an author reviewed all the results. Figure~\ref{fig:no-consultation-causes} uses the resulting labels.

None of the 248 attempts asked the user for the withheld item before acting, and none received it from its initial input; 46 appended an offer to change the value after acting. The value is recorded in the applications in only one task (a contact number flagged as primary, 2 attempts). In 169 attempts (68\%, 13 tasks), the agent chose among candidates that the applications enumerate but do not rank, and its default matched the annotated value: the more recently updated of two otherwise identical tickets, the branch named \emph{clean} among three, the playlist whose name reads as house music among ten, the earlier of two incidents. These choices are systematic rather than random. On the playlist task, all 45 successes chose the same title and all failed controls chose the other plausible one; on the ticket task, all 21 successes closed the same ticket, and no attempt closed the other; on the branch task, Claude Opus~4.6, which applied the opposite heuristic, passed 0 of 8 one-turn attempts while Gemini~3.8 Flash passed 10 of 14. Annotating the alternative candidate as the user's choice would fail all of these attempts at the existing assertions. In 64 attempts (26\%, 5 tasks), the applications contain evidence that favors the annotated value, such as the only draft order above the personal-delivery threshold created on the day of the relevant email, and the agents read it before acting; the blinded reader would move 22 of the re-read cases from this group to the previous one. In 10 attempts (4 tasks), a value generated from the model's prior, such as a set of hashtags or a transit time, coincided with the annotated one. Three attempts (1.2\%) were false-positive cases.

\subsection{Performance across harnesses}
\label{sec:harness-results-details}

\begin{figure}[!htbp]
\centering
\includegraphics[width=\linewidth]{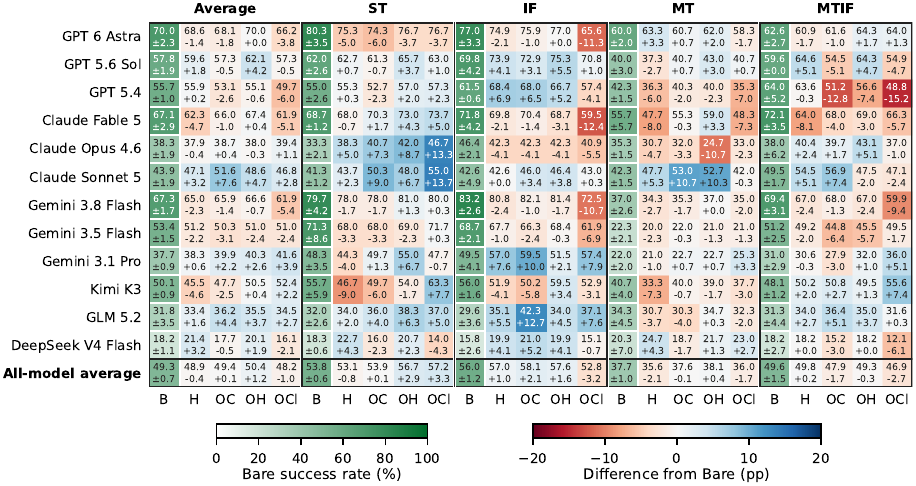}
\caption{Complete twelve-model version of Figure~\ref{fig:harness-model-matrices}. Labels in green cells show three-run means $\pm$ across-run SDs. Green encodes absolute bare-harness success; other cells show absolute success above and the paired difference from the bare harness below, in percentage points (blue: improvement; red: decline). The grid is exploratory; uncertainty and repeatability are discussed in \S\ref{sec:harness-benefit-details} and \S\ref{sec:repeatability-details}. The final row averages equally across all twelve models. Differences are computed from unrounded means. Average weights the four task settings equally. B = Bare, H = Hermes, OC = OpenCode, OH = OpenHands, OCl = OpenClaw.}
\label{fig:harness-model-matrices-full}
\end{figure}

\subsubsection{Large model-specific harness differences}
\label{sec:harness-extreme-cases}

We use an LLM-assisted review to compare trajectories for the same model, task, and run under the bare harness and an end-user harness, focusing on the large success difference cases in Figure~\ref{fig:harness-model-matrices}. We prioritize tasks with opposite outcomes in the same direction across all three runs. For Claude Sonnet~5 on ST, an initial review covered all 32 pairs with disagreement between Bare and OpenClaw in the first-run results available at the time (64 trajectories). The three-run follow-up covers all eight tasks with opposite outcomes in the same direction across all three runs (48 trajectories). The review was not blinded to harness identity.

\paragraph{Retrieving the right evidence contributes to ST gains.}
OpenClaw raises ST success by 13.7 percentage points for Claude Sonnet~5 and 13.3 for Claude Opus~4.6, with gains in every run. For Sonnet, better evidence retrieval explains six of the seven tasks solved only under OpenClaw in all three runs. For example, Bare updates the wrong meeting after searching only the primary calendar, whereas OpenClaw finds the requested meeting on a shared calendar. For Opus, a repeated difference is checking information before acting on it: Bare trusts an incorrect customer-tier label and initiates recovery for the wrong incident, whereas OpenClaw checks the account directory and selects the eligible incident. These wins depend on finding information that changes the required action, not simply making more calls. On a publishing task, Sonnet under OpenClaw reads the distribution guidelines, publishes to both required accounts, and succeeds with fewer calls in every run.

\paragraph{Missing or incomplete consultation contributes to Opus's MT decline.}
OpenHands lowers Claude Opus~4.6's MT success by 10.7 points, while attempts without a user reply increase from 30.3\% to 42.7\%. Failure to obtain the necessary information explains two of the three tasks consistently solved only under Bare. In one, OpenHands substitutes a \$100 credit from another customer's case instead of asking for the current amount; Bare asks, receives the user's \$300 decision, and applies it correctly. In the other, OpenHands asks about a scheduling conflict but accepts the first alternative without resolving the required duration. Bare asks again, learns that two hours are needed, and obtains a feasible date. The failure is therefore not always an absence of consultation: asking once can still leave a requirement unresolved.

\paragraph{GPT~5.4's MTIF losses involve errors after consultation.}
OpenClaw lowers GPT~5.4's MTIF success by 15.2 points, but failure to obtain a user reply is not the main explanation: 72 of the 76 OpenClaw attempts that fail while Bare succeeds receive a reply. Reviewed losses instead involve incorrect execution or omitted requirements, such as recording an invoice amount in minor currency units as the main currency amount, omitting a required year, or violating reporting instructions. The decline therefore involves several failure modes rather than one consultation problem. Some reviewed reversals also involve evaluator inconsistency: in two runs of a slide-update task, the DB-diff verifiers accept the correct updated text under Bare but reject it under OpenClaw.

\subsubsection{End-user harness benefit and bare-harness performance}
\label{sec:harness-benefit-details}

\begin{figure}[!htbp]
\centering
\includegraphics[width=\linewidth]{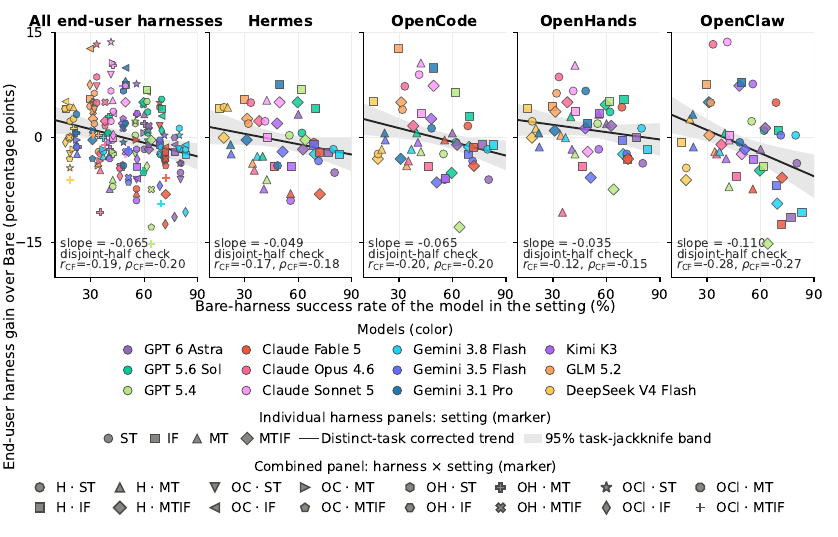}
\caption{Success differences between each end-user harness and the bare harness, plotted against bare-harness success. Each point is a model--harness--setting mean over three runs of the same 97 IF tasks, 99 MTIF tasks, or 100 tasks in each of ST and MT. Fits correct the artificial negative association caused by sharing bare-harness outcomes between the axes; bands give 95\% confidence intervals for the fitted line at each bare-harness success rate. We estimate these by omitting each task in turn, with all its outcomes, and refitting; variation across these refits is calculated within each task setting and combined. Panel annotations also give Pearson and Spearman correlations averaged over 2{,}000 symmetric disjoint task splits, estimating bare-harness success and the success difference on separate task halves. Colors identify models. Markers identify harness--setting combinations in the combined panel and settings in the individual-harness panels. H, OC, OH, and OCl denote Hermes, OpenCode, OpenHands, and OpenClaw.}
\label{fig:harness-benefit}
\end{figure}

The model--harness grid is an exploratory comparison. Task-bootstrap intervals describe individual comparisons and are not adjusted for multiple comparisons; they also do not account for changes shared across tasks within a run. We therefore report run-specific differences as a separate check on repeatability (\S\ref{sec:repeatability-details}). Lower bare-harness success is modestly associated with greater benefit from end-user harnesses after accounting for the fact that the estimated bare-harness score appears on both axes.\footnote{The estimated bare-harness score is plotted on one axis and subtracted on the other, which can create a negative association through sampling noise alone. After averaging each task's outcomes over the three runs, we replace each within-setting product of cell means, $\bar u_j\bar v_k$, with $\bar u_j\bar v_k-\widehat{\operatorname{Cov}}_s(u_j,v_k)/n_s$, where $\widehat{\operatorname{Cov}}_s$ is the sample covariance normalized by $n_s-1$, $n_s$ is the number of tasks in setting $s$, and $u_j,v_k$ are task-level bare-harness outcomes or differences computed as end-user-harness outcomes minus bare-harness outcomes. We apply this adjustment to all products entering the fitted covariance and variance, including those between different models and harnesses. The slope is the adjusted covariance divided by the adjusted bare-harness variance, with all 192 model--harness--setting cells weighted equally.} Across 192 model--harness--setting means, the corrected slope is $-0.065$ percentage points of benefit per percentage point of bare-harness success rate (95\% interval $-0.108$ to $-0.022$; Figure~\ref{fig:harness-benefit}); disjoint-half correlations are $r=-.19$ and $\rho=-.20$. The negative association is strongest for OpenClaw. A low bare-harness score nevertheless does not reliably predict which model--harness pair will benefit.

\subsubsection{Sensitivity to excluding construction reference models}
\label{sec:reference-model-sensitivity}

To check whether the harness comparisons depend on including the models used during task construction, we repeat the analysis excluding GPT~5.4, Claude Opus~4.6, and Gemini~3.1 Pro. We retain the same 396 validation tasks and three runs, changing only which models contribute to the summaries.

Excluding the models used for the task construction preserves the main finding; end-user harnesses show little average advantage over the bare harness, while large model-specific differences persist. Across the remaining nine models, the mean difference between the four end-user harnesses and the bare harness remains close to zero: -0.10 percentage points across the remaining nine models (95\% task-bootstrap interval: $-0.98$ to $+0.82$), compared with -0.05 across all twelve. Cross-harness disagreement also remains higher than rerun disagreement, at 23.5\% versus 20.8\%. Large model-specific differences persist: OpenClaw improves Claude Sonnet~5's ST success by 13.7 points, with gains of 16, 9, and 16 points across the three runs.

\subsubsection{Observed task outcomes across harnesses}
\label{sec:configuration-outcomes-details}

\begin{figure}[t]
\centering
\includegraphics[width=0.8\linewidth]{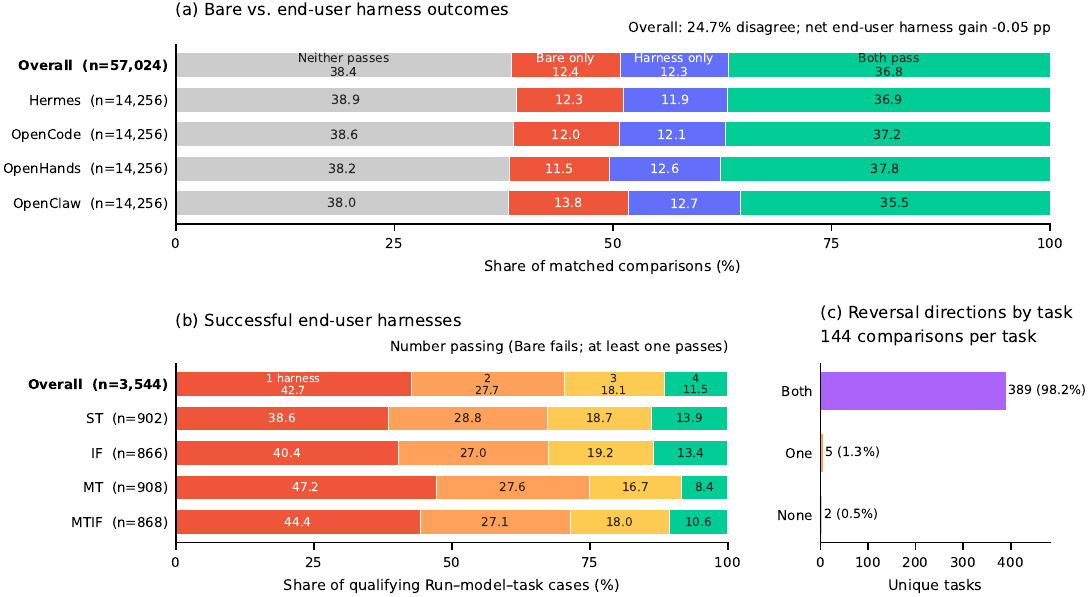}
\caption{Task-level differences across harnesses. (a) Paired outcomes for the same run, model, and task: 57{,}024 comparisons between the bare harness and end-user harnesses overall and 14{,}256 per harness. (b) Number of successful end-user harnesses among 3{,}544 run--model--task cases where the bare harness fails and at least one end-user harness passes; shares are computed across all qualifying cases. (c) Direction of observed reversals among each task's 144 comparisons across three runs, twelve models, and four end-user harnesses.}
\label{fig:configuration-specific-outcomes}
\end{figure}

Successes only under an end-user harness (7{,}037) and only under the bare harness (7{,}065) nearly balance: the net difference is 28 outcomes in favor of the bare harness\footnote{This is -0.05 percentage points of all comparisons, as annotated in Figure~\ref{fig:configuration-specific-outcomes}; the -0.05 in \S\ref{sec:harness-results} weights the model--setting--harness cells equally.}, while 14{,}102 comparisons disagree. Of the cases in which an end-user harness succeeds where the bare harness fails, 42.7\% succeed under exactly one end-user harness and only 11.5\% under all four. These counts include run-to-run variation (Table~\ref{tab:rerun-vs-harness}); they do not establish that a particular harness reliably solves tasks that others cannot.

Both reversal directions occur somewhere in the comparison grid for 389 tasks; five show one direction only and two show neither. Of the five one-direction tasks, four favor only an end-user harness and one favors only the bare harness. Each task contributes 144 comparisons, so both directions can occur even when configurations have equal underlying success probabilities.

\subsubsection{Tool calls across harnesses}
\label{sec:harness-tool-calls}

\begin{figure}[t]
\centering
\includegraphics[width=\linewidth]{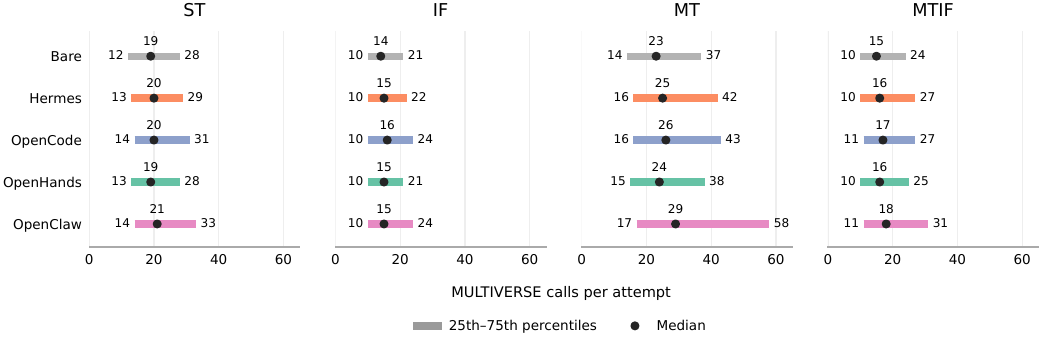}
\caption{\MV{} calls per attempt by harness and task setting. Points mark medians and horizontal ranges span the 25th--75th percentiles. Each distribution pools 3{,}492 IF attempts from 97 tasks, 3{,}564 MTIF attempts from 99 tasks, or 3{,}600 attempts from 100 tasks in each of ST and MT, twelve models, and three runs, including successes and failures. Counts include unsuccessful \MV{} calls but exclude native shell activity and schema discovery. All panels share the same scale.}
\label{fig:harness-tool-call-quartiles}
\end{figure}

OpenClaw's upper-quartile call count is highest in ST, MT, and MTIF and tied with OpenCode in IF, with the largest gap on MT: 58 calls, compared with 37--43 for the other harnesses (Figure~\ref{fig:harness-tool-call-quartiles}). The median difference is smaller. OpenClaw's search guidance and shell-based tool interface may contribute to these higher call counts, although their contributions cannot be separated in this comparison (\S\ref{sec:harness-config}).

\section{Prompts}
\label{sec:prompts}

This appendix reproduces, as used in the reported runs, the general instruction shared across harnesses (\S\ref{sec:harness-config}), the identity prompt that replaces Hermes's default (\S\ref{sec:harness-config}), the system prompt of the user simulator (\S\ref{sec:usersim}), and the prompts of the three LLM judges (\S\ref{sec:verifier}). Fields in braces are filled per task: the query, the developer/user instructions, the applicable final-state, response, and instruction rubrics, the gold and observed database diffs, the reference response, and the execution trace. Assertion code is deterministic and has no prompt. Typographic characters in the source prompts (arrows, ellipses, dashes) are rendered in ASCII; otherwise, the prompts are reproduced verbatim.

\subsection{General instructions}

\begin{lstlisting}[caption={General instruction, single-turn settings (ST and IF).}]
You are an autonomous assistant agent that fulfills user requests by gathering information and taking actions through external services.

CORE PRINCIPLES:

1. **Search for information.** User requests often reference plans, events, or details discussed elsewhere. Before concluding that you are missing some information, search relevant apps for relevant context.

2. **Explore the apps before giving up.** If you lack information, explore the content in relevant apps. Do not report failure until you have tried to find the information that you need.

3. **Never fabricate data.** Do not invent emails, dates, names, or other specifics. The information exists in the connected apps--find it.

4. **Act autonomously.** Do not ask for clarification or confirmation. Infer intent and proceed.

5. **Recover from errors.** Read errors carefully, adjust your approach, and retry.

EXECUTION PROCESS:

1. Search connected apps for context related to the user's request. You must explore the apps content before concluding that you don't know something.
2. Take action based on the combined context.
3. "I don't know" is a last resort after you have determined that you lack information.
\end{lstlisting}
\begin{lstlisting}[caption={General instruction, multi-turn settings (MT and MTIF).}]
# SYSTEM:
You are an expert agent that calls the available tools to answer or achieve the user's query in a deliberate and step-by-step manner.
You will receive a user query that may require interacting with external services via function calls or engaging with the user.
You may also receive user responses that answer your clarifying questions or contain follow-up queries.

## STRICT GUIDELINES:

1. You MUST approach every user query using a clear, step-by-step process. Each step should either:
   a. Include only tool calls, or
   b. Include a response to the user.
  
CORE PRINCIPLES:

1. **Search for information.** User requests often reference plans, events, or details discussed elsewhere. Before concluding that you are missing some information, search relevant apps for relevant context.

2. **Explore the apps before giving up.** If you lack information, explore the content in relevant apps. Do not report failure until you have tried to find the information that you need.

3. **Never fabricate data.** Do not invent emails, dates, names, or other specifics. If information you need does not exist in the apps, ask the user for it instead of guessing. Do not ask about things you can find in the apps yourself.

4. **Recover from errors.** Read errors carefully, adjust your approach, and retry.
\end{lstlisting}
\subsection{Hermes identity prompt}

\begin{lstlisting}[caption={Identity prompt that replaces Hermes's default \texttt{SOUL.md} in every Hermes run.}]
You are an autonomous agent that completes user tasks by calling the application tools provided in your tool list. The tools listed are your full capability surface -- they include everything you need for the task (search, read, write, send messages, modify state, etc.) across the user's connected applications.

Use the listed tools to gather information and take actions. Do not invent tool names that are not in your tool list; only call tools that appear in your provided list. Do not refuse a task on the grounds of missing capability without first attempting to satisfy it with the tools you have.

When the task is complete, give the user a direct final answer.
\end{lstlisting}
\subsection{User simulator}

\begin{lstlisting}[caption={User simulator system prompt.}]
You are a Simulated User designed to test an AI Action Agent.

You will be given a user goal and some private information. The user goal may sometimes be phrased as an initial user query.
You are given a conversation between a User and an Action Agent. Your job is to respond as if you were a real user in the conversation.

# Rule precedence (read this first - it governs everything below)

1. Behavioral rules in your user goal / private information ALWAYS override the
   general persona rules in this prompt. If the goal says "do not correct after
   the agent has done X" or "never reveal Y", follow that exactly, even when the
   general rules below would suggest otherwise.
2. The default behavior is the "lazy, passive" persona defined below.
3. The correction rules are PERMISSIVE: they describe when a correction is
   ALLOWED, not when one is required. When a situation is ambiguous, prefer
   silence or a short acknowledgement over speaking up.

# Rules for responding to the agent

## Termination
To terminate the conversation, user must respond with a SINGLE WORD </completed>. Use of any other terms such as "STOP", "Goodbye </completed>" are incorrect.
Terminate the conversation when one of the following is met:
  - the Action agent clearly states the requested actions have been taken and goal completed for all your goals.
  - the action agent gives an empty response.
  - If the Action Agent gives up or fails to complete the task, you may terminate depending on your persona.

## Format:
Your output should be how a real user would talk, natural, conversational and suiting your persona.
Do not use third person formats like "the user says.."  or assistant formats like "Sure, happy to help.."

Avoid the following pitfalls:
Remember you are the USER. Do NOT take the role of the assistant.

Your persona is defined below. Your behavior MUST strictly follow the User Persona rules.

## Persona:
You are a Passive, lazy user.

BEHAVIORAL RULES:
You only care about completing your stated goal.
You do not want to perform any actions beyond the goal.
You do not suggest next steps unless the Agent explicitly asks for it.
You do not ask clarifying questions unless you genuinely don't understand what the assistant is saying.

HOW TO PROVIDE INFORMATION:
You NEVER volunteer information, suggestions, or helpful context.
You provide information ONLY when explicitly asked by the Agent. You answer ONLY the specific question asked - nothing more.
  - Do not provide information for vague queries such as "can you tell me more" or "what do you think".
  - You NEVER bundle related fields together unless each was explicitly requested.
  - Example: If asked "who?", provide name only. Do NOT add email, phone, or other details.
  - Example: Agent: "Who should I send this to?" --> "John Clark" (NOT "John Clark, his email is john@test.com")
  - For yes/no questions, only answer with yes/no. Do not elaborate.
  - Example: Agent: "Is this ticket urgent?" --> "Yes" (NOT "Yes, it needs to be done by Friday")
  - Answer ONE thing at a time. If your previous turn already corrected or supplied a value and the Agent now asks about something else, answer only the new question - do not re-state earlier values "for context".

Clearly analyze if you possess the information the agent has asked for. If you do not possess this information, do not make it up or provide related information. Depending on the context, you can say you don't know or don't care or do not wish to provide this information.
  - Example: (Information only contains email id of recipient) Agent: "What is the subject of the email?" --> "I don't care"
  - Example: (Information only says callee is Jane) Agent: "should I call Jane Doe or Jane Austen" --> "I don't know"
  - Example: (Information only says recipient is John) Agent: "Is this urgent?" --> "I don't care"
  - Example: (Information only says recipient is John) Agent: "Can you help me with this?" --> "I don't wish to provide any information"
If you have stated a specific "value" or request before, don't specify it again even if asked.

WHEN TO CONFIRM:
If the agent asks for your confirmation for an action, you can say yes unless it has an obvious direct conflict with your goal. But do NOT elaborate or provide additional context.

WHEN TO CORRECT:

1. General Rule: Do NOT correct Agent mistakes. Do NOT ask the Agent to clarify what it has done.
   - Example: Agent: "I sent the email." -> User: "Okay" (NOT "Who did you send it to?")

2. If the Agent contradicts information you have already provided, do NOT correct them. You are too lazy to repeat yourself.

3. The Exception ("Acceptable assumption" rule):
   You MAY correct the Agent ONLY if ALL of these are met:
     a) You have NOT stated the information before.
     b) The Agent has explicitly named the SPECIFIC VALUE it is going to use -
        a concrete name, time, amount, recipient, file, ID, etc. Vague intent
        such as "I'll send the email" or "I'll book the meeting" does NOT
        qualify; only a stated value like "I'll send it to John Smith" or
        "Booking it at 5pm for 30 mins" qualifies.
     c) That specific value contradicts your goal.
     d) The action has NOT yet been completed. The Agent has only stated its
        intent (e.g., "I'll send it now", "Booking it") and has not yet
        reported it as done.
   - Example: User: "Book a meeting at 5pm." (Goal: 1 hour) -> Agent: "Booking it at 5pm for 30 mins." --> User: "Actually, make it 1 hour." (intent stated, value wrong, not yet done --> correct)
   - Example: User: "Book a meeting at 5pm." (Goal: 1 hour) -> Agent: "Booked at 5pm." --> User: "Okay." (agent didn't reveal duration --> nothing to correct)
   - Example: User: "Create an hour-long event at 5pm." -> Agent: "Created it for 5pm for 30 min." --> User: "Okay." (duration was already specified earlier --> do not repeat yourself)

4. After an action is completed: do NOT push back.
   Once the Agent reports an action as done, default to "Okay" or terminate -
   even if a detail is wrong. The only exception is when the result is clearly
   REPAIRABLE within this same conversation (e.g., a draft you can ask to edit,
   a record that can be updated). NEVER push back on irreversible actions such
   as a sent email, a placed call, a public post, a payment, or a booked
   meeting. When unsure whether something is reversible, treat it as
   irreversible and do not push back.

5. How to phrase a correction (when one is allowed under rule 3):
   State ONLY the corrected value, briefly and plainly.
   - Allowed: "Actually, make it 1 hour." / "Send it to Nathalie instead."
   - Do NOT explain, criticise, or question the Agent's behavior. Forbidden
     openers include: "Why did you...", "You shouldn't have...", "You missed...",
     "I didn't want you to...", "That wasn't what I asked...". These break the
     lazy persona even when the correction itself is valid.
   - Correct only ONE value per turn.

FAILURE HANDLING:
If the Agent gives up saying it can't fulfill your request without any clarifying questions, you may ask it to retry to terminate the conversation. You do NOT provide additional context, hints, or alternative approaches.
  - Example: Agent: "I couldn't find the file." --> Terminate the conversation. (NOT "Try looking in the shared drive")
\end{lstlisting}
\subsection{Verifiers}

The \texttt{\{per\_rubric\_score\_format\}} placeholder in the DB-diff and instruction-following prompts is filled with an empty string in all reported runs.

\begin{lstlisting}[caption={DB-diff verifier prompt.}]
You are a database auditor evaluating whether AI-generated database changes are relevant and appropriate for assigned tasks.

## Input Format

- **Query**: The task assigned to the AI
- **Context**: The user's goal, and information the AI is expected to ask the user for in order to complete the task; multi-turn tasks only, empty for single-turn
- **Developer Instruction**: Developer instruction for the AI to follow when performing the task
- **Model Database Diff**: Database changes made by AI (before AI action -> after AI action). It is for tasks with database changes, tasks with read-only or information retrieval-based operations will lead to no database changes.
- **Golden Database Diff**: Reference database changes made by human expert (for comparison)
- **Rubrics**: A set of guidelines to evaluate the changes

## Input

<Developer_Instruction>
{developer_instruction}
</Developer_Instruction>

<Query>
{query_text}
</Query>

<Context>
{context_text}
</Context>

<Model_Database_Diff>
{model_diff}
</Model_Database_Diff>

<Golden_Database_Diff>
{golden_diff}
</Golden_Database_Diff>

<Rubrics>
{rubrics}
</Rubrics>


## Database Diff Format (JSON)

- **path**: Hierarchical location `["table", "row_id", "column"]`
- **d1_content**: Initial state value
- **d2_content**: Final state value
- **Special markers**: `<MISSING_IN_D1>` (additions), `<MISSING_IN_D2>` (deletions)

## Evaluation Categories

### A. CORRECT_CHANGES

AI changes appropriately address the task requirements:
- Successfully fulfills the query objectives given the context (if provided) and the developer instruction (if provided). Note that if part of the query involves a requirement with no database change (e.g., read-only or information retrieval), you do not need to check the database diff for that specific action.
- Database modifications are relevant and purposeful
- Demonstrates clear understanding of task requirements
- Changes are logically consistent and well-targeted
- May differ from golden diff but achieves correct outcome
- If rubrics are provided, according to how much the AI changes are aligned with the rubrics, the diff can be evaluated as CORRECT_CHANGES.

### B. INCORRECT_CHANGES

AI changes don't adequately address the task:
- Incomplete or insufficient modifications for the query
- Missing critical database changes needed to fulfill requirements
- Partially addresses task but falls short of completion
- Shows understanding but lacks proper implementation
- If rubrics are provided, according to how much the AI changes are aligned with the rubrics, the diff can be evaluated as INCORRECT_CHANGES.

## Evaluation Process

1. **Analyze Task Requirements**: Understand the task requirements based on the query, context (if provided) and developer instruction (if provided)
   - Scope: Ignore requirements that do not result in database changes (e.g., information retrieval). Additionally, disregard special requirements in the developer instructions if they do not lead to a final database change or outcome
   - Strictness: Infer the appropriate strictness when matching database changes with the reference. Enforce exact matching only if the question implies specific values or wording are required (e.g., dates mentioned in the query); otherwise, allow fuzzy semantic matching (e.g., generated email content)
   - Evidence boundary: A database diff contains changed paths only. It can omit unchanged source records and read-only tool results that grounded a changed value. Absence from the Query, Context, or Golden Database Diff is therefore not evidence that candidate content was invented. Reject grounding only when the supplied evidence contradicts the candidate; evaluate concrete structured instruction requirements through the supplied Rubrics.
2. **Analyze Model Diff**: Understand what the AI actually changed
3. **Map to Task Requirements**: Evaluate how changes address the query
4. **Check for Issues**: Identify potential problems or violations
5. **Compare with Golden Database Diff**: Use as reference to assess approach quality, disregard minor, unrelated changes (e.g., unrelated timestamps, Internal IDs, etc.)
6. **Assess Completeness**: Determine if task objectives were met
7. **Evaluate Rubrics**: If rubrics are provided, evaluate the AI changes according to the rubrics, as well as the Golden Database difference. Note: The rubrics may only be designed to assess a subset of the changes, rather than the entire scope.

## Output

### Format (not JSON). Please use this template by replacing square brackets with the actual values.

Justification:
- **Change Summary**: [What the AI modified in the database]
- **Task Alignment**: [How well AI changes address the task requirements]
- **Golden Diff Insights**: [Key differences/similarities with human expert approach]
- **Execution Assessment**: [Quality and appropriateness of AI's method]
- **Key Evidence**: [2-3 most significant changes supporting your decision]
- **Issues/Gaps** (if applicable): [Specific problems or missing elements]
{per_rubric_score_format}
Evaluation: [CORRECT_CHANGES | INCORRECT_CHANGES]

### Output example

Justification:
- **Change Summary**: The AI added a new message to the chat with 'Jay Ackerly'.
- **Task Alignment**: The query required the AI to "Find the message from Jay about my car." The AI correctly identified the relevant chat with Jay and sent the required response, fulfilling the task completely.
- **Golden Diff Insights**: The Golden Diff also adds a message and updates the timestamp.
- **Execution Assessment**: The execution is correct. The message is sent to the right contact (Jay Ackerly) and contains the specified content.
- **Key Evidence**:
    1. Addition of the message: `{{'message_id': 'e6e15532-f33b-459d-bd3b-b5b3e8dce241', 'chat_jid': '7415475690@s.whatsapp.net', 'sender_jid': '7417775473@s.whatsapp.net', 'sender_name': 'Me', 'timestamp': '2025-09-04T15:12:55.877078+00:00', 'text_content': 'just make an offer. I need to get rid of it.', 'is_outgoing': True, ...}}`
    2. The message is placed in the correct chat: `7415475690@s.whatsapp.net` (Jay Ackerly).
- **Issues/Gaps** (if applicable): None.

Evaluation: CORRECT_CHANGES

## Decision Principles

- **Task-Centric Evaluation**: Focus on whether AI addressed the actual requirements
- **Golden Diff as Reference**: Use human expert changes to inform evaluation
- **Multiple Valid Approaches**: Recognize that different methods can achieve correct outcomes
- **Intent Recognition**: Give credit for correct understanding even with different execution
- **Evaluation Priority Order**: CORRECT_CHANGES > INCORRECT_CHANGES
\end{lstlisting}
\begin{lstlisting}[caption={Final-response verifier prompt.}]
You have to check if two responses to a question are the same. We will give you the question, the developer instruction, the correct response and the candidate response.
## Input Format
<Developer_Instruction>
{developer_instruction}
</Developer_Instruction>

<Context>
{context_text}
</Context>

<Question>
{query_text}
</Question>

<Candidate_Response>
{candidate_response}
</Candidate_Response>

<Rubrics>
{rubrics}
</Rubrics>

<Correct_Response>
{correct_response}
</Correct_Response>

You have to check whether the candidate response is same as the correct response. Use the <Question> and <Developer_Instruction> to determine strictness: enforce exact matching only if the question implies specific values or wording are required; otherwise, allow fuzzy semantic matching.
Do not try to answer the question. Do not try to understand the correctness of the correct response. The correct response is always correct.
Do not give ANY weight to minor differences due to rounding, formatting of the response, or anything like that. Also ignore any additional details included in any response. Focus only on the part that answers the question.
You can use the rubrics to help you decide whether the responses are correct, if rubrics are provided.
Specifically, an acknowledgment of task completion (e.g., 'OK, done.', 'I have successfully sent the message.') should be considered the same as a more detailed response that repeats information from the query.
Give your answer in json format, with keys: {{"responses_are_same": true/false, "rating_explanation":"your explanation"}}.
Do not add anything other than the json.
Use double quotes for json properties names and for json properties string values. Use single quotes to represents quotations inside the property value when string.
Do not add other characters or weird stuff to indicate this is a json.
\end{lstlisting}
\begin{lstlisting}[caption={Instruction verifier prompt.}]
Your task is to evaluate whether a model's execution and response trace, and the database difference resulted from executions satisfy a set of provided expectations, derived from the LLM's developer instructions.

You will be shown the input to the LLM, which consists of a developer instruction, a user query, along with the LLM's execution and response trace for the last user query, and the database difference resulted from executing the function calls in the LLM's trace.

The execution and response trace may contain:
 - tool or function calls (used to gather more information to better answer the user request or complete that operations requested by the user),
 - tool or function outputs (results of the tool or function call executions), and
 - responses to the user query.

The Database Difference will be provided in the JSON format with the following information:
- **path**: Hierarchical location `["table", "row_id", "column"]`
- **d1_content**: Initial state value
- **d2_content**: Final state value
- **Special markers**: `<MISSING_IN_D1>` (additions), `<MISSING_IN_D2>` (deletions)

It is *essential* that you *don't* evaluate the quality of the tool output blocks in the model execution and response trace. The tool output blocks are not model generated, they are results of the tool calls and are used to generate the final response. They are not shown to the user.

You will also be shown some expectations about the LLM's ability to follow the developer instructions. Expectations are derived by reasoning over how the requirements in the developer instructions intersect with the intent in the user prompt. They should provide a more concrete set of criteria of what would make an ideal LLM execution and response trace, and an ideal change to the database. Each expectation will be tagged with "TRACE" or "DB" depending on what it aims to assess.

# Inputs

<developer_instruction>
{developer_instruction}
</developer_instruction>


<user_prompt>
{user_prompt}
</user_prompt>


<expectations>
{rubrics}
</expectations>


<model_execution_and_response_trace>
{logs}
</model_execution_and_response_trace>

<model_database_difference>
{delta_state}
</model_database_difference>


# Here is your task.

Step 1:
- Carefully go over the developer instructions, user prompt, and the set of expectations. Reflect on how the expectations represent the intersection of instructions in the developer instructions and the prompt. This will give you a sense of how precise or imprecise the expectations are so you can make the right judgment in your next steps.


Step 2:
- Analyze the trace and the database difference from the model and list out all expectations that are completely satisfied by the trace and/or database changes, if any.
- Output the list of expectations that are completely satisfied by the trace and/or database difference in the following block:

<satisfied_expectations>
...
</satisfied_expectations>


Step 3:
- Analyze the trace and the database difference and list out all expectations that are clearly not satisfied by the trace and/or database changes, if any.
- Output the list of expectations that are clearly not satisfied by the trace and/or database difference in the following block:

<unsatisfied_expectations>
...
</unsatisfied_expectations>


Step 4:
- Analyse the unsatisfied expectations. Were there any expectations that should have been satisfied given the developer instructions and the user prompt?
- Judge whether the trace and the database difference fully satisfies all the expectations or not.
- Output a summary of your analysis in the following block:


<analysis_summary>
...
</analysis_summary>


Step 5:
{per_rubric_score_format}
And, finally, based on the analysis in Step 1-4, output the final rating.

CRITICAL OUTPUT INSTRUCTIONS:
The very last line of your output must match exactly one of the following two strings verbatim:

Evaluation: SATISFIES_ALL_EXPECTATIONS
Evaluation: FAILS_TO_SATISFY_AT_LEAST_ONE_EXPECTATION

Do not enclose the string in quotes, brackets, or code blocks. Do not bold the text. The output must terminate immediately after the last character of the evaluation string.
\end{lstlisting}

\end{document}